\documentclass[3p]{elsarticle}
\usepackage{lineno}
\modulolinenumbers[5]

\journal{Artificial Intelligence}

\usepackage[utf8]{inputenc}
\usepackage{url}

\usepackage{breakurl}
\usepackage[breaklinks]{hyperref}
\usepackage{fullpage, indentfirst}
\usepackage[dvipsnames]{xcolor}
\usepackage{graphicx, float}
\usepackage{amsmath, amssymb, mathtools, bbm, amsthm}
\usepackage{subcaption}
\usepackage{wrapfig}
\usepackage{multicol, multirow}
\usepackage{array, booktabs}
\usepackage{pifont} %
\newcommand{\bftab}{\fontseries{b}\selectfont}
\usepackage{ifthen}

\biboptions{sort&compress}

\usepackage{algorithm, algpseudocode}
\algnewcommand{\Comment}[1]{\hfill// \eqparbox{COMMENT\thealgorithm}{#1}}
\makeatletter
\newcounter{phase}[algorithm]
\newlength{\phaserulewidth}
\newcommand{\setphaserulewidth}{\setlength{\phaserulewidth}}

\makeatother
\setphaserulewidth{.7pt}

\usepackage[referable]{threeparttablex}
\newtheorem{theorem}{Theorem}

\theoremstyle{remark}
\newtheorem{remark}{Remark}

\definecolor{porange}{RGB}{231, 117, 0}
\newboolean{showrevision}

\begin{document}

\setboolean{showrevision}{false}
\newcommand{\new}[1]{\ifthenelse{\boolean{showrevision}}{\textcolor{porange}{#1}}{#1}}
\newcommand{\newcaption}[1]{\ifthenelse{\boolean{showrevision}}{\caption{\textcolor{porange}{#1}}}{\caption{#1}}}

\definecolor{darkgreen}{RGB}{0, 100, 0}
\newcommand{\myspace}{\underline{\hspace{5pt}}}
\newcommand{\mytoprule}[1]{\specialrule{#1}{3pt}{3pt}}
\newcommand{\cmark}{\color{darkgreen}\ding{51}} %
\newcommand{\xmark}{\color{red}\ding{55}} %
\newcolumntype{C}[1]{>{\centering\arraybackslash} m{#1} } %

\newcommand{\reals}{{\mathbb{R}}}
\newcommand{\expect}{{\mathbb{E}}}
\newcommand{\entropy}{{\mathcal{H}}}

\newcommand{\state}{{s}}
\newcommand{\ctrl}{{a}}
\newcommand{\obs}{{o}}
\newcommand{\traj}{{\xi}}
\newcommand{\trajObs}{{\xi_\obs}}
\newcommand{\tdisc}{{t}}

\newcommand{\xSet}{{\mathcal{S}}}
\newcommand{\cSet}{{\mathcal{A}}}
\newcommand{\oSet}{{\mathcal{O}}}

\newcommand{\target}{{\mathcal{T}}}
\newcommand{\constraint}{{\mathcal{K}}}
\newcommand{\failure}{{\mathcal{F}}}

\newcommand{\dyn}{{f}}
\newcommand{\costFuncPolicy}{{C}}
\newcommand{\costFunc}{{c}}
\newcommand{\rewardFuncPolicy}{{R}}
\newcommand{\rewardFunc}{{r}}
\newcommand{\outcome}{{\mathcal{V}}}
\newcommand{\valFunc}{{V}}
\newcommand{\qFunc}{{Q}}
\newcommand{\consFunc}{{g}}
\newcommand{\targFunc}{{\ell}}
\newcommand{\policy}{{\pi}}
\newcommand{\policySet}{{\Pi}}

\newcommand{\replay}{{\mathcal{B}}}

\newcommand{\env}{{E}}
\newcommand{\envSetPs}{{\mathcal{M}}}
\newcommand{\envSetPr}{{\mathcal{M}'}}
\newcommand{\envSpace}{{\mathcal{E}}}
\newcommand{\envSpacePs}{{\mathcal{E}}}
\newcommand{\envDist}{{D}}
\newcommand{\latent}{{z}}
\newcommand{\gaussian}{{\mathcal{N}}}
\newcommand{\PAC}{{\textup{PAC}}}
\newcommand{\KL}{{\textup{KL}}}

\newcommand{\perf}{\text{p}}
\newcommand{\backup}{\text{b}}
\newcommand{\shield}{\text{sh}}
\newcommand{\backupProb}{{\rho}}
\newcommand{\shieldProb}{{\epsilon}}

\begin{frontmatter}
\title{Sim-to-Lab-to-Real: Safe Reinforcement Learning with Shielding \\ and Generalization Guarantees}

\author[ece]{Kai-Chieh Hsu\fnref{equal}}
\ead{kaichieh@princeton.edu}
\author[mae]{Allen Z. Ren\fnref{equal}}
\ead{allen.ren@princeton.edu}
\author[ece]{Duy P. Nguyen}
\author[mae]{Anirudha Majumdar\fnref{equal_advise}}
\author[ece]{Jaime F. Fisac\fnref{equal_advise}}

\fntext[equal]{Equal contributions in alphabetical order}
\fntext[equal_advise]{Equal contributions in advising}
\address[ece]{Department of Electrical and Computer Engineering, Princeton University, United States}
\address[mae]{Department of Mechanical and Aerospace Engineering, Princeton University, United States}

\begin{abstract}
Safety is a critical component of autonomous systems and remains a challenge for learning-based policies to be utilized in the real world. In particular, policies learned using reinforcement learning often fail to generalize to novel environments due to unsafe behavior. \new{In this paper, we propose Sim-to-Lab-to-Real to bridge the reality gap with a probabilistically guaranteed safety-aware policy distribution.} To improve safety, we apply a dual policy setup where a performance policy is trained using the cumulative task reward and a backup (safety) policy is trained by solving the Safety Bellman Equation based on Hamilton-Jacobi (HJ) reachability analysis. In \textit{Sim-to-Lab} transfer, we apply a supervisory control scheme to shield unsafe actions during exploration; in \textit{Lab-to-Real} transfer, we leverage the Probably Approximately Correct (PAC)-Bayes framework to provide lower bounds on the expected performance and safety of policies in unseen environments. \new{Additionally, inheriting from the HJ reachability analysis, the bound accounts for the expectation over the worst-case safety in each environment. We empirically study the proposed framework for ego-vision navigation in two types of indoor environments with varying degrees of photorealism.} We also demonstrate strong generalization performance through hardware experiments in real indoor spaces with a quadrupedal robot. See
\url{https://saferoboticslab.github.io/SimLabReal/}
for supplementary material.
\end{abstract}


\end{frontmatter}

\section{Introduction}

Reinforcement Learning (RL) techniques have been increasingly popular in training autonomous robots to perform complex tasks such as traversing uneven outdoor terrains \cite{kumar2021rma} and navigating through cluttered indoor environments \cite{zhu2017target}. 
Through interactions with environments and feedback in the form of \new{a reward signal}, robots learn to reach target locations relying on onboard sensing (e.g., RGB-D cameras). 
In order to achieve good empirical \emph{generalization} performance in different environments, the robot needs to be trained in multiple environments and collect experiences continuously. 
Due to tight hardware constraints and high sample complexities of RL techniques, in most cases, the training is performed solely in simulated environments. 

However, the robots' performance often degrades sharply when they are deployed in the real world, where there can be substantial changes in environments such as different lighting conditions and noise in robot actuation. This performance drop opens the need for research on \emph{Sim-to-Real} transfer. The typical approach is to simulate a large number of environments with randomized properties and train the policy to work well across environments, with the expectation that real environments at deployment time will be well captured by
the rich distribution of training variations. This technique, namely \emph{domain randomization}, has helped bridge the \textit{Sim-to-Real} gap substantially
\citep{tobin2017domain, muratore2021robot, sadeghi2017cad2rl}. In the field of visual navigation, conditions such as camera poses, scene layout, and wall textures can be randomized. However, previous \emph{Sim-to-Real} techniques do not explicitly address \emph{safety} of the robots. Usually, it is worth compromising the performance (e.g., success rate and time needed for reaching the target) to allow better safety of the system (e.g., avoiding dangerous collisions with humans or furniture).
While safety violations are inconsequential in simulation, robots trained without safety considerations will tend to exhibit similar unsafe behavior once deployed in real environments.
Another drawback of these techniques is that they do not provide any guarantees on robots' performance or safety when they are deployed in different real environments. A ``certificate'' of robots' \emph{generalization} performance and safety is necessary before they are deployed in safety-critical environments (e.g., households with children).

\begin{figure}[!t]
    \centering
    \includegraphics[width=\textwidth]{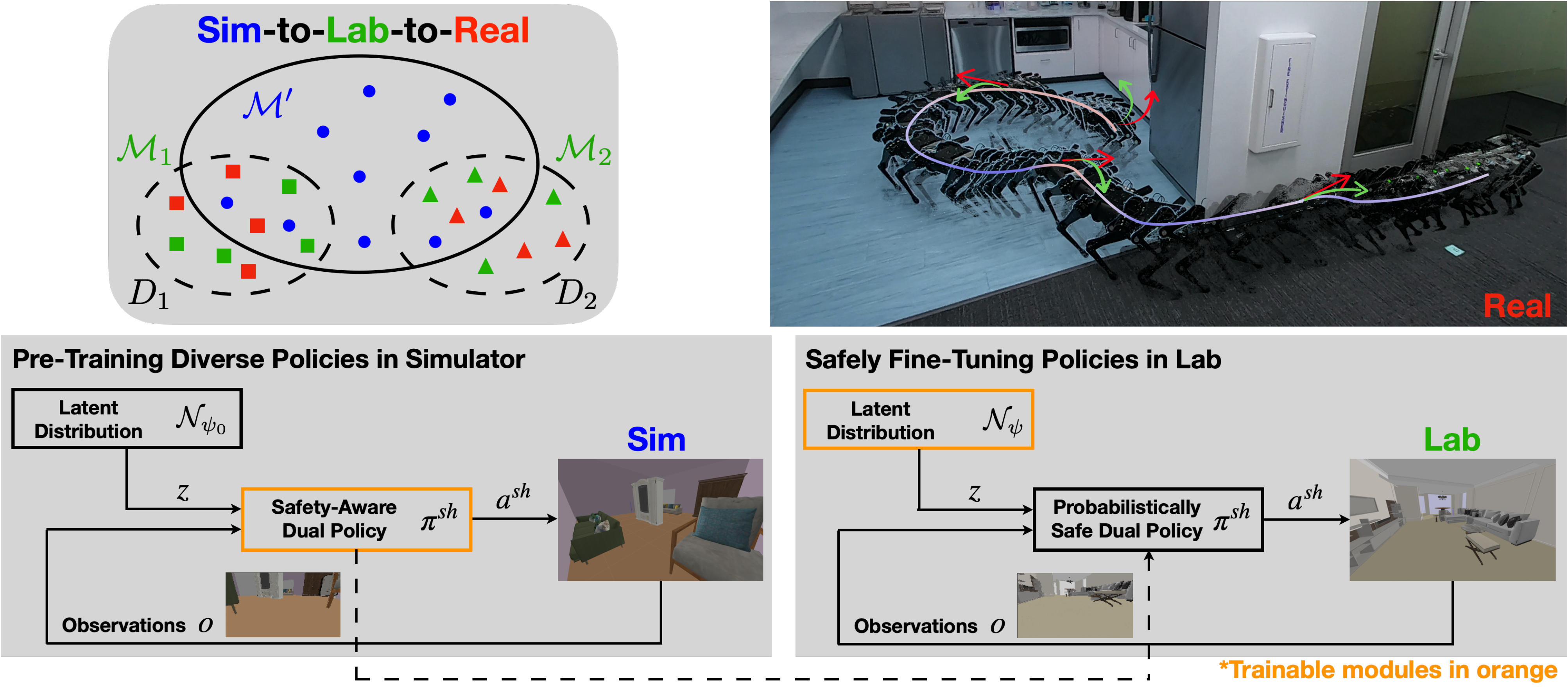}
    \newcaption{\textbf{Overview of the Sim-to-Lab-to-Real framework.}
        \textbf{Top Left:} During \emph{Sim} stage, we train robot policies in a wide variety of environments and conditions $\envSetPr$ (Blue circles). Then the same policies from \emph{Sim} can be fine-tuned in different, more specific settings $\envSetPs_{1,2,...}$ (Green triangles/rectangles) during \emph{Lab} training, which are in the same distribution $\envDist_{1,2,..}$ of Real environments (Red triangles/rectangles).
        For example, we may first train using environments of randomized furniture configurations in \emph{Sim}, and then fine-tune policies in realistic room layouts \cite{fu20213dfront} before deploying in Real indoor spaces.
        \textbf{Bottom:} In \textit{Sim} stage, Sim-to-Lab-to-Real trains a safety-aware dual policy conditioned on latent variable sampled from a distribution, and then safely fine-tunes the latent distribution in \textit{Lab} stage to adapt to a specific environment distribution.
        \textbf{Top right:} Sample trajectory of a quadrupedal robot running trained policy in a real kitchen environment. The backup policy (Green arrow) overrides the performance policy (Red arrow) when the safety critic value (colored trajectory) exceeds some threshold, steering the robot away from obstacles.}
    \label{fig:sim2lab2real}
\end{figure}

In this work, we explore an intermediate training stage between \emph{Sim} and \emph{Real}, which we call \emph{Lab}, that aims to systematically bridge the \emph{Sim-to-Real} gap by explicitly enforcing hard safety constraints on the robot and certifying the performance and safety before deployment in \emph{Real}. The proposed \emph{Sim-to-Lab-to-Real} framework is motivated by the conventional engineering practice whereby, before deploying autonomous systems in the real world, designers usually test them in a realistic but controlled environment, such as a test track for autonomous cars \new{or testing warehouse facilities for quadrupeds at Boston Dynamics \citep{bd2022}}. This standard pipeline opens up an opportunity for autonomous systems to further improve performance and safety in the \emph{Lab} stage. Our insight is that (1) in simulation, where environments can be easily randomized and data is easily collected, the robot can be trained in a wide range of environments and conditions; (2) after that, the robot needs to \emph{fine-tune} in more specific environments before being deployed in similar ones in the real world; (3) this training stage can also certify the system before \emph{Real} deployment, especially if the training can provide \emph{guarantees} on its performance and safety in the real world. Through such extensive training and validation, we can deploy the system confidently in the real environments. Fig.~\ref{fig:sim2lab2real} demonstrates the overview of the proposed Sim-to-Lab-to-Real framework.

Fine-tuning in the \emph{Lab} stage differs from training in the \emph{Sim} stage in that the \textit{Lab} stage is more \emph{safety-critical}. In other words, we want the autonomous system to safely explore in this stage to improve the performance. In order to realize safe \emph{Sim-to-Lab} transfer, we need to consider both (1) how safety is formulated and (2) how safety is ensured throughout training. 
Typical approaches in safe RL combine the safety objective with the performance objective, including adding large negative reward when violating the safety constraints or minimizing the worst-case performance using conditional value at risk (CVaR) formulation \cite{chow2014cvar, chow2017riskconstrained}. However, these methods do not attempt to explicitly enforce hard safety constraints and face a fragile balance between performance and safety. \new{Specifically, these methods require hand-tuning the weights of different components in the objective function, which makes them often fail to generalize well to unseen tasks or environments.} Our approach instead builds upon a dual policy setup, where a performance policy optimizes task reward and a backup (safety) policy keeps the robot away from failure conditions. We then apply a \textit{least-restrictive control law} \cite{fisac2019AGS} (or \emph{shielding}) with the safety state-action value function from the backup policy: the performance policy is only overridden by the backup policy when the safety state-action value function predicts that the proposed performance action would result in an inevitable safety violation in the future.
The backup policy is pre-trained in the \textit{Sim} stage and ready to ensure safe exploration once \emph{Lab} training starts.
\new{Based on the safety policy training developed in \citep{fisac2019bridging, hsu2021safety} using \textit{Hamilton-Jacobi (HJ)  reachability-analysis}, our backup agent can learn from \textit{near-failure} with \textit{dense} signals.
Unlike previous work that uses binary safety failure indicators \cite{srinivasan2020learning, thananjeyan2021recovery},
our training does not rely on experiencing safety violations,
which enables the backup policy to be updated in safety-critical conditions.}
As we show in Sec.~\ref{sec:hj_risk_comparison}, the number of safety violations is reduced by $4\%$--$77\%$ compared to previous safe RL work in different settings. 

We also would like to provide ``certificates'' on the performance and safety of the robot after \textit{Lab} training. However, this can be very challenging since typically it is not possible to fully specify the environment distribution where the robot is deployed (e.g., range of wind velocities for drone navigation, or minimum distance between obstacles for home robot navigation). Traditional techniques that provide performance guarantees for control policies, such as from robust control \cite{zhou1998essentials, xu2002robust} and model-based reachability analysis \cite{majumdar2017funnel, singh2019robust}, typically assume an explicit description of such uncertainty affecting the system (e.g., bound on actuation noise) and/or the environment. 
To tackle the challenge, we apply the \emph{Probably Approximately Correct (PAC)-Bayes Control} framework \citep{majumdar2021pac, farid2021task, veer2020probably}, which provides lower bounds on the expected performance and safety when testing learned policies in unseen environments, while (1) not assuming explicit knowledge of the environment and (2) suited for systems with \new{high-dimensional observations like vision.} 
The framework also naturally fits our setup, as the two training stages of PAC-Bayes Control, \emph{prior} and \emph{posterior}, can be assigned to the \emph{Sim} and \emph{Lab} stages. 
As required by the PAC-Bayes setup, we train a \emph{distribution of policies} by conditioning the performance (and backup) policy on latent variables sampled from a distribution.
After training a \emph{prior} policy distribution in \emph{Sim} stage, we fine-tune the distribution in \emph{Lab}, obtaining a \emph{posterior} policy distribution and its associated generalization guarantee. While previous work in PAC-Bayes Control does not consider explicit policy architecture for safety, we now combine it with HJ reachability analysis and improve the generalization bounds for performance and safety by $40\%$ (Sec.~\ref{sec:pac_bound}).

\subsection{Statement of Contributions}

The primary contribution of this work is to propose \emph{Sim-to-Lab-Real}, a framework that combines HJ reachability analysis and the PAC-Bayes Control framework to improve safety of robots during training and real-world deployment, and provide generalization guarantees on robots' performance and safety in real environments. Additionally, we make the following contributions:
\begin{itemize}
    \item We propose an algorithm for concurrently training the performance policy that optimizes task reward and a backup policy that follows the Safety Bellman Equation \eqref{eq:safety_bellman} in Sec.~\ref{sec:sim_training}. We introduce annealing parameters that allow gradual learning of performance and safety in the \emph{Sim} stage. We also demonstrate that HJ-reachability RL can learn the safety state-action value function end-to-end from images and enable safe exploration with a shielding scheme.
    \item We propose a modification of off-policy actor-critic algorithms that incorporates the policy distribution regularization from PAC-Bayes Control in Sec.~\ref{sec:lab_training}. By constraining the KL divergence between the prior and posterior policy distribution in expectation with batch samples and a weighting coefficient, we optimize the generalization bound in the \emph{Lab} stage efficiently. With a shielding-based policy architecture, we are able to significantly improve the bound compared to previous PAC-Bayes Control works.
    \item We demonstrate the ability of our framework to reduce safety violations during training and improve empirical performance and safety, as well as generalization guarantees, compared to other safe learning techniques and previous work in PAC-Bayes Control in Sec.~\ref{sec:experiments}. We set up ego-vision navigation tasks in two types of environments including one with realistic indoor room layout and visuals. We also validate our approach and generalization guarantees with a quadruped robot navigating in real indoor environments (Sec.~\ref{sec:phy_exp}).
\end{itemize}

\section{Related Work}

\paragraph{Safe Exploration}
Ensuring safety during training has long been a problem in the reinforcement learning community. On one hand, the RL agent usually needs to experience failure in order to learn to be safe. On the other hand, being too conservative hinders exploring the state/action space sufficiently. Constrained MDP (CMDP) is a frequently used framework in safe exploration to satisfy constraints by changing the optimization objective to include some forms of risk \citep{garcia2015survey}. CMDP faces two main challenges: how to incorporate the safety constraints in RL algorithms and how to efficiently solve the constrained optimization problem. \citet{chow2017riskconstrained} use Lagrangian methods to transform the constrained optimization into an unconstrained one over the primal variable (policy) and the dual variable (penalty coefficient). A recent line of works building on reachability analysis argues that optimizing the sum of rewards and penalties is not an accurate encoding of safety \citep{ hsu2021safety}.
\new{Instead, they encode the safety specification of dynamical systems by finding the optimal safety value function, which is a solution to a Hamilton-Jacobi-Bellman/Isaacs variational inequality \citep{bansal2017hamilton, fisac2015reach}. With this safety value function, they apply a least-restrictive control law to shield any performance-oriented policy by overriding with an optimal safe action only when the agent is at states with critically low safety values \cite{fisac2019AGS, leung2020infusing}. \citet{cheng2019e2e} propose a similar shielding framework, utilizing the related control barrier function (CBF) concept. If the system dynamics are control-affine and a CBF is available, a smooth safety override can be computed efficiently by solving a quadratic program with a linear CBF constraint. However, these methods all assume that the dynamics and the environment are at least approximately known. Moreover, they also require that the reachability value function or CBF is available before the learning starts, which is non-trivial for high-dimensional dynamics and/or unknown environments.}

\new{To mitigate the curse of dimensionality and address generalization to novel environments, we build upon reachability RL \citep{hsu2021safety, fisac2019bridging}, which finds an approximate safety value function.}
Recent methods in \citep{srinivasan2020learning, thananjeyan2021recovery, dalal2018safe, chen2021safe} address the safety problem by similar learning-based methods and shielding schemes as proposed in this work. However, the major differences lie in how the safety state-action value function, or \textit{safety critic}, is trained and where the backup actions come from. \citet{dalal2018safe} assume the safety of systems can be ensured by adjusting the action in a single time step (no long-term effect). Thus, they learn a linear safety-signal model and formulate a quadratic program to find the closest control to the reference control such that the safety constraints are satisfied. \citet{srinivasan2020learning} and \citet{thananjeyan2021recovery} learn the safety state-action value function from only sparse and binary safety labels. \citet{srinivasan2020learning} use this function to filter out the \textit{unsafe} actions from the performance policy and resample actions until the backup agent deems the proposed actions safe, while \citet{thananjeyan2021recovery} let the backup agent directly take over. The concurrent work \citep{chen2021safe} uses the same reachability RL to learn the backup agent. Our method is distinct in that (1) we propose the two-stage training to further reduce the safety violations in training and (2) we train the reachability RL end-to-end from images without pre-training the visual encoder. \new{We compare our reachability critic with risk critic \citep{srinivasan2020learning, thananjeyan2021recovery} as detailed in Sec. \ref{sec:hj_risk_comparison}. Our method reduces the number of safety violations by up to $77\%$ in Lab training and $38\%$ in testing.}

\new{A different line of works use learning-based methods to capture the residual error between the nominal model and the real dynamics, which results from model mismatch and/or uncertainties. Then, they combine learned models with model-based RL or model predictive control (MPC) to allow safe exploration. \citet{berkenkamp2016safe} use Gaussian process (GP) to estimate the performance of control parameters and they only deploy parameters that are predicted to be higher than a predefined threshold. On the other hand, \citet{koller2018learning} use GP to estimate the residual error and then utilize this model to over-approximate the forward-reachable set (FRS). They formulate a terminal constraint in MPC to only deploy policies whose FRS reaches a known control-invariant set (under some safety controllers). \citet{liu2020robust} learn the unknown residual with regression and quantify the residual error bound by formulating a covariance shift problem. Our method is distinct in that we use a model-free approach since we only assume we have high-dimensional measurements like RGB images, which cannot be easily handled with model-based methods. Secondly, we do not assume having access to a safe set a priori.}

\paragraph{Generalization Theory and Guarantees} In supervised learning, generalization theory provides a principled guarantee on the true expected loss on new samples drawn from the underlying (but unknown) data distribution, after training a model using a finite number of samples. Foundational frameworks include Vapnik-Chervonenkis (VC) theory \cite{vapnik2015uniform} and Rademacher complexity \cite{bousquet2003introduction}; however the resulting bounds are generally extremely loose for neural networks. 
More recent approaches based on PAC-Bayes generalization theory \cite{mcallester1999some} have provided non-vacuous bounds for neural networks in supervised learning \cite{dziugaite2017computing, perez2021tighter}. Majumdar et al \cite{majumdar2021pac} apply the PAC-Bayes framework in policy learning settings and provide generalization guarantees for control policies in unseen environments. Follow-up work has provided strong guarantees in different robotics settings including for learning neural network policies for vision-based control \cite{ren2020generalization, veer2020probably, gurgen2021learning, agarwal2021stronger, farid2022failure}. Unlike supervised learning settings where picking a PAC-Bayes prior can be difficult, previous work in control settings has encoded different domain knowledge in the prior, such as diverse trajectories from human demonstrations \citep{ren2020generalization}. Our work also encodes diverse navigation strategies into the prior through maximum entropy learning \citep{eysenbach2018diayn}. In addition, previous work has not adopted safety-related policy architectures nor considered safety \emph{during training}. Combining PAC-Bayes theory with reachability safety analysis, we are able to provide stronger guarantees on performance and safety.

\paragraph{Safe Visual Navigation in Unseen Environments} Robot navigation has witnessed a long history of research \cite{bonin2008visual}, and many of the approaches have focused on explicit mapping of the environment combined with long-horizon planning in order to reach a goal location \cite{sim2006autonomous, thrun1996integrating}. Some recent works apply a map-less approach \cite{zhu2017target, bansal2020combining} or builds a map-like belief of the world \cite{gupta2017cognitive} instead. They often take an end-to-end learning approach and start to tackle generalization to previously unseen environments. Similar to them, we train from pixels to actions, and use RGB images as the policy input without any depth information or mapping of the environment. Furthermore, we place more emphasis on the safety of the robot; we aim to train the robot avoiding any collision with obstacles and reaching some target location without the need of explicit mapping (e.g., initial and target locations can be in the same living room). There has been work that explicitly aims to improve safety of the navigating agent. A popular approach is to detect any novel environment or location (often using a neural network) and resort to conservative actions when novelty is detected \cite{richter2017safe, wellhausen2020safe}. A slightly different approach is to estimate the uncertainty of the policy output and act cautiously when the policy is uncertain where to go \cite{lutjens2019safe, kahn2017uncertainty}. However, these work learn the notion of novelty and uncertainty purely from data, often in the form of binary signals, which can be sample inefficient and not generalizable to unseen domains. Closer to our work, there has been a line of work in applying Hamilton-Jacobi reachability analysis in visual navigation. Bajcsy et al \cite{bajcsy2019efficient} solves for the reachability set at each step but relies on a map generated using onboard camera. Li et al \cite{li2020generating} proposes supervising the visual policy using expert data generated by solving a reachability problem. As detailed in the following section, our work also leverages reachability analysis but does not build a map of the environment nor relies on offline data generated by a different (expert) agent.

\new{\paragraph{Adaptive Sim-to-Real Transfer} Directly applying policies trained in simulation to real environments can lead to bad performance and safety, and there has been work that adaptively bridges the Sim-to-Real gap. One line of work addresses the mismatch in robot and environment dynamics by explicitly searching for simulation parameters (e.g., mass, friction coefficient) that result in trajectories matching the real rollouts \citep{ramos2019bayessim, chebotar2019closing, lim2021planar}. \citet{mehta2020active} propose active domain randomization that looks for simulation parameters that leads to different trajectories than reference ones, and those parameters are deemed important to train upon. A different approach \citep{muratore2021data} searches for simulation parameters by directly optimizing task reward in real environments without matching the dynamics. A work closer to ours is Multi-Fidelity RL by \citet{cutler2014multifidelity}, in which lower-fidelity environment (i.e., simulation) determines exploration heuristics for higher-fidelity environment (i.e., real world), and higher-fidelity environment learns model parameters for lower-fidelity environment. In a similar spirit, we learn safety-aware policy in lower-fidelity simulation for safer exploration in the Lab stage, where the policy distribution is fine-tuned. An important distinction of our work from previous ones is that we jointly address the Sim-to-Real gap in robot perception, environment configuration and dynamics. In addition, we provide probabilistic guarantees on the performance and safety of policies being deployed in real environments.}

\section{Problem Formulation and Preliminaries}
\label{sec:formulation}

We consider a robot with discrete-time dynamics given by 
\begin{equation}
    \state_{t+1} = \dyn_\env(\state_t, \ctrl_t),
    \label{eq:dynamics}
\end{equation}
with state $\state \in \xSet \subseteq \reals^{n_\state}$, control input $\ctrl \in \cSet \subseteq \reals^{n_\ctrl}$, and
\new{environment $\env \in \envSpace$ that the robot interacts with (e.g., an indoor space with furniture including initial and goal locations of the robot). Below we introduce the different conditions of the environments considered in the three stages. See Figure.~\ref{fig:sim2lab2real}a for visualization.}
\vspace{1mm}

\noindent \new{
\textbf{Environment - Sim.} In the Sim stage, we assume there is a set of training environments $\envSetPr \subset \envSpace$ (e.g., synthetic indoor spaces with randomized arrangement of furniture), $\envSetPr := \{\env_1, \env_2, \cdots, \env_{N'} \}$. There is no assumption on how $\envSetPr$ is distributed in $\envSpace$. 
}
\vspace{1mm}

\noindent \new{
\textbf{Environment - Lab.} In the Lab stage, we are concerned with more specific conditions, and there can be different distributions of environments $D_1, D_2, ...$  (e.g., office or home spaces, dimensions of the obstacles), with which the policies trained in Sim can be fine-tuned. We assume \emph{no} explicit knowledge of each distribution $\envDist_i$; instead, we assume there is a set of $N_i$ training environments drawn i.i.d. from $\envDist_i$ available for the robot to train in; we denote these training datasets by $\envSetPs_i := \{\env_1, \env_2, \cdots, \env_{N_i} \} \sim \envDist_i^{N_i}$. With a slight abuse of notations and for convenience, we consider a single target condition when introducing the rest of formulation and the approach, and denote the concerned distribution $D$, the training set $\envSetPs$, and the number of training environments $N$.
}
\vspace{1mm}

\noindent \new{
\textbf{Environment - Real.} In the Real stage, we assume the robot is deployed in environments from the same distribution $D$ but \emph{unseen} during the Lab stage.
}
\vspace{1mm}

\new{
Next we introduce the rest of problem settings including the robot sensor, the policy, and robot's task involving the reward function and the failure set. These settings hold the same for all three stages, except for the failure set which we do not require knowledge of at Real deployment.
\vspace{1mm}
}

\noindent \new{\textbf{Sensor.}} In all environments, we assume the robot has a sensor (e.g., RGB camera) that provides an observation $\obs = h_\env(\state)$ using a sensor mapping $h: \xSet \times \envSpace \to \oSet$. 
\vspace{1mm}

\noindent \new{\textbf{Task and Policy.}} Suppose the robot's task can be defined by a reward function, and let $R_\env(\pi)$ denote the cumulative reward gained over a (finite) time horizon by a deterministic policy $\policy: \oSet \to \cSet$ when deployed in an environment $E$. We assume the policy $\pi$ belongs to a space $\Pi$ of policies. We also allow policies that map \emph{histories} of observations to actions by augmenting the observation space to keep track of observation sequences. We assume $R_\env(\pi)\in [0,1]$, but make no further assumptions such as continuity or smoothness. We use $\traj^{\state, \policy}_\env: [0, T] \times \envSpace \to \xSet$ to denote the trajectory rollout from state $\state$ using policy $\pi$ in the environment $\env$ up to a time horizon $T$. 
\vspace{1mm}

\noindent \new{\textbf{Failure set.}} We further assume there are environment-dependent failure sets $\failure_\env \subseteq \xSet$, that the robot is not allowed to enter. In training stages, we assume the robot \new{has access} to Lipschitz functions $\consFunc: \xSet \times \envSpace \to \reals$ such that $\failure_\env$ is equal to the zero superlevel set of $\consFunc_\env$, namely, $\state \in \failure_\env \Leftrightarrow \consFunc_\env(\state) \geq 0$. For example, $\consFunc_\env(\state)$ can be the signed distance function to the nearest obstacle around state $\state$. Thus, $\consFunc_\env(\state)$ is called the safety margin function throughout the paper.
\vspace{1mm}

\subsection{Goal}
Our goal is to 
learn policies that \emph{provably generalize} to unseen environments \new{at the Real stage}. This is very challenging since we do not have explicit knowledge of the underlying distribution $\envDist$. 
We employ a slightly more general formulation where a \emph{distribution} $P$ over policies $\policy \in \policySet$ instead of a single policy is used.    
In addition to maximizing the policy reward, we want to minimize the number of safety violations, i.e., the number of times that the robot enters failure sets. Our goal can then be formalized by the following optimization problem, \new{which we would like to lower bound as the guarantee}:

\vspace{-10pt}
\begin{align}
\centering
\rewardFuncPolicy^\star := \underset{P \in \mathcal{P}}{\sup} \ \rewardFuncPolicy_{\envDist}(P),
\ \text{where} \ & \rewardFuncPolicy_{\envDist}(P) := \new{\underset{E \sim \envDist}{\mathbb{E}} ~ \bigg[
    \underset{\pi \sim P}{\mathbb{E}} \Big[ \rewardFuncPolicy_\env(\pi) \Big]
\bigg]}, \\
& \rewardFuncPolicy_\env(\pi) := \overline{\rewardFuncPolicy}_\env(\policy) \mathbbm{1} \Big\{ \forall t \in [0, T], \traj^{\state, \policy}_\env(t) \notin \failure_\env \Big\},
\label{eq:opt}
\end{align}
where $\overline{\rewardFuncPolicy}_\env(\policy) \in [0,1] $ denotes the task reward that does not penalize safety violation, $\mathcal{P}$ denotes the space of probability distributions on the policy space $\Pi$, and \new{$\mathbbm{1}\{\cdot\}$ is the indicator function. Here the task reward can be either dense (e.g., normalized cumulative reward) or sparse (e.g., reaching the target or not).}

\subsection{Generalization Bounds} Recently, PAC-Bayes generalization bounds have been applied to policy learning settings in order to provide formal generalization guarantees in unseen environments. We briefly introduce this framework here, as it will be used in our overall approach presented in Section \ref{sec:method overview}. \new{First it requires training a prior policy distribution $P_0$, which we do in the Sim stage with the set of environments $M'$. Then in the Lab stage, we fine-tune $P_0$ with environments $M$ to obtain the posterior distribution $P$.} Now, define the \emph{empirical reward} of $P$ as the average expected reward across training environments in $\envSetPs$:
\vspace{-3pt}
\begin{equation}
\label{eq:training_reward}
\rewardFuncPolicy_{\envSetPs}(P):=\frac{1}{N} \sum_{\env \in \envSetPs} \underset{\pi \sim P}{\mathbb{E}} \Big[ \rewardFuncPolicy_\env(\pi) \Big].
\end{equation}
The following theorem can then be used to lower bound the true expected reward $R_\envDist(P)$.

\begin{theorem}[PAC-Bayes Bound for Control Policies; adapted from \citep{majumdar2021pac}] 
\label{thm:pac bayes control}
Let $P_0\in\mathcal{P}$ be a prior distribution. Then, for any $P\in\mathcal{P}$, and any $\delta \in (0,1)$, with probability at least $1 - \delta$ over sampled environments \new{$\envSetPs \sim \envDist^N$}, the following inequality holds:
\begin{equation*}
\rewardFuncPolicy_\envDist(P) \geq \rewardFuncPolicy_{\PAC}(P, P_0) := \rewardFuncPolicy_\mathcal{\envSetPs}(P) - \sqrt{C(P, P_0)}
\label{eq:maurer-pac-bound}, \ \textup{where} \ C(P,P_0) := \frac{\KL(P \| P_0) + \log (\frac{2\sqrt{N}}{\delta})}{2N},
\end{equation*}
\new{and $\KL(P || Q)$ stands for Kullback-Leibler (KL) divergence between probability distribution $P$ and $Q$.}
\end{theorem}

Maximizing the lower bound $\rewardFuncPolicy_{\PAC}$ can be viewed as maximizing the empirical reward $\rewardFuncPolicy_\envSetPs(P)$ along with a regularizer $C$ that prevents overfitting by penalizing the deviation of the posterior $P$ from the prior $P_0$. By fine-tuning $P_0$ to $P$ and maximizing the bound in the Lab stage, we can provide a generalization guarantee for trained policies in 
unseen environments in the Real stage.

\new{
\begin{remark}
    In exchange for assuming almost nothing about the environment distribution $\envDist$ and providing statistical guarantees that hold in arbitrarily \emph{high confidence} ($1-\delta$) instead of only in \emph{expectation} over sampled environments (e.g., conformal prediction \citep{shafer2008tutorial}), the PAC-Bayes framework requires at least a few hundred Lab environments ($N \geq 100$) to achieve reasonably tight generalization bounds. This requires substantial resources for training the policies in the Lab stage. In this work we use simulated environments for Lab training, but we envision that training in real environments is well scalable for industry practitioners with extensive training resources. Please refer to Sec.~\ref{sec:conclusion} for more discussion.
\end{remark}
}

\section{Method Overview}
\label{sec:method overview}

\begin{figure}[!t]
    \centering
    \includegraphics[width=.65\textwidth]{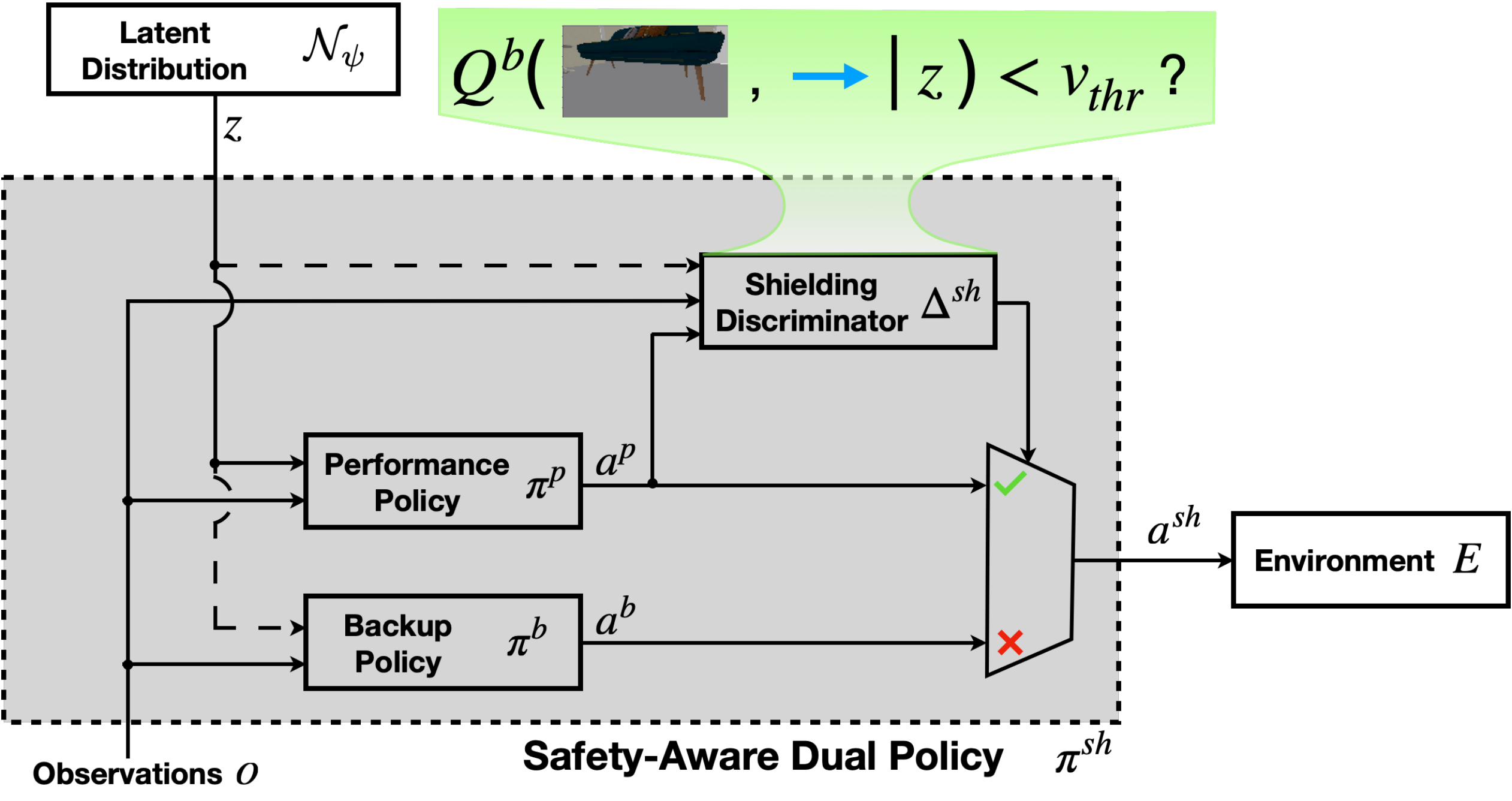}
    \newcaption{\textbf{Architecture of the safety-aware policy distribution}: we consider a dual policy setup where the performance policy $\pi_p$ (and backup policy $\pi_b$, optionally) is conditioned on latent variables sampled from a distribution encoding diverse behavior. The safety state-action value function $Q_b(o_t, a_t)$ from the backup policy is used as the shielding discriminator $\Delta^\shield$, which determines whether the proposed action by the performance policy, $a_p$, is safe. The action from the backup policy, $a_b$, overrides only if necessary.}
    \label{fig:safety_layer}
\end{figure}

Our proposed \textit{Sim-to-Lab-to-Real} framework bridges the reality gap with probabilistic guarantees by learning a safety-aware policy distribution. Fig.~\ref{fig:safety_layer} shows the architecture of the safety-aware dual policy. It explicitly handles safety through the use of a shielding \new{discriminator}, which monitors the candidate actions from the performance policy and overrides them with backup actions only when deemed necessary. We condition the performance policy $\pi_p$ (and the backup policy $\pi_b$ and the shielding \new{discriminator} $\Delta^\shield$) on a latent variable $z$ sampled from some distribution, encoding different trajectories to follow (and different shielding strategies to take).
With these tools, we divide the Sim-to-Real gap into two components, i.e., \textit{Sim-to-Lab} and \textit{Lab-to-Real}, which we \new{tackle} by a two-stage training pipeline as shown in Fig.~\ref{fig:sim2lab2real}. We show how to jointly train a dual policy conditioning on a latent distribution in Sec.~\ref{sec:sim_training}. The details of \emph{Lab} training and derivations of generalization guarantees are provided in Sec.~\ref{sec:lab_training}.

For training, we use a proxy reward function $\rewardFunc_\env: \xSet \times \cSet \times \envSpace \to \reals$, such as dense reward in distance to target, as a single-step surrogate for the task reward $\overline{\rewardFuncPolicy}_\env(\policy)$. Additionally, for every interaction with the environment, the robot receives a safety cost $\consFunc_\env(\state)$ (e.g., distance to nearest obstacle).
We train both \emph{performance} and \emph{backup} policies with modifications of the off-policy Soft Actor-Critic (SAC) algorithm \cite{Haarnoja2018SAC}. We denote the neural network weights of the actor and the critic $\theta$ and $w$. We use superscripts $(\cdot)^\perf$ and $(\cdot)^\backup$ to denote critics, actors, and actions from the performance or backup agent. In order to parameterize the policy distribution, we condition the performance (and the backup) policy on a latent variable $\latent \in \reals^{n_\latent}$. We assume the latent variable is sampled from a multivariate Gaussian distribution with diagonal covariance as $\latent \sim \gaussian(\mu, \Sigma)$, where $\mu \in \reals^{n_\latent}$ is the mean and $\Sigma \in \reals^{n_\latent \times n_\latent}$ is the diagonal covariance matrix. For notational convenience, we denote $\sigma \in \reals^{n_\latent}$ the element-wise square-root of the diagonal of $\Sigma$, and define $\psi = (\mu, \sigma)$, $\gaussian_\psi := \gaussian(\mu, \textup{diag}(\sigma^2))$.
\new{This parameterization enables our framework to quantify the difference between the policy distribution after Sim training and Lab training, by which we can use PAC-Bayes Control to give probabilistic guarantees.}

\section{Pre-Training a Diverse Dual Policy in Simulation}
\label{sec:sim_training}

The goal of the first training stage is to train the dual policy jointly with the fixed latent distribution in simulation, where training is not safety-critical (safety violations are not restricted). In this training stage, we use the environment dataset $\envSetPr$ that contains environments that are not necessarily similar to those from the target environment distribution $\envDist$. Similar to domain randomization techniques, we use environments and conditions with randomized properties, such as random arrangement of furniture in indoor space and random camera tilting angle on the robot. 

In the following subsections, we first review how to learn a backup policy by reachability RL optimizing for the worst-case safety. Then, we propose a shielding scheme with physical meaning to override unsafe candidate actions proposed by the performance policy. Additionally, we incorporate information-theoretic objectives to induce diversity into the learned policy distribution, which helps with fine-tuning the policy distribution and achieve stronger generalization guarantees in the next training stage. Finally, we show how to jointly train two agents, \textit{performance} and \textit{backup}, to realize all the above-mentioned goals. 

\subsection{Safety through Reachability Reinforcement Learning}
Failures are usually catastrophic in safety-critical settings; thus worst-case safety, instead of an average safety over the trajectory, should be considered.
For training the backup policy, we incorporate tools from reachability reinforcement learning \citep{fisac2019bridging, hsu2021safety} and optimize the discounted safety Bellman equation (DSBE) as below,
\begin{equation}
    Q^\backup(\obs_\tdisc, \ctrl_\tdisc) := (1-\gamma) \consFunc_\env(\state_\tdisc) + \gamma \max \Big\{\consFunc_\env(\state_\tdisc), \ \min_{\ctrl_{\tdisc+1} \in \cSet} Q^\backup \big( \obs_{\tdisc+1}, \ctrl_{\tdisc+1} \big) \Big\},
    \label{eq:safety_bellman}
\end{equation}
where $\obs_\tdisc = h_\env(\state_\tdisc)$ and $\gamma$ is the discount factor. This discount factor represents how much attention the RL agent places on future outcomes: if $\gamma$ is small, the RL agent only cares about ``imminent danger'', and as $\gamma \rightarrow 1$, one recovers the infinite-horizon safety state-action value function. In the training, we initialize $\gamma=0.8$ and gradually anneal $\gamma$ towards $1$ during the process.

The safety state-action value function in \eqref{eq:safety_bellman} captures the maximum cost $\consFunc_\env$ along the trajectory starting from $\state_\tdisc$ with action $\ctrl_\tdisc$ assuming that the safest control input is applied at every instant thereafter. Thus, $\min_{\ctrl_\tdisc \in \cSet} Q(\obs_\tdisc, \ctrl_\tdisc) > 0$ indicates that the robot is predicted to \emph{inevitably} violate safety in the future if $\ctrl_\tdisc$ is taken. By utilizing this (annealed) DSBE, we have an exact encoding of the property we want our system to satisfy. The DSBE allows the backup agent to learn the safety state-action value function from near-failure executions, which significantly reduces failure events during training. Additionally, the DSBE enables the backup agent to update using a dense learning signal, which is suitable for the joint training of performance and backup agents. To our knowledge, this work demonstrates the first instance of reachability RL in fully end-to-end training with extremely high-dimensional inputs (RGB images), without the need for pre-training a vision encoder as in \cite{chen2021safe}.

\subsection{Shielding}
\begin{figure}[!t]
    \centering
    \includegraphics[width=0.9\textwidth]{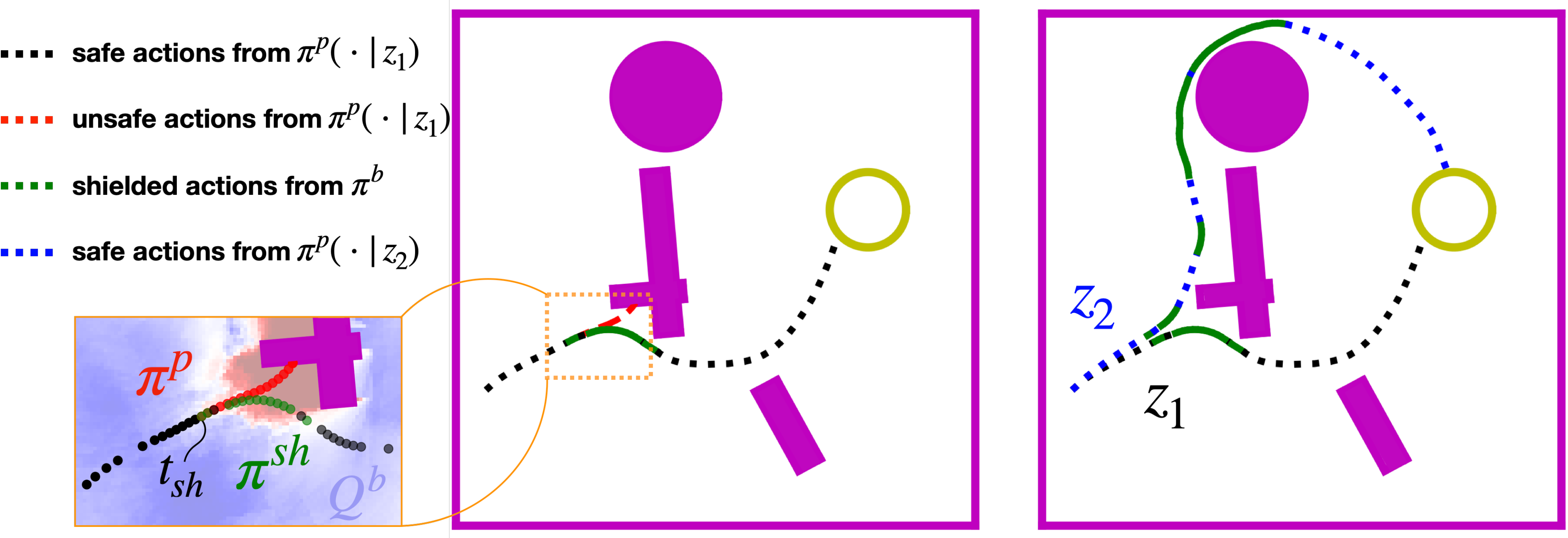}
    \caption{\textbf{Rollout trajectories of the safety-aware policy distribution:} the latent variables sampled from the distribution induce a diverse exploration motives and value-based shielding manages to override the unsafe actions. Red dashed line shows the unshielded actions; Black/Blue dotted lines show the safe actions by the performance policy; Green lines show the backup actions overriding unsafe actions. The inset shows safety values $Q(\obs, \policy^\backup(\obs))$ with the observation $\obs$ taken when the heading angle fixed to the one at time instant $t_{sh}$.}
    \label{fig:diverse_shielding_toy}
\end{figure}

We leverage a least-restrictive control law, i.e., \textit{shielding}, to reduce the number of safety violations in both training and deployment. Suppose we have two policies: performance-pursuing policy $\policy^\perf$ and safety-pursuing (backup) policy $\policy^\backup$. Before we apply a candidate action from the performance-pursuing policy, we use a shielding \new{discriminator} $\Delta^\shield$ to check if it is safe. We replace the proposed action with the action from the backup policy if and only if that candidate action is deemed to result in safety violations in the future. The shielding criterion is summarized in \eqref{eq:shielding}. This ensures minimum intervention by the backup policy while the performance policy guides the robot towards the target \cite{fisac2019AGS, alshiekh2017safe}.

\begin{equation}
    \policy^\shield(\obs) = \left\{ 
        \begin{array}{ll}
            \policy^\perf(\obs), & \Delta^\shield(\obs, \policy^\perf, \qFunc^\backup) = 1\\ 
            \policy^\backup(\obs), & \text{otherwise}
        \end{array}
    \right..
    \label{eq:shielding}
\end{equation}

The safety value function learned by DSBE represents the maximum cost along the trajectory in the future if following the learned policy. If we define the safety margin function $\consFunc_\env(\state)$ to be the closest distance to the obstacles, then $Q^\backup(\obs, \ctrl)$ represents the closest distance of the robot to the obstacles in the future. Based on this, we propose a \textit{value-based shielding} with the threshold having a physical interpretation, i.e., a margin from the failure. Once the robot receives the current observation $\obs$ and uses performance policy to generate action $\policy^\perf(\obs)$, the backup policy overrides the action if and only if $Q^\backup(\obs, \ctrl^\perf) > v_{thr}$.
\new{In other words, the shielding discriminator is defined as below
\begin{equation}
    \Delta^\shield(\obs, \policy^\perf, \qFunc^\backup) := \mathbbm{1} \big\{ \qFunc^\backup(\obs, \policy^\perf(\obs)) \leq v_{thr} \big\}. \label{eq:shielding_criterion}
\end{equation}}
Fig.~\ref{fig:diverse_shielding_toy} shows an example of shielding that prevents applying unsafe actions from the performance policy (replace the red dotted lines with green dotted lines in the inset). We compare the safety state-action value function based on DSBE with ones by sparse safety indicators \citep{srinivasan2020learning, thananjeyan2021recovery} in Sec.~\ref{sec:hj_risk_comparison} and Fig.~\ref{fig:backup_critic}; our approach affords much better safety during training and deployment.

\subsection{Diversity through Maximization of Latent-based Mutual Information}
During Sim training, we also maximize the diversity of robot behavior encoded by the latent distribution, which has shown in different work \cite{ren2020generalization, eysenbach2018diayn} to result in better performance after fine-tuning the distribution, which we perform in the Lab stage. Each latent variable is sampled from the distribution. As the policy is conditioned on the latent variable, it should lead to different trajectories around obstacles and towards the target (Fig.~\ref{fig:diverse_shielding_toy}). With a single policy instead of a distribution, it is prone to overfit to some set of environments and fails to adapt in new environments (Fig.~\ref{fig:traj_diversity}).

In order to distinguish the resulting trajectories using latent variables, we maximize mutual information between observations of trajectories $\trajObs$ and latent variables $\latent$, which can be lowered bounded by sum of mutual information between each observation and the latent variable $I(\trajObs;\latent) \geq \sum_{t=1}^T I(\obs_\tdisc;\latent)$ \citep{jabri2019unsupervised} (\textbf{observation-marginal MI}). We can further lower bound $I(O; Z) \geq \mathbb{E}_{\latent \sim \gaussian_{\psi_0}, \obs \sim p(\cdot | \policy^\shield, \latent)} [\log q_{\phi}(\latent|\obs)]-\mathbb{E}_{\latent \sim \gaussian_{\psi_0}} [\log p(\latent)]$, where the posterior $p(\latent|\obs)$ is approximated by a learned discriminator $q_\phi(\latent|\obs)$, parameterized by a neural network with weights $\phi$ \citep{eysenbach2018diayn}. Intuitively, in order to make trajectories recognizable by the discriminator, the trajectories need to be diverse. Similar to \cite{eysenbach2018diayn}, before updating the policies after sampling a batch of experiences, we augment the proxy reward by a weighted mutual information reward with coefficient $\beta$:
\begin{equation}
    \rewardFunc_\text{aug}(\state_\tdisc, \ctrl_\tdisc, \obs_\tdisc, \latent) = \rewardFunc(\state_\tdisc, \ctrl_\tdisc) + \beta \big[ \log q_\phi(\latent|\obs_\tdisc) - \log p(\latent) \big].
    \label{eq:diverse_rwd}
\end{equation}
This encourages the value function to assign higher reward to regions more recognizable by the discriminator. Concurrently, we train the discriminator by maximizing $\log q_\phi(\latent|\obs)$ with Stochastic Gradient Descent (SGD),
\begin{equation}
    \phi \leftarrow \phi + \nabla_\phi \expect_{\obs} \Big[ \log q_\phi(\latent|\obs) \Big].
    \label{eq:disc_update}
\end{equation}
As shown in Fig.~\ref{fig:safety_layer}, we can additionally condition backup policies with the latent distribution; the robot may avoid obstacles in different directions, and such skills might be beneficial when there is a distributional shift of obstacle placement and geometry in the Lab stage.

The backup agent can also depend on a latent variable. Since the backup agent can intervene at any state and condition on any latent, we instead optimize the conditional mutual information between action and latent given the current observation $I(A; Z|O)$ (\textbf{observation-conditional MI}). We modify the backup policy training objective \eqref{eq:safety_bellman} as below
\begin{equation}
    \theta^{\backup^*} = \arg \min_\theta L(\theta) := \expect_{\obs, \latent} \Big[ \expect_{\ctrl \sim \policy_\theta(\cdot|\obs, \latent)} Q(\obs, \ctrl; \latent) \Big] - \nu I(A; Z|O),
\end{equation}
where the Q-function is now conditioned on a latent variable and $\nu$ is the coefficient balancing the safety cost and the diversity. Through derivations in \ref{app:diverse_backup}, we modify the SAC formulation and the backup actor is updated as,
\begin{equation}
    \theta \leftarrow \theta - \nabla_\theta  ~ \expect_{(\obs, \latent) \sim \mathcal{B}, \ctrl \sim \policy_\theta(\cdot|\obs, \latent)} \bigg[ Q(\obs, \ctrl; \latent) - \nu \log \policy_\theta(\ctrl | \obs, \latent) + \nu \log \frac{1}{n_s} \sum_{i=1, \latent_i \sim p(\latent)}^{n_s} \policy_\theta(\ctrl | \obs, \latent_i) \bigg],
\label{eq:dads-update}
\end{equation}
where $\mathcal{B}$ is the replay buffer. \new{Intuitively, for specific action $\ctrl$ given current observation $\obs$, we want it to have high probability for policy conditioned on a specific latent variable $\latent$ and low probability for other latent variables $\{\latent_i\}$ sampled from the distribution. Note that when the backup agent is also conditioned on latent variable $\latent$, the shielding discriminator in \eqref{eq:shielding_criterion} now becomes $\qFunc^\backup(\obs, \policy^\perf(\obs), \latent) \leq v_{thr}$.} While similar formulations have been explored in previous work \citep{kumar2020smerl, eysenbach2018diayn, sharma2020dynamicsaware} to achieve diverse trajectories/skills in RL, to our best knowledge, we are the first to consider a continuous distribution of latent variables instead of a discrete one. We find this brings difficulty in training, exacerbated by using robot observations instead of true states (e.g., ground-truth locations of the robot); nonetheless, we show effectiveness of such diversity-induced training in Sec.~\ref{sec:sensitivity}.

\subsection{Joint Training of Performance and Backup Policies.}
Now we are ready to perform joint training of the dual policy. In the Sim stage, we fix the latent distribution to be a zero-mean Gaussian distribution with diagonal covariance $\mathcal{N}_{\psi_0}$, where $\psi_0 = (0, \sigma_0)$. For each episode during training, we sample a latent variable $\latent \sim \mathcal{N}_{\psi_0}$ and condition the performance policy (and the backup policy) on it for the whole episode. The training procedure is illustrated in Algorithm~\ref{algo:sim}.

Since we train both policies with modifications of the off-policy SAC algorithm, we can use transitions from actions proposed by either backup policy or performance policy. The transitions from both policies are stored in a shared replay buffer and are sampled at random to update the parameters of actors and critics for both performance and backup agents. At every step during training, the robot needs to select a policy to follow. We introduce a parameter $\backupProb$, the probability that the robot chooses an action proposed by the backup policy. We initialize $\rho$ to 1, meaning that at the beginning, all actions are sampled from the backup policy. Our intuition is that the backup policy needs to be trained well before shielding mechanism is introduced in the training. We gradually anneal $\rho$ to $0$.
Additionally, to realize a safe Sim-to-Lab transfer, we want the performance agent to be aware of the backup agent. Thus, we also apply shielding during Sim training. However, since the backup actor and critic may not be able to shield successfully in the beginning, we introduce a parameter $\shieldProb$, which is the probability that the shielding is activated at this time step. This parameter can be viewed as how much we trust the backup policy and to what extent we want it to shield the performance policy. We typically initialize $\shieldProb$ to $0$ and anneal it to $1$ gradually.
The influence of $\backupProb$ and $\shieldProb$ are further analyzed in Sec.~\ref{sec:sensitivity}.

\begin{algorithm}[!t]
\caption{Joint training in simulator}
\begin{algorithmic}[1]
    \Require $\envSetPr, \policy^\perf, \policy^\backup, q_\phi, \mathcal{N}_{\psi_0} := \gaussian (0,\sigma I), \rho=1, \epsilon=0, \gamma=\gamma_\text{init}$
    \State Sample $E \sim \envSetPr$ and $\latent \sim \mathcal{N}_{\psi_0}$, reset environment \Comment{Same latent for whole episode}
	\For{$t \leftarrow 1$ to \emph{num\textunderscore prior\textunderscore step}}
        \State With probability $\rho$, sample action $\ctrl_\tdisc \sim \policy^\backup(\cdot | \obs_\tdisc, \latent)$; else sample $\ctrl_\tdisc \sim \policy^\perf(\cdot | \obs_\tdisc, \latent)$
        \State With probability $\epsilon$, apply shielding $\ctrl^{\shield}_\tdisc = \policy^\shield(\policy^\backup, \qFunc^\backup, \obs_\tdisc, \ctrl_\tdisc, \latent)$
        \State Step environment $\rewardFunc_\tdisc, \obs_{t}, \state_{\tdisc+1} = \dyn_\env (\state_\tdisc, \ctrl^{\shield}_\tdisc)$
        \State Save $(\obs_{\tdisc+1}, \obs_\tdisc, \ctrl_\tdisc, a^\shield_\tdisc, \latent, \rewardFunc_\tdisc)$ to replay buffer
        \State Update $\policy^\perf, \policy^\backup, q_\phi$ \Comment{See Algorithm \ref{algo:update}}
        \State Anneal $\rho \rightarrow 0, \epsilon \rightarrow 1, \gamma \rightarrow 1$
        \If{timeout or failure}
        \State Sample $E \sim \envSetPr$ and $\latent \sim \mathcal{N}_{\psi_0}$, reset environment
        \EndIf
	\EndFor \\
	\Return $\policy^\perf, \policy^\backup, \gaussian_{\psi_0}$
	\end{algorithmic}
\normalsize
\label{algo:sim}
\end{algorithm}

\begin{algorithm}[!t]
\caption{Updating the performance policy, backup policy, and the discriminator}
\begin{algorithmic}[1] 
	\For{$t \leftarrow 1$ to \emph{num\textunderscore policy\textunderscore update}}
	    \State Sample batch $\{(\obs_{\tdisc+1}, \obs_\tdisc, \ctrl_\tdisc, \latent, \rewardFunc_\tdisc)\}$ from the replay buffer \Comment{Action re-labeling}
	    \State Augment $\rewardFunc_\tdisc$ with mutual information reward \eqref{eq:diverse_rwd}
        \State Update $\policy^\perf$ to maximize $r_\text{aug}$ with SAC \Comment{Observation-marginal MI}
	    \State Sample batch $\{(\obs_{\tdisc+1}, \obs_\tdisc, \ctrl_\tdisc^{\shield}, \latent, \rewardFunc_\tdisc)\}$ from the replay buffer
        \State Update $\policy^\backup$ to minimize $\consFunc_\env(s)$ with modified SAC by \eqref{eq:safety_bellman} and \eqref{eq:dads-update} \Comment{Observation-conditional MI}
    \EndFor
	\For{$t \leftarrow 1$ to \emph{num\textunderscore discrminator\textunderscore update}}
	    \State Sample batch $\{(\obs_\tdisc, \latent)\}$ from the replay buffer
	    \State Update $q_\phi$ with SGD \eqref{eq:disc_update} \new{\Comment{Observation-marginal MI}}
    \EndFor
	\end{algorithmic}
\normalsize
\label{algo:update}
\end{algorithm}

The details of updating the policies and the discriminator are shown in the Algorithm~\ref{algo:update}. Notice that while we train the backup policy $\policy^\backup$ using the executed action $\ctrl_\tdisc^\shield$, the performance policy $\policy^\perf$ is trained using the originally proposed action $\ctrl_\tdisc$ (``action re-labeling''). This ensures that the performance agent learns to associate its proposed action with the transition outcome, and avoids keeping proposing unsafe actions.

After the joint training, we obtain the trained dual policies $\policy^\perf$ and $\policy^\backup$, and the latent distribution $\mathcal{N}_{\psi_0}$ that encodes diverse solutions in the environments. We now fix the weights of the two policies, and consider the latent variable $\latent$ also part of their parameterization. This gives rise to the space of policies $\Pi := \{\policy^\perf_\latent, \policy^b_z : \mathcal{O} \mapsto \mathcal{A} \ | \ \latent \in \reals^{n_z} \}$; hence, the latent distribution $\mathcal{N}_{\psi_0}$ can be equivalently viewed as a distribution on the space $\Pi$ of policies. In the next section, we will consider $\mathcal{N}_{\psi_0}$ as a \emph{prior} distribution $P_0$ on $\policy$ and ``fine-tune'' it by searching for a \emph{posterior} distribution $P = \mathcal{N}_\psi$, which comes with the generalization guarantee from PAC-Bayes Control. 
\section{Safely Fine-Tuning Policies in Lab}
\label{sec:lab_training}

In the second training stage, we consider more safety-critical training environments such as test tracks for autonomous cars or indoor lab space, where the conditions can be more realistic and closer to real environments. After pre-training the performance and backup policies with shielding, the robot can safely explore and fine-tune the prior policy distribution $P_0$ in a new set of environments $\envSetPs$ sampled from the unknown distribution $\envDist$. Leveraging the PAC-Bayes Control framework, we can provide ``certificates'' of generalization for the resulting posterior policy distribution $P$.

The PAC-Bayes generalization bound $R_\text{PAC}$ associated with $P$ from Eq.~\eqref{eq:maurer-pac-bound} consists of two parts: (1) $R_\envSetPs(P)$, the empirical reward of $P$ as the average expected reward across training environments in $\envSetPs$ \eqref{eq:training_reward}, which can be optimized using SAC algorithm; (2) a regularizer $C(P,P_0)$ that penalizes the posterior $P$ for deviating significantly from the prior $P_0$,
\begin{equation}
    C(P, P_0) := \frac{\KL(P \| P_0) + \log (\frac{2\sqrt{N}}{\delta})}{2N}.
\end{equation}

Note that the only term in $C(P,P_0)$ that involves $P$ is the KL divergence term between $P$ and $P_0$. To minimize $C(P,P_0)$, we modify the SAC objective to include minimization of the KL divergence term. Also, we consider stochasticity of the policy from the latent distribution instead of the policy network; this leads to removing the policy entropy regularization in SAC and adding a weighted KL divergence term to the actor loss:
\begin{equation}
    \max_P \expect_{\obs, \latent} \Big[ \expect_{\ctrl \sim \pi_\theta(\cdot|\obs, \latent)} \big[ Q^p (o, \ctrl) \big] \Big] - \alpha \text{KL}(P, P_0) ,
\end{equation}
where $\alpha \in \reals$ is a weighting coefficient to be tuned. In practice, we find the gradient of the KL divergence term heavily dominates the noisy gradient of actor and critic, and thus we approximate the KL divergence with an expectation on the posterior:
\begin{equation}
    \max_P \expect_{\obs, \latent} \Big[ \expect_{\ctrl \sim \pi_\theta(\cdot|\obs, \latent)} \big[ Q^p (o, \ctrl) \big] - \alpha \log \frac{P(z)}{P_0(z)} \Big].
\label{eq:ps-update}
\end{equation}

Below we show the algorithm for this stage of training. To avoid safety violations, we always apply value-based shielding to the proposed action, and continue to apply action-relabeling when updating $P$.

\begin{algorithm}[!ht]
\caption{Safely fine-tuning the policy distribution} 
    \begin{algorithmic}[1]
    \Require $\envSetPs, \pi^p, \pi^b, P = P_0$
    \State Sample $E \sim \envSetPs$ and $\latent \sim P$, reset environment
	\For{$t \leftarrow 1$ to \emph{num\textunderscore posterior\textunderscore step}}
        \State Sample $a_t \sim \pi^p(\cdot | o_t, z)$
        \State Apply value-based shielding $a^{\textup{sh}}_t = \pi^\textup{sh}(\pi^b, \qFunc^\backup, o_t, a_t, z)$
        \State Step environment $r_t, o_{t}, s_{t+1} \sim \mathcal{P}(\cdot | s_t, a^{\textup{sh}}_t)$
        \State Save $(o_{t+1}, a_t, z, r_t)$ to replay buffer \Comment{Action re-labeling}
        \State Update $P$ using SAC with weighted regularization \eqref{eq:ps-update}
        \If{timeout or failure}
        \State Sample $E \sim \envSetPs$, $z \sim P$, reset environment
        \EndIf
	\EndFor \\
	\Return $P$
	\end{algorithmic}
\normalsize
\label{algo:lab}
\end{algorithm}

\subsection{Computing the Generalization Bound.}
After training, we can calculate the generalization bound using the optimized posterior $P$. First, note that the empirical reward $R_\envSetPs(P)$ involves an expectation over the posterior and thus cannot be computed in closed form. Instead, it can be estimated by sampling a large number of policies $z_1,...,z_L$ from $P$: $\hat{R}_\envSetPs(P):= \frac{1}{NL}\sum_{E \in \envSetPs}\sum_{i=1}^L R_\env (\pi^{p,b}_{z_i})$, and the error due to finite sampling can be bounded using a sample convergence bound $\overline{R}_\envSetPs$ \cite{langford2002not}. The final bound $R_\text{bound}(P) \leq R_\envDist(P)$ is obtained from $\overline{R}_\envSetPs$ and $C(P, P_0)$ by a slight tightening of $C_\text{PAC}$ from Theorem \ref{thm:pac bayes control} using the KL-inverse function \cite{majumdar2021pac}. Please refer to Appendix A2 in \cite{ren2020generalization} for detailed derivations. Overall, our approach provides generalization guarantees in novel environments from the distribution $\envDist$: as policies are randomly sampled from the posterior $P$ and applied in test environments, the expected success rate over all test environments is guaranteed to be at least $R_\text{bound}(P)$ (with probability $1-\delta$ over the sampling of training environments; $\delta=0.01$ for all experiments in Sec.~\ref{sec:experiments}). Through reachability shielding during training and generalization guarantees for the resulting policies, we bridge the \emph{Lab-to-Real} gap with a probabilistically guaranteed safety-aware policy distribution.

\section{Experiments}
\label{sec:experiments}

Through extensive simulation and hardware experiments, we aim to answer the following questions: does our proposed Sim-to-Lab-to-Real achieve (1) lower safety violations during Lab training compared to other safe learning methods, (2) stronger generalization guarantees on performance and safety compared to previous work in PAC-Bayes Control, and (3) better empirical performance and safety during deployment compared to all baselines? We also evaluate (a) the relative importance of the Sim stage and Lab stage, (b) how the value threshold in shielding affects safety and efficiency, (c) how the regularization weight in \eqref{eq:ps-update} affects generalization guarantees and empirical performance after training, (d) how the two annealing parameters during Sim training, $\epsilon$ and $\rho$, affect training performance, and (e) how diversity components, mutual information maximization during Sim training and latent dimension, affect performance after Lab training.

\subsection{Experiment Setup}

\begin{figure}[!ht]
    \centering
    \includegraphics[width=\textwidth]{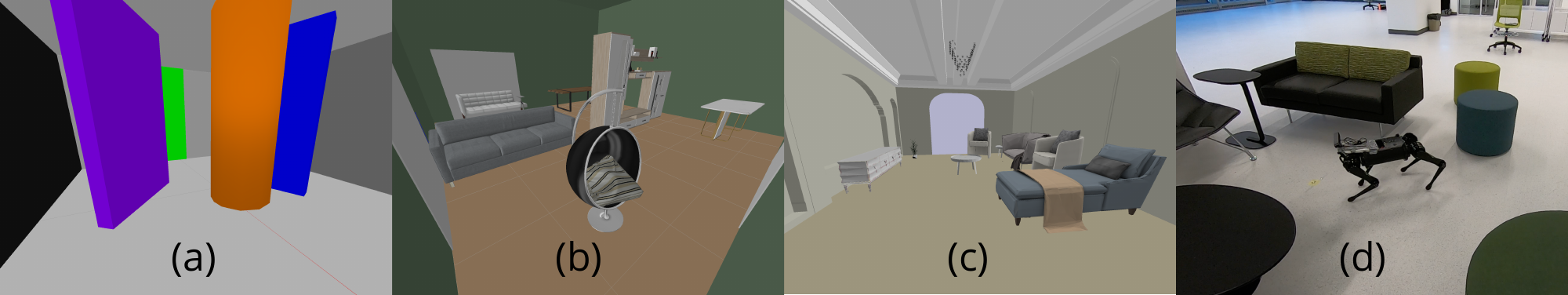}
    \caption{\textbf{Samples of environments used in experiments:} (a) Sim training in Vanilla-Env; (b) Sim training in Advanced-Env; (c) Advanced-Realistic training; (d) physical robot deployment.}
    \label{fig:environment}
\end{figure}

\subsubsection{Environments}
We evaluate the proposed methods by performing ego-vision navigation task in two types of environment. The first type (\textbf{Vanilla-Env}) consists of undecorated rooms of $2m \times 2m$ with randomly placed cylindrical and rectangular obstacles of different dimensions and poses, and the robot needs to bypass them and the walls to reach a green door (a smaller circular region in front it) (Fig.~\ref{fig:environment}a). A virtual camera is simulated with 120 degree field of view both vertically and horizontally, outputting RGB images of $48 \times 48$ pixels. We treat the robot as a point mass when checking collision. 

The second type of environments (\textbf{Advanced-Env}) uses realistic furniture models from the 3D-FRONT dataset \cite{fu20213dfront} (Fig.~\ref{fig:environment}b); the robot needs to safely reach some target location (a smaller circular region) using given distance and heading angle towards the target. A virtual camera is simulated at the front of the robot with 72 degree field of view vertically and 128 degree horizontally (matching the ZED 2 camera used in Real deployment), outputting RGB images of $90 \times 160$ pixels. When checking collisions, we approximate the robot as a circular shape of radius 25cm, roughly the same as the quadrupedal robot in Real deployment. 

For both types of environments, the control loop runs at 10Hz and the maximum number of steps is 200. The robot is commanded with forward velocity ($[0.5, 1]$ m/s for performance policy and $[0.2,0.5]$ m/s for backup policy) and angular velocity ($[-1,1]$ rad/s for both policies). We use dense proxy reward that is proportional to the percentage of distance traveled between initial location and goal, and the safety signal is calculated as the minimum distance to obstacles and walls. Additionally, we assume the robot is given $\ell_\env(\state)$ and $\Delta_\env(\state)$, distance and relative bearing to the goal.

For Sim training, we randomize obstacle and furniture configurations to cover possible scenarios as much as possible. We also randomize camera poses (tilt and roll angles) in Advanced-Env to account for possible noise in real experiments. Sim training uses 100 environments in Vanilla-Env and 500 environments in Advanced-Env. After Sim training, we can fine-tune the policies in different types of Lab environments. For Vanilla-Env, we consider:
\begin{itemize}
    \item \textbf{Vanilla-Normal}: shares the same environment parameters as ones in the Sim stage.
    \item \textbf{Vanilla-Dynamics}: increases the lower bound of forward and angular velocity (more aggressive maneuvers).
    \item \textbf{Vanilla-Task}: adds an additional condition on success that the the robot needs to enter the target region with a yaw angle within a small range instead of $2 \pi$ (no restriction) in the Sim stage. The robot may pass through the target region and re-enter it with the required yaw orientation. The robot knows the lower bound and the upper bound of the required yaw range.
\end{itemize}
and for Advanced-Env, we consider:
\begin{itemize}
    \item \textbf{Advanced-Dense}: assigns a higher density of furniture in the rooms, resulting in smaller clearances between them.
    \item \textbf{Advanced-Realistic}: uses realistic room layouts (Fig.~\ref{fig:environment}c) and associated furniture configurations from the 3D-FRONT dataset, which are similar to real environments. We perform Lab-to-Real transfer with policies trained in this Lab (Fig.~\ref{fig:environment}d). More details about the dataset and room layouts can be found in \ref{app:env-setup}.
\end{itemize}

\subsubsection{Policy}
We parameterize the performance and backup agents with neural networks consisting of convolutional layers and then fully connected layers. The actor and critic of each agent share the same convolutional layers. In Vanilla-Env, a single RGB image is fed to the convolution layers, and the latent variable is appended to the output of the last convolutional layer before fully connected layers. In Advanced-Env, we stack 4 previous RGB images while skipping 3 frames between two images to encode the past trajectory of the robot. Then, the stacked images are concatenated with the first $10$ dimensions of the latent variable by repeating each dimension to the image size. Rest of the dimensions is appended to the output of the last convolutional layer. In addition to the image observation, the actors and critics also receive two auxiliary signals $\ell_\env(\state)$ and $\Delta_\env(\state)$, which are also appended to the output of the last convolutional layer. The details of neural network architecture and training can be found in \ref{app:hyperparams}. 

\subsubsection{Baselines}

We compare our methods to five prior RL algorithms that neglect safety violations (Base and PAC{\myspace}Base \cite{majumdar2021pac}) or address safety by reward shaping (RP and PAC{\myspace}RP) or use a separate safety agent (SQRL \cite{srinivasan2020learning} and Recovery RL \cite{thananjeyan2021recovery}).
Sim-to-Lab-to-Real varies from SQRL and Recovery RL in that the latter trains the safety critic by the sparse safety indicators as below,
\begin{equation*}
        Q^\backup(\obs_t, \ctrl_t) := \mathcal{I}_\env(\state_t) + \gamma \big( 1-\mathcal{I}_\env(\state_t) \big)  \min_{\ctrl_{t+1} \in \cSet} Q^\backup \big( \obs_{t+1}, \ctrl_{t+1} \big),
    \label{eq:risk_bellman}
\end{equation*}
where $\mathcal{I}_\env(\state_t) = \mathbbm{1}\{ \consFunc_\env(\state_t) > 0\}$ is the indicator function of the safety violations.
In Sim-to-Lab-to-Real, the safety state-action values represent the robot's closest \textit{distance} to the obstacles in the future, while in Recovery RL and SQRL, the values represent the \textit{probability} that the robot will hit the obstacle (but the probability strongly depends on the discount factor used). The major distinction between Sim-to-Lab-to-Real and PAC-Bayes control is that the latter does not handle the safety explicitly but instead hopes to use diverse policies and fine-tuning to prevent unsafe maneuver. We give a brief description of these methods below and summarize the similarities and differences in Table~\ref{tab:method_comp}. 
\begin{itemize}
    \item \textbf{Sim-to-Lab-to-Real (ours)}: trains a distribution over dual policies conditioned on latent variables with guarantees on generalization to novel environments. We present two variants: \textbf{PAC{\myspace}Shield{\myspace}Perf}, whose performance policy is conditioned on latent variables, and \textbf{PAC{\myspace}Shield{\myspace}Both}, whose both performance and backup policies are conditioned on latent variables (Fig.~\ref{fig:safety_layer}).
    \item \textbf{Shield (ours)}: trains a dual policy without conditioning on latent variables, thus no distribution over policies \new{nor generalization guarantees}.
    \item \textbf{PAC-Bayes Control \cite{majumdar2021pac}:} trains a distribution over policies conditioned on latent variables that optimizes for either only task reward (PAC{\myspace}Base) or reward with penalty (PAC{\myspace}RP).
    \item \textbf{Base:} trains a single policy that optimizes the task reward only.
    \item \textbf{Reward Penalty (RP):} trains a single policy but augments the task reward with penalty on safety violations, $\hat{\rewardFunc}_\env(\state, \ctrl) = \rewardFunc_\env(\state, \ctrl) - \lambda \mathbbm{1}\{ \consFunc_\env(\state) > 0\}$.
    \item \textbf{Safety Critic for RL (SQRL) \citep{srinivasan2020learning}}: trains a dual policy. The backup critic optimizes the Lagrange relaxation of CMDP, $\hat{J}(\policy) = J(\policy) + \nu \expect_{\ctrl \sim \policy} [ (v_{thr} - Q^\backup(\obs, \ctrl) ]$, with a rejection sampling method that re-samples action if $Q^\backup(\obs, \ctrl) > v_{thr}$.
    \item \textbf{Recovery RL \citep{thananjeyan2021recovery}:} trains a dual policy. The backup critic is trained in the same method as SQRL, but the backup action is from the backup actor instead of being re-sampled from performance policy.
\end{itemize}

\begin{table}[!ht]
    \centering
    \small
    \newcaption{Major distinctions among Sim-to-Lab-to-Real and baseline methods.}
    \begin{tabular}{C{0.24\linewidth}C{0.12\linewidth}C{0.3\linewidth}C{0.18\linewidth}}
        \textbf{Methods} & \textbf{Dual Policy} & \textbf{Safety Treatment} & \textbf{\shortstack{Generalization\\Guarantees}}\\
        \mytoprule{1pt}
        Sim-to-Lab-to-Real (ours) & \cmark & \cmark (Reachability safety critic) & \cmark \\
        Shield (ours) & \cmark & \cmark (Reachability safety critic) & \xmark \\
        PAC+Base \cite{majumdar2021pac} & \xmark & \xmark & \cmark \\
        PAC+Reward Penalty \cite{majumdar2021pac} & \xmark & \cmark (Reward with safety penalty) & \cmark \\
        Base & \xmark & \xmark & \xmark \\
        Reward Penalty & \xmark & \cmark (Reward with safety penalty) & \xmark \\
        Safety Critic for RL \citep{srinivasan2020learning} & \xmark & \cmark (Risk safety critic) & \xmark \\
        Recovery RL \citep{thananjeyan2021recovery} & \cmark & \cmark (Risk safety critic) & \xmark \\
    \end{tabular}
    \label{tab:method_comp}
\end{table}

\subsection{Results}
\begin{figure}[!t]
    \centering
    \begin{subfigure}[b]{.475\textwidth}
        \centering
        \includegraphics[width=\textwidth]{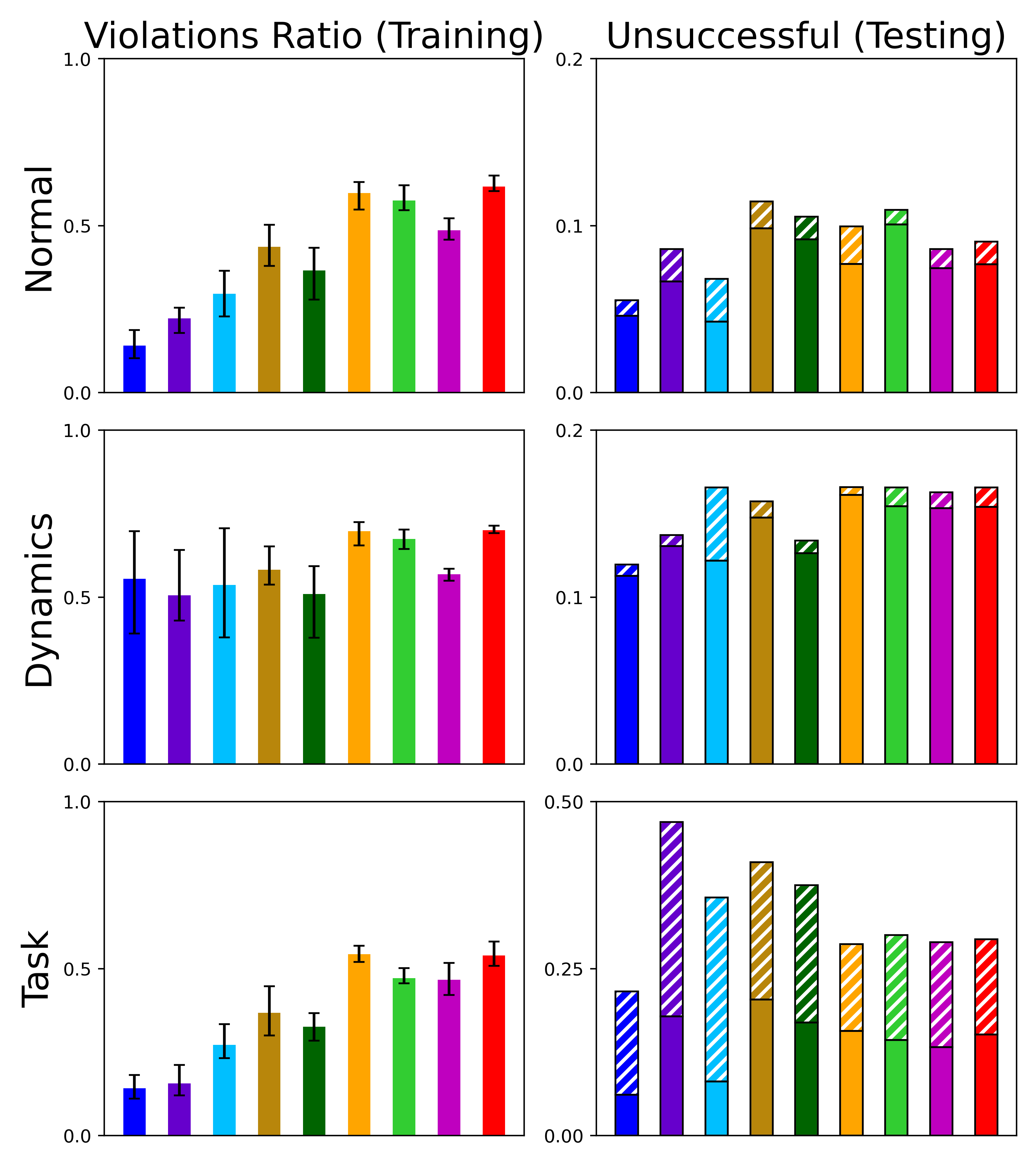}
        \caption{Vanilla-Env (averaged over 5 seeds)}
        \label{fig:main_toy}
    \end{subfigure}
    \begin{subfigure}[b]{.475\textwidth}
        \centering
        \includegraphics[width=\textwidth]{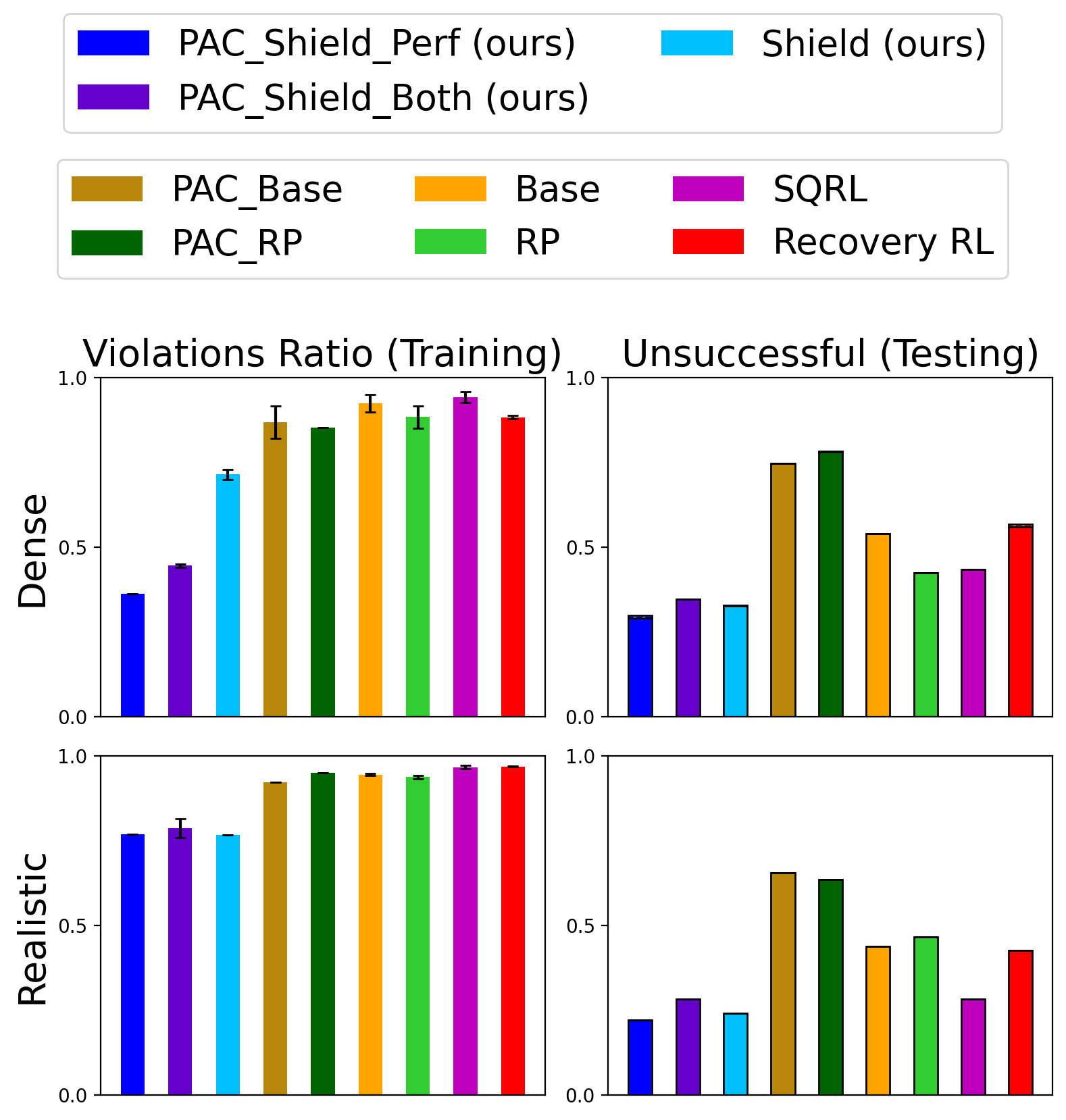}
        \caption{Advanced-Env (averaged over 3 seeds)}
    \end{subfigure}
    \caption{\textbf{Comparison of safety violations during Lab training and unsuccessful trials at test time:} Sim-to-Lab-to-Real (PAC{\myspace}Shield{\myspace}Perf and PAC{\myspace}Shield{\myspace}Both) has the lowest safety violations in both training and deployment. First, it showcases the benefits of using a shielding scheme in contrast with Base, RP and vanilla PAC-Bayes. Second, reachability safety critic enables safer exploration and safety satisfaction in deployment as compared to SQRL and recovery RL. Additionally, Sim-to-Lab-to-Real has lower unsuccessful ratio in deployment than Shield, which shows a diverse but safe policy distribution not only provides a generalization bound but also improves the empirical performance to novel environments.}
    \label{fig:main}
\end{figure}

We compare all the methods by (1) safety violations in Lab training and (2) success and safety at deployment (Figure~\ref{fig:main}). We calculate the ratio of number of safety violations to the number of episodes collected during training. For deployment, we show the percentage of failed trials (solid bars in Figure~\ref{fig:main}) and unfinished trials (hatched bars). We summarize the main findings below:
\new{
\begin{enumerate}
    \item Across Lab training, our proposed Sim-to-Lab-to-Real (PAC{\myspace}Shield{\myspace}Perf and PAC{\myspace}Shield{\myspace}Both) achieves the fewest safety violations. This demonstrates the efficacy of the reachability safety state-action value function for shielding. Compared to the risk-based safety critics in SQRL \citep{srinivasan2020learning} and Recovery RL \citep{thananjeyan2021recovery}, our safety critics can learn from near-failure and with dense cost signals, as discussed in \ref{sec:hj_risk_comparison}. Adding penalty in the reward function does not reduce safety violations significantly.
    \item In testing environments, Sim-to-Lab-to-Real achieves the lowest unsuccessful fraction of trajectories (solid bars plus hatched bars). This indicates that training a diverse and safe policy distribution achieves better generalization performance to novel environments. Sim-to-Lab-to-Real also achieves the fewest safety violations (solid bars) at test time. This suggests that explicitly enforcing hard safety constraints improves the safety not only in training but also in testing. In Sec.~\ref{sec:pac_bound} we show stronger generalization guarantees (for both performance and safety) compared to PAC-Bayes baselines.
    \item In Sec.~\ref{sec:phy_exp} we show Sim-to-Lab-to-Real achieves the best performance and safety among baselines when the policies are deployed on a quadrupedal robot navigating through real indoor environments. The empirical performance and safety also validate the theoretical generalization guarantees from PAC-Bayes Control.
    \item In Sec.~\ref{sec:sensitivity} we show that high diversity of trajectories from the latent distribution results in better generalization at test time. Without diversity maximization in Sim training, the resulting trajectories can concentrate close to a single one and hinder downstream fine-tuning in Lab. However, we also find that in Advanced-Env, PAC{\myspace}Base and PAC{\myspace}RP (distribution over policies) perform worse than Base and RP (single policy). We find that high diversity without shielding may hinder training progress due to frequent safety violations interfering with strategy exploration.
    \item We find that adding latent distribution to the backup policy introduces difficulty during Sim training, and leading to similar, if not worse, performance and safety at test time. We suspect that PAC{\myspace}Shield{\myspace}Both would take more samples to converge well in training and requires more careful tuning of hyperparameters. Following discussions focus on results of PAC{\myspace}Shield{\myspace}Perf, in which only the performance policy is conditioned on the latent variable.
    \item Compared to other Labs, violation ratios in Advanced-Realistic tend to be higher, although our methods still reduce safety violations by 20-25$\%$. Also, there are few unfinished trials at test time (the robot neither reaches the target nor collides with obstacles). Given the tight spacing in realistic indoor environment (Fig.~\ref{fig:backup_critic}b), the non-trivial dimensions of the quadruped robot, and the complex visuals, the backup policy can fail to ensure safety in some environments.
\end{enumerate}
}

\subsubsection{Reachability vs. Risk-Based Safety Critic} \label{sec:hj_risk_comparison}
\begin{figure}[!t]
    \centering
    \begin{subfigure}[b]{.475\textwidth}
        \centering
        \includegraphics[width=\textwidth]{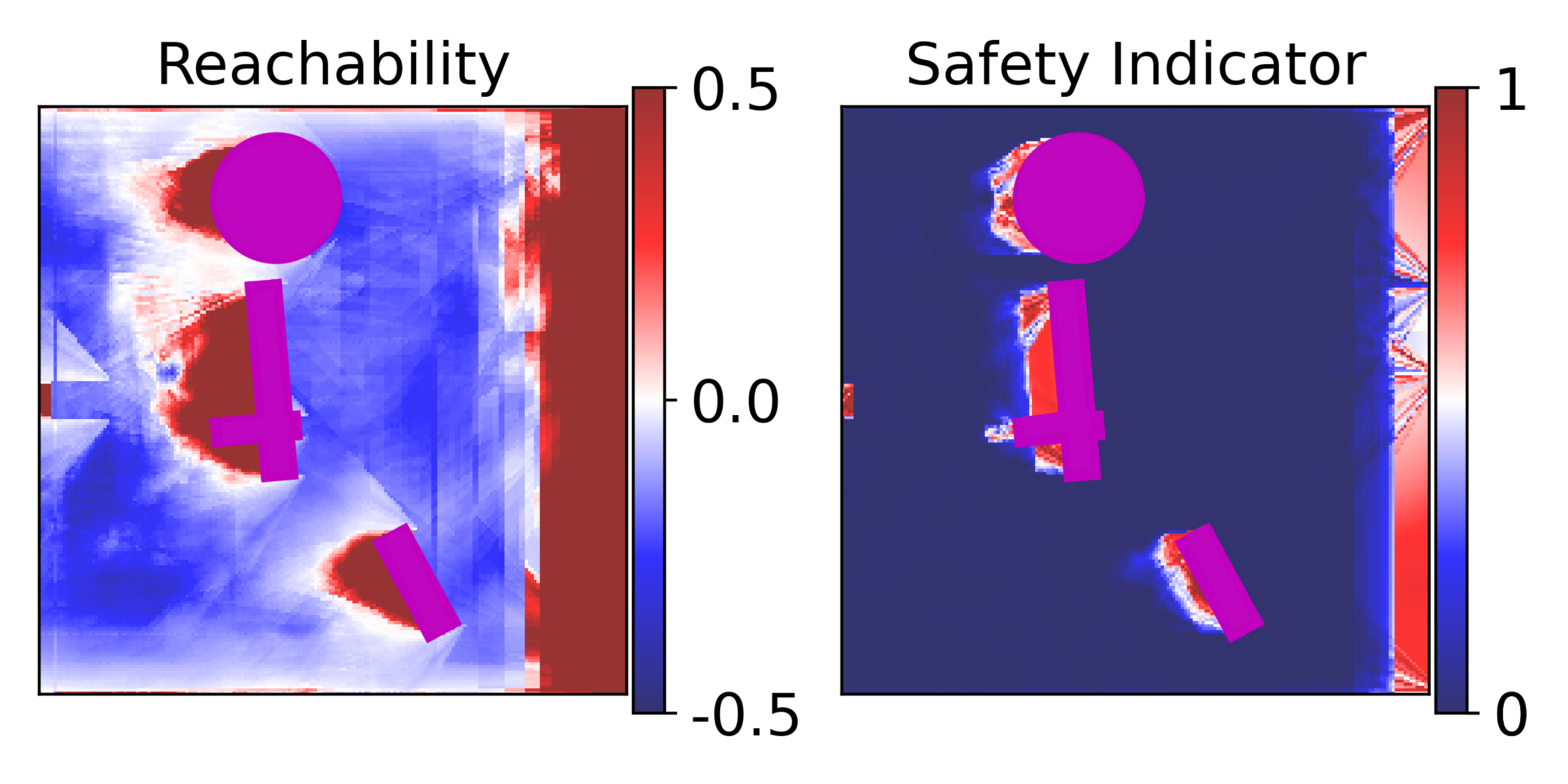}
        \caption{Lab: Vanilla-Normal}
    \end{subfigure}
    \begin{subfigure}[b]{.49\textwidth}
        \centering
        \includegraphics[width=\textwidth]{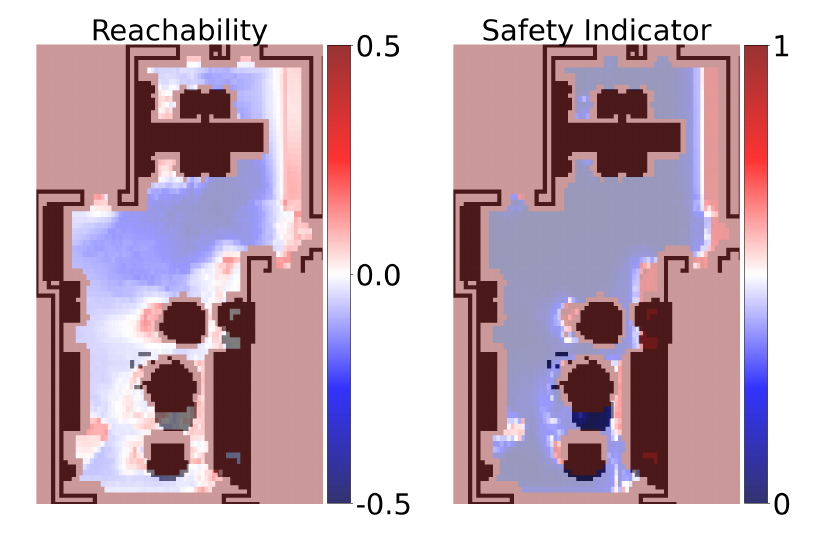}
        \caption{Lab: Advanced-Realistic}
    \end{subfigure}
    \newcaption{\textbf{2D slices of safety state-action value functions when the robot is facing to the right:} we train the safety critic using RL modified from Hamilton-Jacobi reachability analysis (``Reachability''), while SQRL and recovery RL train it with sparse binary indicators (``Safety Indicator''). Reachability safety critic better captures the unsafe region - there is a more gradual change in safety value near the obstacles (from blue to red color, lower to higher value), indicating that the robot is getting closer to the obstacles. In contrast, risk-based critic shows a more binary separation between safe and unsafe regions, leaving the robot little room and time to steer away from obstacles. The unsafe regions are also thinner than those learned with reachability. Thus, our reachability-based approach achieves fewer safety violations during both training and deployment.}
    \label{fig:backup_critic}
\end{figure}

Sim-to-Lab-to-Real and previous safe RL methods differ in the metric used to quantify safety and train the backup agent. By utilizing reachability RL, we have an exact encoding of the property we want our system to satisfy, i.e., the \textit{distance} should be no closer to obstacles than a specific threshold. In contrast, SQRL and Recovery RL define safety by the \textit{risk} of colliding with obstacles in the future and use binary safety indicators. We argue that risk-based threshold can easily overfit to specific scenarios since the probability heavily depends on the discount factor used. In addition, reachability objective allows the backup agent to learn from near failure, while the risk critic in SQRL and Recovery RL needs to learn from complete failures. Fig.~\ref{fig:backup_critic} shows 2D slices of the safety state-action values in both environment settings. Reachability critics provide thicker unsafe regions next to obstacles, while risk-based critics fail to recognize many unsafe regions or consider unsafe only when very close to obstacles. Among different Lab setups, compared to the baselines, our method reduces safety violations by $77\%$, $4\%$, $76\%$, $62\%$, and $23\%$ in training and $38\%$, $26\%$, $54\%$, $34\%$, and $28\%$ in deployment.\\

\begin{figure}[!t]
    \centering
    \begin{subfigure}[b]{.375\textwidth}
        \centering
        \includegraphics[height=6cm]{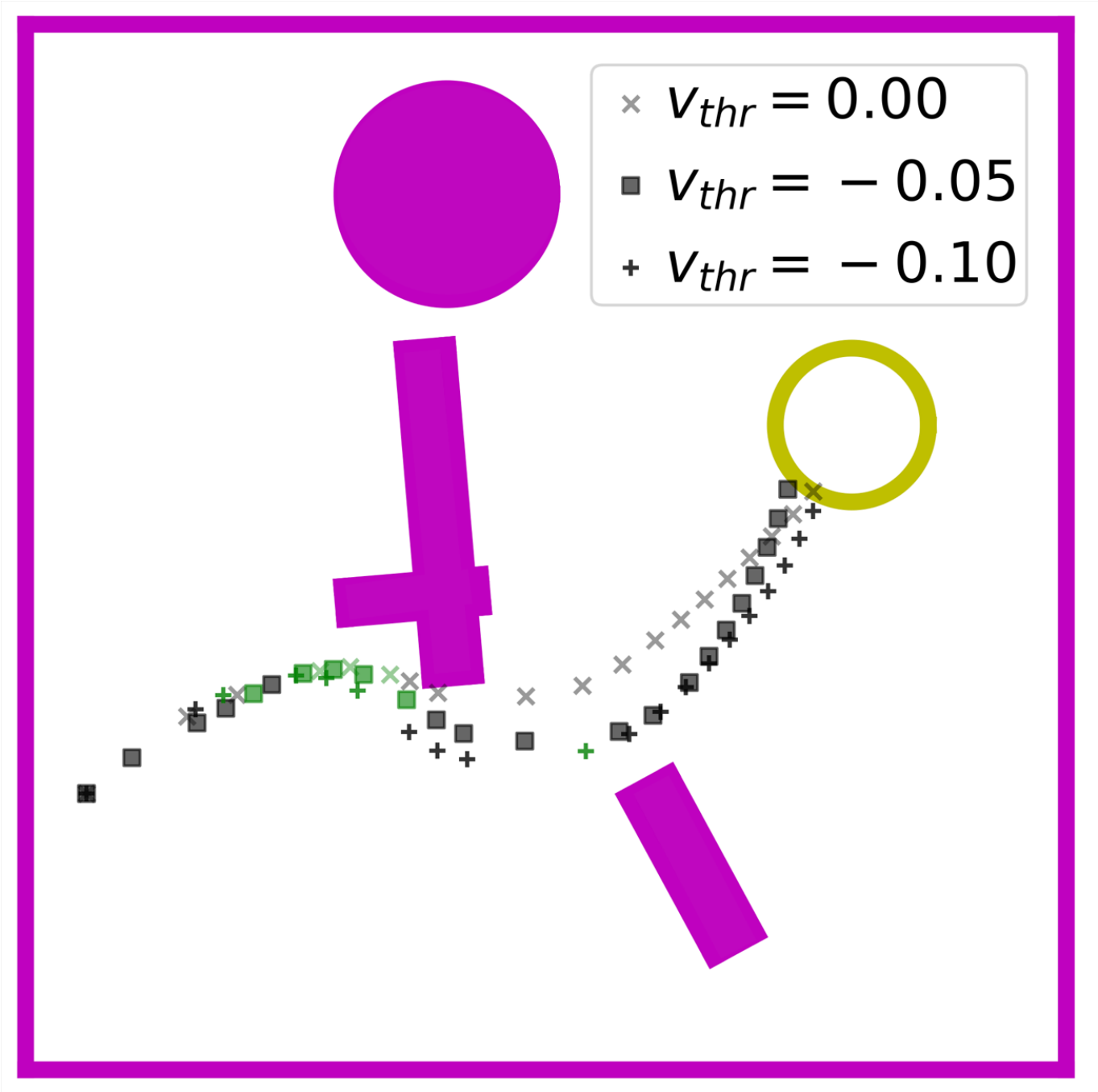}
        \caption{Lab: Vanilla-Normal}
    \end{subfigure}
    \begin{subfigure}[b]{.575\textwidth}
        \centering
        \includegraphics[height=6cm]{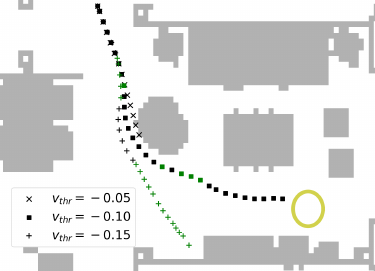}
        \caption{Lab: Advanced-Realistic}
    \end{subfigure}
    \caption{\textbf{Rollout trajectories using different value threshold for shielding:} higher threshold (more negative) results in more conservative maneuver, i.e., keeping farther away from obstacles (purple in (a) and grey in (b)). In Advanced-Env, the complex visuals and tight spacing cause challenges in learning the backup agent. We tend to find a relatively conservative threshold ($v_{thr}=-0.10$) works well in practice, and too high threshold can prevent the robot from reaching the goal and accidentally steer it towards tight space.}
    \label{fig:shielding_threshold}
\end{figure}

\begin{figure}[!t]
    \centering
    \includegraphics[width=\textwidth]{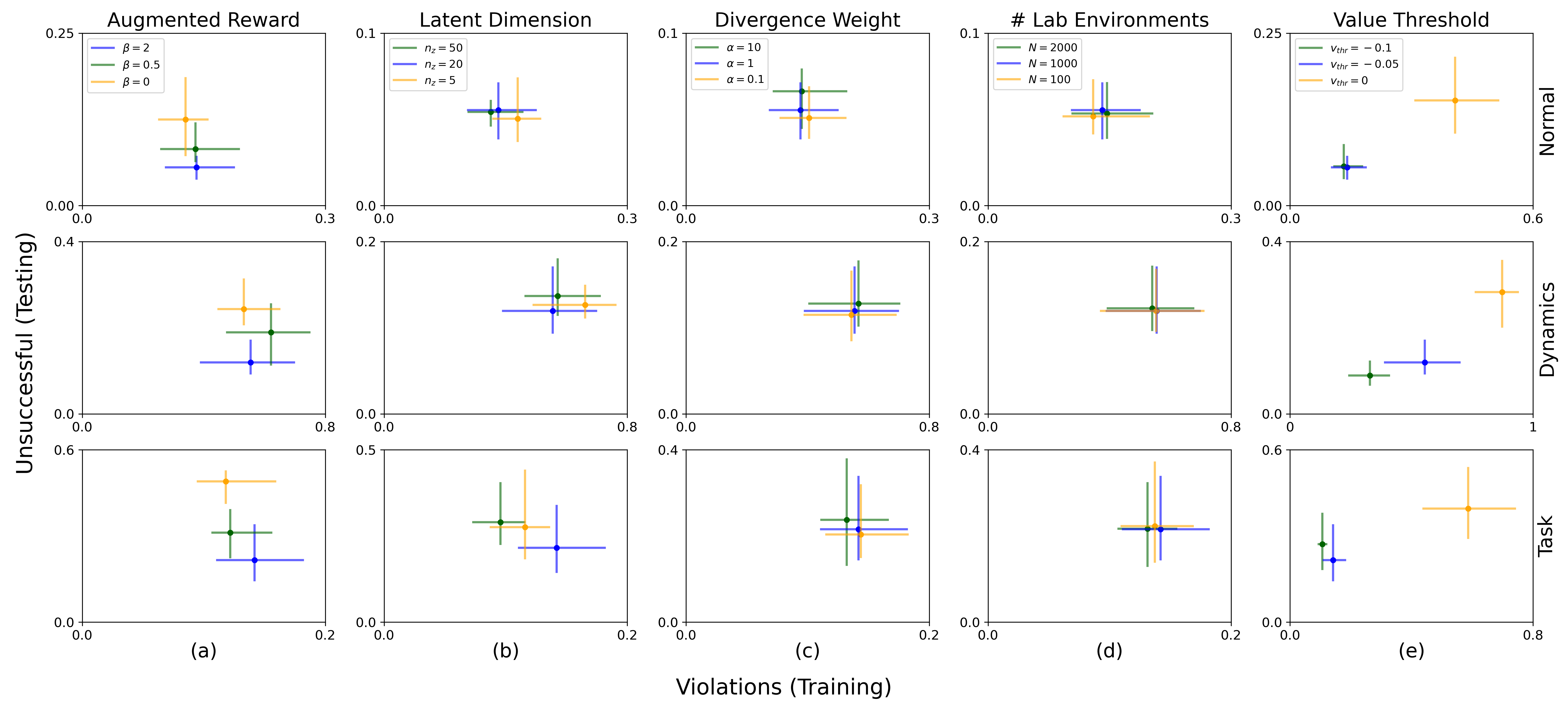}
    \caption{\textbf{Sensitivity analyses:} we study the influence of different hyper-parameters to Sim-to-Lab-to-Real. The results are averaged over $5$ seeds in Vanilla-Env. If not specified, the hyper-parameters default to $\beta=2$, $n_z=20$, $\alpha=1, N=1000$, and $v_{thr}=-0.05$, as shown in blue. Results suggest the augmented reward in Sim training and the value threshold in shielding are the two most important hyper-parameters.}
    \label{fig:sensitivity}
\end{figure}

\noindent \textbf{Sensitivity analysis: value threshold.} Through experiments, we find the value threshold used in shielding essential to performance and safety. We first investigate how the threshold using during \emph{training} affects the final results among the three Lab settings in Vanilla-Env, which are shown in Fig.~\ref{fig:sensitivity}e. $v_{thr}=0$ naturally results in more safety violations during training compared to $v_{thr}=-0.05$ and $v_{thr}=-0.10$. Policies trained with $v_{thr}=0$ also performs the worst at test time, which indicates that less shielding during training makes the robot learn unsafe or aggressive maneuver. 
Next we evaluate how the value threshold affects robot trajectories at \emph{test} time. Fig.~\ref{fig:shielding_threshold} shows the trajectories using different thresholds in the two settings. Small threshold leads to robot passing very closely next to obstacles, while a bigger threshold leads to more conservative behavior. We also would like to highlight the challenges of learning safe policies in Advanced-Env. As shown in the figure, with $v_{thr}=-0.15$ the robot avoids the first obstacle, and then the backup policy steers the robot away from the target, potentially deeming the clearance next to the target not sufficient. However, this brings the robot near the wall, and due to imperfect training of the backup actor, the robot fails to escape. With tight spacing and large dimensions of the robot in Advanced-Env, we find the backup agent more difficult to train, and the final test performance and safety can be sensitive to the shielding threshold. In Advanced-Realistic, average test success rate with $v_{thr}=-0.05, -0.1, -0.15$ are 0.678, 0.786, and 0.762 respectively. Future work could look into adapting the threshold after short experiences in different environments online.

\subsubsection{Generalization Guarantees} \label{sec:pac_bound}
\begin{table}[!ht]
\footnotesize
    \centering
    \caption{\textbf{Results of PAC-Bayes guarantees and test success and safety:} to compute the bound, each environment has $1000$ policies sampled from the latent distribution and tested. The results in the first two rows are based on PAC{\myspace}Shield{\myspace}Perf.}
    \centerline{
    \begin{tabular}{ccccc}
        \toprule
        & \phantom{a} & \multicolumn{3}{c}{Advanced-Realistic} \\
        \cmidrule{3-5} 
        Method && PAC{\myspace}Shield{\myspace}Perf & PAC{\myspace}Base & SQRL \\
        $\#$ Lab Environments && 1000 & 1000 & 1000 \\ \midrule
        Success Bound && 0.701 & 0.297 & - \\
        True Expected Success       && \bftab 0.786 & 0.366 & 0.712 \\
        Real Robot Success  && \bftab 0.767 & 0.433 & 0.667 \\ \midrule
        Safety Bound && 0.708 & 0.304 & - \\
        True Expected Safety       && \bftab 0.794 & 0.367 & 0.713 \\
        Real Robot Safety  && \bftab 0.867 & 0.433 & 0.667 \\
    \end{tabular}
    }
    \vspace{3mm}
    \centerline{
    \begin{tabular}{cccccccccccccc}
        \toprule
        \phantom{a} & \multicolumn{5}{c}{Vanilla-Normal} & \phantom{a} & \multicolumn{5}{c}{Vanilla-Dynamics} \\
        \cmidrule{2-6} \cmidrule{8-12}
        Method & \multicolumn{4}{c}{PAC{\myspace}Shield{\myspace}Perf} & PAC{\myspace}Base && \multicolumn{4}{c}{PAC{\myspace}Shield{\myspace}Perf} & PAC{\myspace}Base \\
        Divergence Weight & $1$ & $1$ & $1$ & $10$ & $1$ & & $1$ & $1$ & $1$ & $10$ & $1$\\
        $\#$ Lab Environments & 100 & 1000 & 2000 & 1000 & 1000 & & 100 & 1000 & 2000 & 1000 & 1000 \\ \midrule
        Success Bound & 0.778 & 0.876 & 0.900 & 0.896 & 0.735 && 0.692 & 0.820 & 0.839 & 0.828 & 0.778 \\
        True Expected Success       & 0.948 & 0.945 & 0.947 & 0.934 & 0.886 && 0.881 & 0.880 & 0.878 & 0.872 & 0.843 \\ \midrule
        Safety Bound & 0.793 & 0.911 & 0.917 & 0.913 & 0.816 && 0.717 & 0.835 & 0.851 & 0.837 & 0.815 \\
        True Expected Safety       & 0.954 & 0.954 & 0.954 & 0.953 & 0.902 && 0.888 & 0.887 & 0.887 & 0.883 & 0.852 \\
        \bottomrule
    \end{tabular}
    }
    \vspace{3mm}
    \centerline{
    \begin{tabular}{cccccccccccccccc}
        \phantom{a} & \multicolumn{5}{c}{Vanilla-Task} & \phantom{a} & \multicolumn{6}{c}{Advanced-Dense} \\
        \cmidrule{2-6} \cmidrule{8-13}
        Method & \multicolumn{4}{c}{PAC{\myspace}Shield{\myspace}Perf} & PAC{\myspace}Base && \multicolumn{5}{c}{PAC{\myspace}Shield{\myspace}Perf} & PAC{\myspace}Base \\
        Divergence Weight & $1$ & $1$ & $1$ & $10$ & $1$ && $2$ & $2$ & $2$ & $1$ & $5$ & 2 \\
        $\#$ Lab Environments & 100 & 1000 & 2000 & 1000 & 1000 && 100 & 500 & 1000 & 500 & 500 & 1000\\ \midrule
        Success Bound & 0.578 & 0.757 & 0.792 & 0.777 & 0.468 && 0.402 & 0.578 & 0.623 & 0.512 & 0.557 & 0.254 \\ 
        True Expected Success & 0.847 & 0.851 & 0.844 & 0.853 & 0.590 && 0.577 & 0.663 & 0.703 & 0.621 & 0.644 & 0.327 \\ \midrule
        Safety Bound & 0.769 & 0.884 & 0.899 & 0.887 & 0.663 && 0.412 & 0.579 & 0.630 & 0.518 & 0.564 & 0.259 \\
        True Expected Safety & 0.939 & 0.939 & 0.940 & 0.938 & 0.796 && 0.583 & 0.671 & 0.709 & 0.629 & 0.652 & 0.332 \\
        \bottomrule
    \end{tabular}
    }
    \label{tab:bound}
\end{table}

In this subsection, we evaluate the PAC-Bayes generalization guarantees obtained after Lab training, and the effect of adding reachability shielding in the policy architecture to the bounds. Table~\ref{tab:bound} shows the bounds and test results on safety (not colliding with obstacles) and success (safely reaching the goal) among Lab training. The true expected success and safety are tested with environments that are similar to the Lab training environments (of the same distribution) but unseen before. In all settings, the true expected success and safety are higher than the bound in all settings, which validates the guarantees derived using PAC-Bayes Control. Furthermore, we compare the bound trained using PAC{\myspace}Shield{\myspace}Perf with previous PAC-Bayes Control method (PAC{\myspace}Base) in the Vanilla-Env and Advanced-Realistic. With shielding, the generalization bound improves in all settings. In the difficult setting of Advanced-Realistic, the bound improves from $0.366$ to $0.786$ for task completion and from $0.367$ to $0.794$ for safety satisfaction. Thus, explicitly enforcing hard safety constraints not only improves empirical outcomes but also provides stronger certification to policies in novel environments. In Sec.~\ref{sec:phy_exp} we also demonstrate empirical results of physical robot experiments validating the guarantees. \\

\newpage
\noindent \textbf{Sensitivity analysis: weight of policy distribution regularization ($\alpha$).} When optimizing the generalization bound \eqref{eq:ps-update}, we place a weighting coefficient $\alpha$ to balance gradients of the training reward and of the estimated KL divergence between the prior and posterior policy distribution, $P_0$ and $P$. Here we study the effect of using different values of $\alpha$ in the generalization bound and test performance. Fig.~\ref{fig:sensitivity}c shows that too strong regularization ($\alpha=10$) prevents the Lab training from tuning the prior distribution sufficiently, resulting in worse testing performance after training. The effect of different $\alpha$ is more prominent in Advanced-Dense training. With same 500 training environments, $\alpha=2$ achieves 0.578 on success bound while 0.512 for $\alpha=1$ and 0.557 for $\alpha=5$.\\

\noindent \textbf{Sensitivity analysis: number of Lab environments ($N$).} Thm.~\ref{thm:pac bayes control} indicates the PAC-Bayes bound depends on the number of environments used in the Lab training. Fig.~\ref{fig:sensitivity}d demonstrates that in Vanilla-Env, $N$ does not have a significant effect on training safety violations and test performance. We suspect that training in Vanilla-Env does not require a large number of environments for generalization. In the more difficult Advanced-Dense, with the same $\alpha=2$, higher $N=1000$ achieves the best test success (0.703) and safety (0.709) compared to smaller $N=100$ and $N=500$ (Table.~\ref{tab:bound}), \new{which demonstrates that a higher number of Lab environments help fine-tuning the policies achieving strong generalization in complex environments.}

\subsubsection{Physical Experiments} \label{sec:phy_exp}
\begin{figure}[!t]
    \centering
    \includegraphics[width=\textwidth]{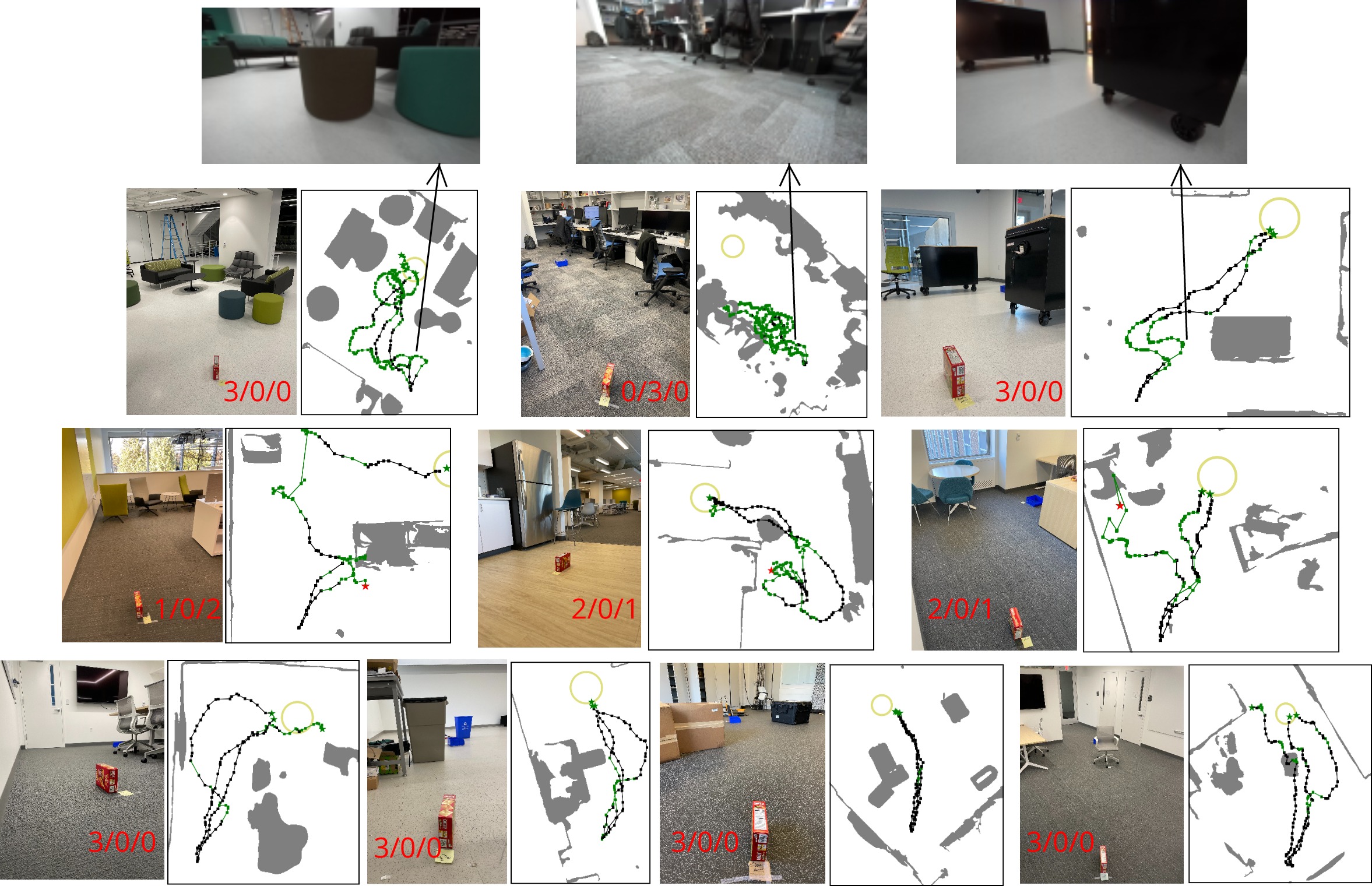}
    \caption{\textbf{Environments for physical robot experiments and robot trajectories/observations with PAC{\myspace}Shield{\myspace}Perf:} we run the policy three times in each environment by sampling different latent variables from the posterior distribution. The three numbers in images indicates success/unfinished/failure split. Green dots indicates shielding in effect. Green star indicates success in reaching the target. Red star indicates colliding with obstacles. We scan the environment using an iPad Pro tablet before experiments to generate the 2D map (which the robot does not have access to). The robot trajectory is obtained using localization algorithm of the onboard camera, and is inaccurate at places (intersecting obstacles; not exactly reaching the target but the robot deems so, which we consider success).}
    \label{fig:phy_exp}
\end{figure}

To demonstrate empirical performance and safety of trained policies in real environments (Lab-to-Real transfer) and verify the generalization guarantees, we evaluate the policies in real indoor environments in the Engineering Quadrangle building at Princeton University. We deploy a Ghost Spirit quadrupedal robot equipped with a ZED 2 camera at the front (Fig.~\ref{fig:environment}d), matching the same dynamics and observation model used in Advanced-Realistic Lab. For the distance and relative bearing to the goal, before each trial the robot is given the ground-truth measurement at the initial location, and then it uses the localization algorithm native to the stereo camera \new{to update the measurement at each step.}

We pick ten different locations with furniture configurations and difficulty similar to those in Advanced-Realistic Lab. Based on test results after Lab training, we run policies trained with PAC{\myspace}Shield{\myspace}Perf (best performance overall), PAC{\myspace}Base (PAC-Bayes baseline with low generalization guarantees), and SQRL (best overall among other baselines). Each policy is evaluated at one environment 3 times (30 trials total). The results are shown in Table.~\ref{tab:bound}. Our policy is able to achieve the best performance (0.767) and safety (0.867), validating the theoretical guarantees from PAC-Bayes Control. The upper-right of Fig. \ref{fig:sim2lab2real} shows a trajectory when running policies trained with PAC{\myspace}Shield in a kitchen environment where the backup policy and the shielding discriminator help the robot avoid hitting the obstacles and reach the target successfully.

Fig.~\ref{fig:phy_exp} shows the 10 real environments and robots' trajectories when running policies trained with PAC{\myspace}Shield{\myspace}Perf. Green dots indicate shielding in effect, which is activated often near obstacles. The first and third images on top of the figure show the robot's view when shielding successfully guides robot away from the sofa stool and the cabinet. In the second environment, the backup policy keeps shielding the robot away from center of the room with value threshold $v_{thr}=-0.10$, and all three trials ended as unfinished. This is possibly due to the cluttered scene of desks at the top half of the observation. We also test with small value threshold $v_{thr}=-0.05$ during experiments, and the robot is able to reach the target without shielding always activated. This highlights the need for adapting the shielding value threshold online in future work.

\subsubsection{Other Studies} \label{sec:sensitivity}

\vspace{2pt}
\noindent \textbf{Ablation Study: importance of two-stage training.}
We evaluate the significance of Lab training by testing the prior policy distribution (without fine-tuning in Lab) in Vanilla-Env. Without Lab training, the unsuccessful ratio in deployment increases by $16\%$, $8\%$ and $14\%$. This suggests that Lab training is essential to policies adapting to real dynamics and new environment distributions. Additionally, we test the importance of Sim training with \textit{Shield} (no policy distribution). Without Sim training, the safety violations in Lab training increases by $60\%$, $11\%$ and $65\%$. This demonstrates that Sim training enables the backup agent to monitor and override unsafe behavior from the beginning of Lab training.\\

\begin{figure}[!t]
    \centering
    \begin{subfigure}[b]{.475\textwidth}
        \centering
        \includegraphics[width=\textwidth]{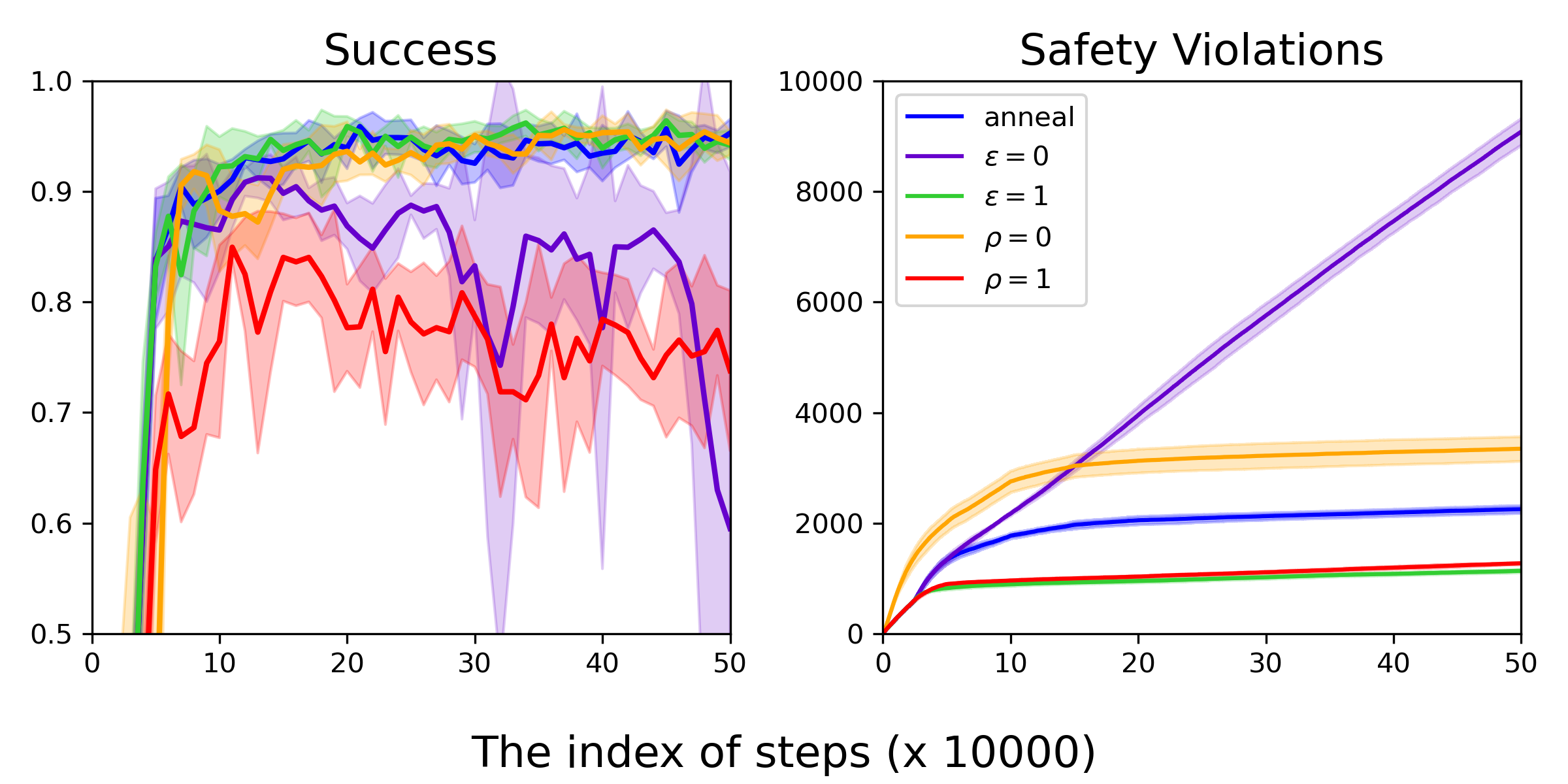}
        \caption{Vanilla-Env}
    \end{subfigure}
     \begin{subfigure}[b]{.475\textwidth}
        \centering
        \includegraphics[width=\textwidth]{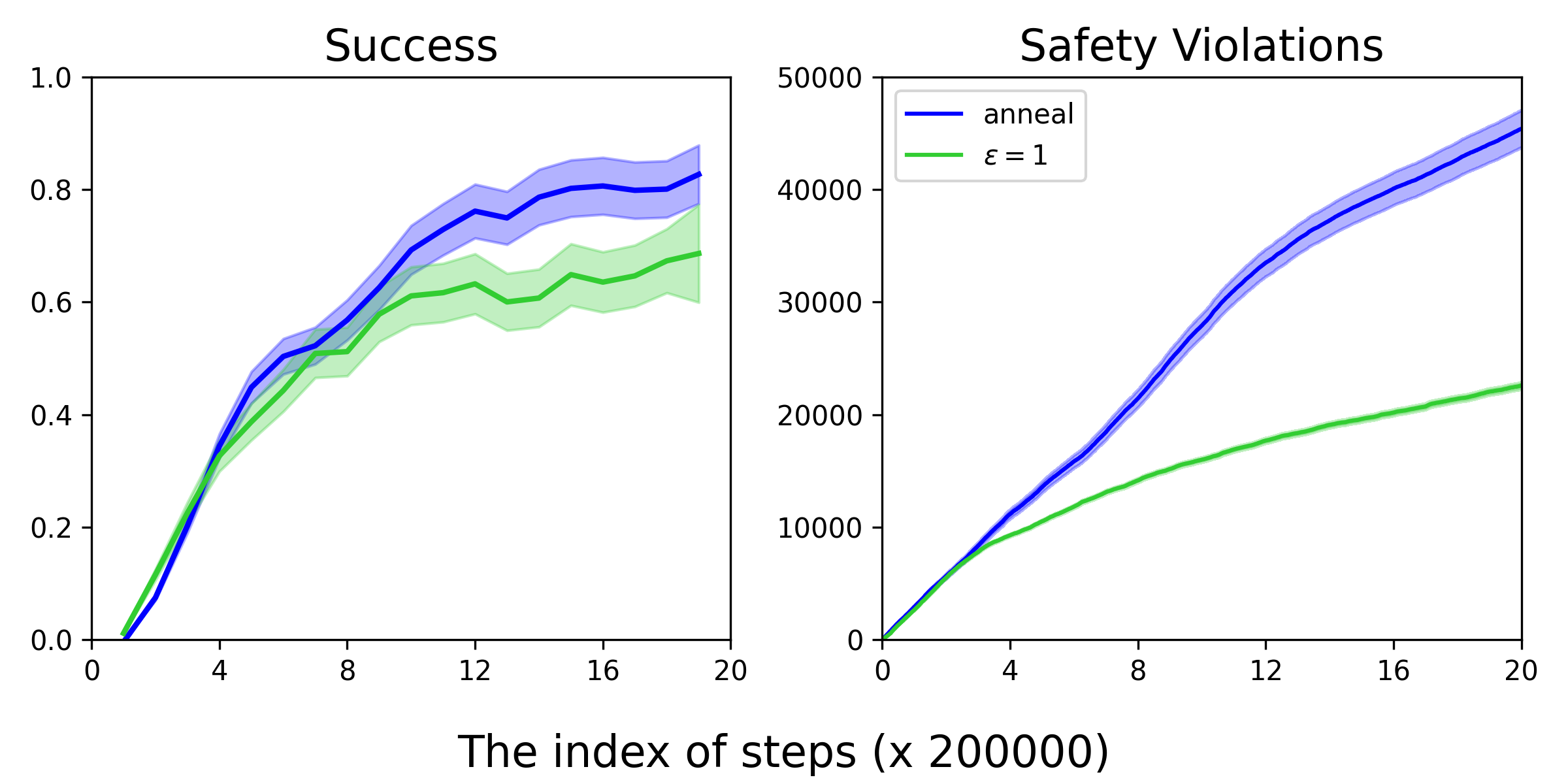}
        \caption{Advanced-Env}
    \end{subfigure}
    \newcaption{\textbf{Effect of $\backupProb$ and $\shieldProb$ scheduling in Sim training:} annealing $\backupProb$ and $\shieldProb$ helps balance between safety violations and  task completion. If not specified, for Vanilla-Env, $\backupProb$ initializes at $1$ and decays by $0.5$ every $25000$ steps, and $\shieldProb$ initializes at $0$ with $1-\shieldProb$ decaying by $0.5$ every $50000$ steps. For Advanced-Env, $\backupProb$ initializes at $0.5$ and decays by $0.5$ every $500000$ steps, and $\shieldProb$ initializes at $0$ with $1-\shieldProb$ decaying by $0.5$ every $200000$ steps. The results are over $5$ random seeds for Vanilla-Env and $3$ random seeds for Advanced-Env.}
    \label{fig:scheduling}
\end{figure}

\noindent \textbf{Sensitivity analysis: the probability of sampling actions from the backup policy ($\backupProb$) and the probability of activating shielding ($\shieldProb$).}
One of the main contributions of our work is the effective joint training of both performance and back agents (realized in Sim training). The two parameters, $\backupProb$ and $\shieldProb$, directly affect the exploration in Sim training. With high $\rho$ or high $\epsilon$, the RL agent basically only explores conservatively within a small safe region. However, in the beginning of the training, we should allow the RL agent to collect diverse state-action pairs. On the other hand, we also gradually anneal $\backupProb \rightarrow 0$ and $\shieldProb \rightarrow 1$ since we want the performance policy to be aware of the backup policy. In other words, the performance policy is effectively in \textit{shielded environments} towards end of Sim training. Fig.~\ref{fig:scheduling} shows the Sim training progress under different $\backupProb$ and $\shieldProb$ scheduling. With constant $\backupProb=0$ or $\shieldProb=0$, the number of safety violations is much higher than that with both parameters annealing. Even worse, $\shieldProb=0$ results in the number of safety violations increase at constant speed and the training success fluctuates significantly. On the other hand, with $\backupProb=1$ or $\shieldProb=1$, the number of safety violations is only half as that with both parameters annealing. However, this is at the expense of exploration and leads to worse success rate in deployment. In Vanilla-Env $\backupProb=1$ leads to very poor training success. Although in Vanilla-Env $\shieldProb=1$ does not have significant effect on training success, in the Advanced-Env, insufficient exploration hinders training progress. Also note that Sim training is not safety-critical and we do not aim to reduce safety violations then.\\

\begin{figure}[!t]
    \centering
    \includegraphics[width=0.6\textwidth]{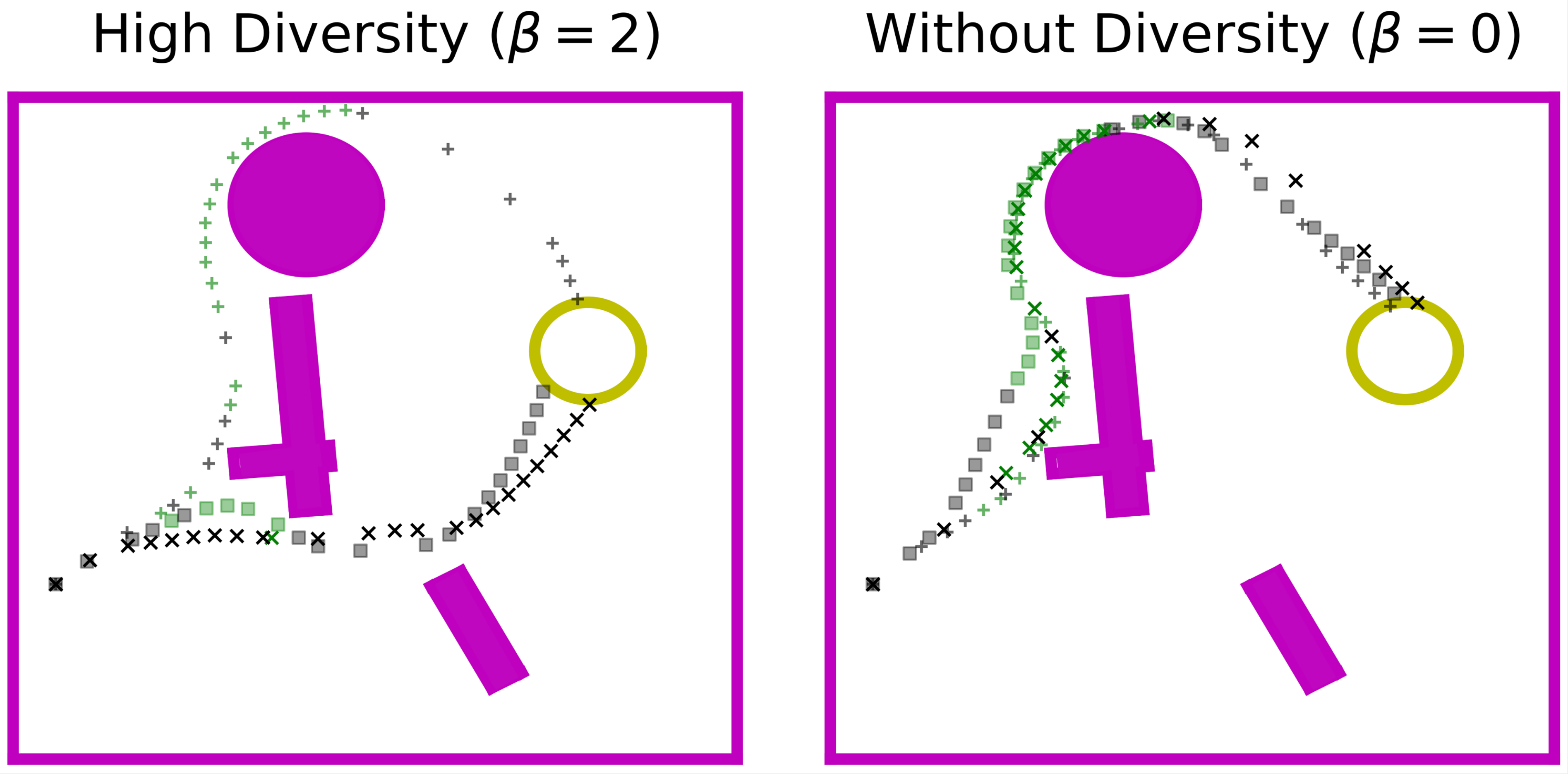}
    \caption{\textbf{High augmented reward coefficient induces a diverse policy distribution:} the diversity is essential to fine-tuning the latent distribution in Lab training and to good generalization to novel environments. Black markers indicate actions from the performance policy being executed, and green markers are for actions from the backup policy.}
    \label{fig:traj_diversity}
\end{figure}

\noindent \textbf{Sensitivity analysis: diversity-induced Sim training.}
We argue that training a diverse and safe policy distribution helps improve safety and performance in novel environments. There are two hyper-parameters in our algorithm affecting the diversity, i.e., augmented reward coefficient $\beta$ and latent dimension $n_z$. Fig.\ref{fig:sensitivity}a and Fig.\ref{fig:sensitivity}b show the violation ratio in Lab training and unsuccessful ratio in testing under different $(\beta, n_z)$ choices. We find that training without augmented reward ($\beta=0$) results in the lowest violation ratio; however, the unsuccessful ratio in testing is the highest. In fact, we observe that with $\beta=0$, rollout trajectories conditioned on different latent variables almost converge to a single trajectory as shown in Fig.~\ref{fig:traj_diversity}. This reflects why safety is better satisfied but at the expense of generalization. On the other hand, when the coefficient is sufficiently large ($\beta=2$), the policy distribution becomes diverse and generalizes well to unseen testing environments. Note that even with high diversity, safety can still be well ensured with shielding. For the second source of diversity, our proposed Sim-to-Lab-to-Real is robust to different latent dimension.

\section{Conclusion}
\label{sec:conclusion}

In this work, we propose the Sim-to-Lab-to-Real framework that combines Hamilton-Jacobi reachability analysis and PAC-Bayes generalization guarantees to  \new{bridge the Sim-to-Real reality gap with a probabilistically guaranteed safety-aware policy distribution.} Joint training of a performance and a backup policy in Sim training (1st stage) enables a safety-aware exploration during Lab training (2nd stage). By optimizing the generalization bounds in Lab training, our approach is able to probabilistically certify robot performance and safety before deployment. We demonstrate significant reduction in safety violations in training and stronger performance and safety during test time. Results from experiments with a quadrupedal robot in real indoor space validate the theoretical guarantees.
\new{
    \subsection{Discussion: Environment distribution.} As elaborated in Sec.~\ref{sec:formulation}, the generalization guarantees obtained through our framework assumes no distribution shift between Lab and Real in terms of environments. To bridge the discrepancy, we model the real environments by using (1) photorealistic dataset of indoor room layouts and furniture models and (2) dynamics from system identification of the real robot and camera poses. Additionally, we note that previous works in PAC-Bayes Control \citep{majumdar2021pac, ren2020generalization, veer2020probably} have consistently shown real deployment validating the bounds. Even under a slight of shift in distribution, we believe that a certificate of performance and safety is useful and provides confidence for deploying the system.
}
\new{
    \subsection{Discussion: Large-scale Lab training.} We acknowledge that one limitation of our framework is that, in exchange for assuming close to nothing about the environment distribution and providing statistical guarantees that hold in arbitrarily \emph{high confidence} instead of in \emph{expectation} only (e.g., conformal prediction \citep{shafer2008tutorial}), we require at least a few hundred environments for ``Lab’’ training to achieve tight PAC-Bayes generalization guarantees (e.g., $<10\%$ difference between empirical performance and theoretical guarantee), which means performing ``Lab’’ training with \emph{real} conditions can be difficult for us researchers in university labs with limited hardware, computation, and human resources. In this work, we resort to performing ``Lab'' training in realistic simulated environments.
}

\new{
    Nonetheless, we envision that our framework is well suited for industry practitioners who have access to either extensive training facilities (e.g. Google's robot ``farms'' \citep{levine2016learning}, Boston Dynamics' testing warehouse \citep{bd2022}), large-scale distributed systems (e.g. Amazon's warehouses \citep{amazon19}), or vast amounts of ``Lab-like'' data collection (e.g. Cruise and Waymo's thousands–millions of test driver miles \citep{waymo22}). For these practical and often safety-critical applications, our framework can improve safety during training and provide generalization guarantees for performance and safety at deployment. For university labs achieving similar scales of data collection and training, it would be promising to explore (1) crowdsourcing robots training across labs \citep{ichnowski2022fogros} and (2) mechanisms for automatically resetting the robot \citep{eysenbach2018leave} and randomizing the environments.
}

\new{
    On the theoretical front, first it would be worth identifying the most representative environments for training (e.g., using coresets \citep{borsos2020coresets}). PAC-Bayes guarantee holds as long as the policies are ``evaluated'' in the training environments $\envSetPs$ and the training reward $R_\envSetPs$ is evaluated.  We could potentially obtain similar tight generalization guarantees by training on a much smaller set of environments compared to $\envSetPs$ used in this work. Second, recent growing interest in PAC-Bayes bound \citep{guedj2019primer} and other types of generalization guarantees \citep{arora2018stronger} could lead to tighter and also more sample-efficient bounds for certifying the generalization performance and safety.
}

\section*{Acknowledgement}
Allen Z. Ren and Anirudha Majumdar were supported by the Toyota Research Institute (TRI), the NSF CAREER award [2044149], the Office of Naval Research [N00014-21-1-2803], and the School of Engineering and Applied Science at Princeton University through the generosity of William Addy ’82. This article solely reflects the opinions and conclusions of its authors and not ONR, NSF, TRI or any other Toyota entity. We would like to thank Zixu Zhang for his valuable advice on the setup of the physical experiments.

\bibliography{ref_safe_rl.bib, ref_rl.bib, ref_reachability.bib, ref.bib}

\begin{thebibliography}{73}
\expandafter\ifx\csname natexlab\endcsname\relax\def\natexlab#1{#1}\fi
\providecommand{\url}[1]{\texttt{#1}}
\providecommand{\href}[2]{#2}
\providecommand{\path}[1]{#1}
\providecommand{\DOIprefix}{doi:}
\providecommand{\ArXivprefix}{arXiv:}
\providecommand{\URLprefix}{URL: }
\providecommand{\Pubmedprefix}{pmid:}
\providecommand{\doi}[1]{\href{http://dx.doi.org/#1}{\path{#1}}}
\providecommand{\Pubmed}[1]{\href{pmid:#1}{\path{#1}}}
\providecommand{\bibinfo}[2]{#2}
\ifx\xfnm\relax \def\xfnm[#1]{\unskip,\space#1}\fi
\bibitem[{Kumar et~al.(2021)Kumar, Fu, Pathak, and Malik}]{kumar2021rma}
\bibinfo{author}{A.~Kumar}, \bibinfo{author}{Z.~Fu},
  \bibinfo{author}{D.~Pathak}, \bibinfo{author}{J.~Malik},
\newblock \bibinfo{title}{{RMA: Rapid Motor Adaptation for Legged Robots}},
\newblock in: \bibinfo{booktitle}{Proceedings of Robotics: Science and Systems
  (RSS)}, \bibinfo{address}{Virtual}, \bibinfo{year}{2021}.
  \DOIprefix\doi{10.15607/RSS.2021.XVII.011}.
\bibitem[{Zhu et~al.(2017)Zhu, Mottaghi, Kolve, Lim, Gupta, Fei-Fei, and
  Farhadi}]{zhu2017target}
\bibinfo{author}{Y.~Zhu}, \bibinfo{author}{R.~Mottaghi},
  \bibinfo{author}{E.~Kolve}, \bibinfo{author}{J.~J. Lim},
  \bibinfo{author}{A.~Gupta}, \bibinfo{author}{L.~Fei-Fei},
  \bibinfo{author}{A.~Farhadi},
\newblock \bibinfo{title}{Target-driven visual navigation in indoor scenes
  using deep reinforcement learning},
\newblock in: \bibinfo{booktitle}{Proceedings of the IEEE International
  Conference on Robotics and Automation (ICRA)}, \bibinfo{year}{2017}, pp.
  \bibinfo{pages}{3357--3364}. \DOIprefix\doi{10.1109/ICRA.2017.7989381}.
\bibitem[{Tobin et~al.(2017)Tobin, Fong, Ray, Schneider, Zaremba, and
  Abbeel}]{tobin2017domain}
\bibinfo{author}{J.~Tobin}, \bibinfo{author}{R.~Fong},
  \bibinfo{author}{A.~Ray}, \bibinfo{author}{J.~Schneider},
  \bibinfo{author}{W.~Zaremba}, \bibinfo{author}{P.~Abbeel},
\newblock \bibinfo{title}{Domain randomization for transferring deep neural
  networks from simulation to the real world},
\newblock in: \bibinfo{booktitle}{Proceedings of the IEEE/RSJ International
  Conference on Intelligent Robots and Systems (IROS)}, \bibinfo{year}{2017},
  pp. \bibinfo{pages}{23--30}. \DOIprefix\doi{10.1109/IROS.2017.8202133}.
\bibitem[{Muratore et~al.(2021)Muratore, Ramos, Turk, Yu, Gienger, and
  Peters}]{muratore2021robot}
\bibinfo{author}{F.~Muratore}, \bibinfo{author}{F.~Ramos},
  \bibinfo{author}{G.~Turk}, \bibinfo{author}{W.~Yu},
  \bibinfo{author}{M.~Gienger}, \bibinfo{author}{J.~Peters},
  \bibinfo{title}{Robot learning from randomized simulations: A review},
  \bibinfo{year}{2021}. \href{http://arxiv.org/abs/2111.00956}{\tt
  arXiv:2111.00956}.
\bibitem[{Sadeghi and Levine(2017)}]{sadeghi2017cad2rl}
\bibinfo{author}{F.~Sadeghi}, \bibinfo{author}{S.~Levine},
\newblock \bibinfo{title}{Cad2rl: Real single-image flight without a single
  real image},
\newblock in: \bibinfo{booktitle}{Proceedings of Robotics: Science and Systems
  (RSS)}, \bibinfo{address}{Cambridge, Massachusetts}, \bibinfo{year}{2017}.
  \DOIprefix\doi{10.15607/RSS.2017.XIII.034}.
\bibitem[{Fu et~al.(2021)Fu, Cai, Gao, Zhang, Wang, Li, Zeng, Sun, Jia, Zhao,
  and Zhang}]{fu20213dfront}
\bibinfo{author}{H.~Fu}, \bibinfo{author}{B.~Cai}, \bibinfo{author}{L.~Gao},
  \bibinfo{author}{L.-X. Zhang}, \bibinfo{author}{J.~Wang},
  \bibinfo{author}{C.~Li}, \bibinfo{author}{Q.~Zeng}, \bibinfo{author}{C.~Sun},
  \bibinfo{author}{R.~Jia}, \bibinfo{author}{B.~Zhao},
  \bibinfo{author}{H.~Zhang},
\newblock \bibinfo{title}{{3D-FRONT: 3D Furnished Rooms With layOuts and
  semaNTics}},
\newblock in: \bibinfo{booktitle}{Proceedings of the IEEE/CVF International
  Conference on Computer Vision (ICCV)}, \bibinfo{year}{2021}, pp.
  \bibinfo{pages}{10933--10942}.
\bibitem[{Boston-Dynamics(2022)}]{bd2022}
\bibinfo{author}{Boston-Dynamics}, \bibinfo{title}{{I}nside the {L}ab:
  {R}obotics {A}fter {H}ours},
  \bibinfo{howpublished}{\url{https://www.youtube.com/watch?v=Jq0GknnKvXM}},
  \bibinfo{year}{2022}.
\bibitem[{Chow and Ghavamzadeh(2014)}]{chow2014cvar}
\bibinfo{author}{Y.~Chow}, \bibinfo{author}{M.~Ghavamzadeh},
\newblock \bibinfo{title}{Algorithms for cvar optimization in mdps},
\newblock in: \bibinfo{booktitle}{Proceedings of Advances in Neural Information
  Processing Systems (NeurIPS), Montreal, Quebec, Canada},
  \bibinfo{year}{2014}, pp. \bibinfo{pages}{3509--3517}.
\bibitem[{Chow et~al.(2017)Chow, Ghavamzadeh, Janson, and
  Pavone}]{chow2017riskconstrained}
\bibinfo{author}{Y.~Chow}, \bibinfo{author}{M.~Ghavamzadeh},
  \bibinfo{author}{L.~Janson}, \bibinfo{author}{M.~Pavone},
\newblock \bibinfo{title}{Risk-constrained reinforcement learning with
  percentile risk criteria},
\newblock \bibinfo{journal}{Journal of Machine Learning Research (JMLR)}
  \bibinfo{volume}{18} (\bibinfo{year}{2017}) \bibinfo{pages}{6070–6120}.
\bibitem[{{Fisac} et~al.(2019){Fisac}, {Akametalu}, {Zeilinger}, {Kaynama},
  {Gillula}, and {Tomlin}}]{fisac2019AGS}
\bibinfo{author}{J.~F. {Fisac}}, \bibinfo{author}{A.~K. {Akametalu}},
  \bibinfo{author}{M.~N. {Zeilinger}}, \bibinfo{author}{S.~{Kaynama}},
  \bibinfo{author}{J.~{Gillula}}, \bibinfo{author}{C.~J. {Tomlin}},
\newblock \bibinfo{title}{A general safety framework for learning-based control
  in uncertain robotic systems},
\newblock \bibinfo{journal}{IEEE Transactions on Automatic Control (TAC)}
  \bibinfo{volume}{64} (\bibinfo{year}{2019}) \bibinfo{pages}{2737--2752}.
\bibitem[{Fisac et~al.(2019)Fisac, Lugovoy, Rubies-Royo, Ghosh, and
  Tomlin}]{fisac2019bridging}
\bibinfo{author}{J.~F. Fisac}, \bibinfo{author}{N.~F. Lugovoy},
  \bibinfo{author}{V.~Rubies-Royo}, \bibinfo{author}{S.~Ghosh},
  \bibinfo{author}{C.~J. Tomlin},
\newblock \bibinfo{title}{Bridging hamilton-jacobi safety analysis and
  reinforcement learning},
\newblock in: \bibinfo{booktitle}{Proceedings of the International Conference
  on Robotics and Automation (ICRA)}, \bibinfo{year}{2019}, pp.
  \bibinfo{pages}{8550--8556}. \DOIprefix\doi{10.1109/ICRA.2019.8794107}.
\bibitem[{Hsu et~al.(2021)Hsu, Rubies-Royo, Tomlin, and Fisac}]{hsu2021safety}
\bibinfo{author}{K.-C. Hsu}, \bibinfo{author}{V.~Rubies-Royo},
  \bibinfo{author}{C.~J. Tomlin}, \bibinfo{author}{J.~F. Fisac},
\newblock \bibinfo{title}{Safety and liveness guarantees through reach-avoid
  reinforcement learning},
\newblock in: \bibinfo{booktitle}{Proceedings of Robotics: Science and
  Systems}, \bibinfo{address}{Virtual}, \bibinfo{year}{2021}.
  \DOIprefix\doi{10.15607/RSS.2021.XVII.077}.
\bibitem[{Srinivasan et~al.(2020)Srinivasan, Eysenbach, Ha, Tan, and
  Finn}]{srinivasan2020learning}
\bibinfo{author}{K.~Srinivasan}, \bibinfo{author}{B.~Eysenbach},
  \bibinfo{author}{S.~Ha}, \bibinfo{author}{J.~Tan}, \bibinfo{author}{C.~Finn},
  \bibinfo{title}{Learning to be safe: Deep rl with a safety critic},
  \bibinfo{year}{2020}. \href{http://arxiv.org/abs/2010.14603}{\tt
  arXiv:2010.14603}.
\bibitem[{Thananjeyan et~al.(2021)Thananjeyan, Balakrishna, Nair, Luo,
  Srinivasan, Hwang, Gonzalez, Ibarz, Finn, and
  Goldberg}]{thananjeyan2021recovery}
\bibinfo{author}{B.~Thananjeyan}, \bibinfo{author}{A.~Balakrishna},
  \bibinfo{author}{S.~Nair}, \bibinfo{author}{M.~Luo},
  \bibinfo{author}{K.~Srinivasan}, \bibinfo{author}{M.~Hwang},
  \bibinfo{author}{J.~E. Gonzalez}, \bibinfo{author}{J.~Ibarz},
  \bibinfo{author}{C.~Finn}, \bibinfo{author}{K.~Goldberg},
\newblock \bibinfo{title}{Recovery {RL}: Safe reinforcement learning with
  learned recovery zones},
\newblock \bibinfo{journal}{IEEE Robotics and Automation Letters (RAL)}
  \bibinfo{volume}{6} (\bibinfo{year}{2021}) \bibinfo{pages}{4915--4922}.
\bibitem[{Zhou and Doyle(1998)}]{zhou1998essentials}
\bibinfo{author}{K.~Zhou}, \bibinfo{author}{J.~C. Doyle},
  \bibinfo{title}{Essentials of robust control}, volume \bibinfo{volume}{104},
  \bibinfo{publisher}{Prentice hall Upper Saddle River, NJ},
  \bibinfo{year}{1998}.
\bibitem[{Xu and Chen(2002)}]{xu2002robust}
\bibinfo{author}{S.~Xu}, \bibinfo{author}{T.~Chen},
\newblock \bibinfo{title}{Robust h-infinity control for uncertain stochastic
  systems with state delay},
\newblock \bibinfo{journal}{IEEE Transactions on Automatic Control (TAC)}
  \bibinfo{volume}{47} (\bibinfo{year}{2002}) \bibinfo{pages}{2089--2094}.
\bibitem[{Majumdar and Tedrake(2017)}]{majumdar2017funnel}
\bibinfo{author}{A.~Majumdar}, \bibinfo{author}{R.~Tedrake},
\newblock \bibinfo{title}{Funnel libraries for real-time robust feedback motion
  planning},
\newblock \bibinfo{journal}{The International Journal of Robotics Research
  (IJRR)} \bibinfo{volume}{36} (\bibinfo{year}{2017})
  \bibinfo{pages}{947--982}.
\bibitem[{Singh et~al.(2017)Singh, Majumdar, Slotine, and
  Pavone}]{singh2019robust}
\bibinfo{author}{S.~Singh}, \bibinfo{author}{A.~Majumdar},
  \bibinfo{author}{J.-J. Slotine}, \bibinfo{author}{M.~Pavone},
\newblock \bibinfo{title}{Robust online motion planning via contraction theory
  and convex optimization},
\newblock in: \bibinfo{booktitle}{Proceedings of the IEEE International
  Conference on Robotics and Automation (ICRA)}, \bibinfo{year}{2017}, pp.
  \bibinfo{pages}{5883--5890}. \DOIprefix\doi{10.1109/ICRA.2017.7989693}.
\bibitem[{Majumdar et~al.(2021)Majumdar, Farid, and Sonar}]{majumdar2021pac}
\bibinfo{author}{A.~Majumdar}, \bibinfo{author}{A.~Farid},
  \bibinfo{author}{A.~Sonar},
\newblock \bibinfo{title}{{PAC}-{B}ayes {C}ontrol: Learning policies that
  provably generalize to novel environments},
\newblock \bibinfo{journal}{The International Journal of Robotics Research
  (IJRR)} \bibinfo{volume}{40} (\bibinfo{year}{2021})
  \bibinfo{pages}{574--593}.
\bibitem[{Farid et~al.(2021)Farid, Veer, and Majumdar}]{farid2021task}
\bibinfo{author}{A.~Farid}, \bibinfo{author}{S.~Veer},
  \bibinfo{author}{A.~Majumdar},
\newblock \bibinfo{title}{Task-driven out-of-distribution detection with
  statistical guarantees for robot learning},
\newblock in: \bibinfo{booktitle}{Proceedings of the Conference on Robot
  Learning (CoRL)}, \bibinfo{year}{2021}.
\bibitem[{Veer and Majumdar(2021)}]{veer2020probably}
\bibinfo{author}{S.~Veer}, \bibinfo{author}{A.~Majumdar},
\newblock \bibinfo{title}{Probably approximately correct vision-based planning
  using motion primitives},
\newblock in: \bibinfo{booktitle}{Proceedings of the 2020 Conference on Robot
  Learning (CoRL)}, volume \bibinfo{volume}{155} of
  \textit{\bibinfo{series}{Proceedings of Machine Learning Research}},
  \bibinfo{publisher}{PMLR}, \bibinfo{year}{2021}, pp.
  \bibinfo{pages}{1001--1014}.
\bibitem[{Garc{{\'i}}a and Fern{{\'a}}ndez(2015)}]{garcia2015survey}
\bibinfo{author}{J.~Garc{{\'i}}a}, \bibinfo{author}{F.~Fern{{\'a}}ndez},
\newblock \bibinfo{title}{A comprehensive survey on safe reinforcement
  learning},
\newblock \bibinfo{journal}{Journal of Machine Learning Research (JMLR)}
  \bibinfo{volume}{16} (\bibinfo{year}{2015}) \bibinfo{pages}{1437--1480}.
\bibitem[{Bansal et~al.(2017)Bansal, Chen, Herbert, and
  Tomlin}]{bansal2017hamilton}
\bibinfo{author}{S.~Bansal}, \bibinfo{author}{M.~Chen},
  \bibinfo{author}{S.~Herbert}, \bibinfo{author}{C.~J. Tomlin},
\newblock \bibinfo{title}{Hamilton-jacobi reachability: A brief overview and
  recent advances},
\newblock in: \bibinfo{booktitle}{Proceedings of the IEEE 56th Annual
  Conference on Decision and Control (CDC)}, \bibinfo{year}{2017}, pp.
  \bibinfo{pages}{2242--2253}. \DOIprefix\doi{10.1109/CDC.2017.8263977}.
\bibitem[{Fisac et~al.(2015)Fisac, Chen, Tomlin, and Sastry}]{fisac2015reach}
\bibinfo{author}{J.~F. Fisac}, \bibinfo{author}{M.~Chen},
  \bibinfo{author}{C.~J. Tomlin}, \bibinfo{author}{S.~S. Sastry},
\newblock \bibinfo{title}{{Reach-Avoid Problems with Time-Varying Dynamics,
  Targets and Constraints}},
\newblock in: \bibinfo{booktitle}{Proceedings of the 18th International
  Conference on Hybrid Systems: Computation and Control}, HSCC ’15,
  \bibinfo{address}{New York, NY, USA}, \bibinfo{year}{2015}, p.
  \bibinfo{pages}{11–20}. \DOIprefix\doi{10.1145/2728606.2728612}.
\bibitem[{Leung et~al.(2020)Leung, Schmerling, Zhang, Chen, Talbot, Gerdes, and
  Pavone}]{leung2020infusing}
\bibinfo{author}{K.~Leung}, \bibinfo{author}{E.~Schmerling},
  \bibinfo{author}{M.~Zhang}, \bibinfo{author}{M.~Chen},
  \bibinfo{author}{J.~Talbot}, \bibinfo{author}{J.~C. Gerdes},
  \bibinfo{author}{M.~Pavone},
\newblock \bibinfo{title}{On infusing reachability-based safety assurance
  within planning frameworks for human–robot vehicle interactions},
\newblock \bibinfo{journal}{The International Journal of Robotics Research}
  \bibinfo{volume}{39} (\bibinfo{year}{2020}) \bibinfo{pages}{1326--1345}.
\bibitem[{Cheng et~al.(2019)Cheng, Orosz, Murray, and Burdick}]{cheng2019e2e}
\bibinfo{author}{R.~Cheng}, \bibinfo{author}{G.~Orosz}, \bibinfo{author}{R.~M.
  Murray}, \bibinfo{author}{J.~W. Burdick},
\newblock \bibinfo{title}{End-to-end safe reinforcement learning through
  barrier functions for safety-critical continuous control tasks},
\newblock in: \bibinfo{booktitle}{Proceedings of the Thirty-Third AAAI
  Conference on Artificial Intelligence}, AAAI'19/IAAI'19/EAAI'19,
  \bibinfo{publisher}{AAAI Press}, \bibinfo{year}{2019}.
  \DOIprefix\doi{10.1609/aaai.v33i01.33013387}.
\bibitem[{Dalal et~al.(2018)Dalal, Dvijotham, Vecerik, Hester, Paduraru, and
  Tassa}]{dalal2018safe}
\bibinfo{author}{G.~Dalal}, \bibinfo{author}{K.~Dvijotham},
  \bibinfo{author}{M.~Vecerik}, \bibinfo{author}{T.~Hester},
  \bibinfo{author}{C.~Paduraru}, \bibinfo{author}{Y.~Tassa},
  \bibinfo{title}{Safe exploration in continuous action spaces},
  \bibinfo{year}{2018}. \href{http://arxiv.org/abs/1801.08757}{\tt
  arXiv:1801.08757}.
\bibitem[{Chen et~al.(2021)Chen, Francis, Oh, Nyberg, and
  Herbert}]{chen2021safe}
\bibinfo{author}{B.~Chen}, \bibinfo{author}{J.~Francis},
  \bibinfo{author}{J.~Oh}, \bibinfo{author}{E.~Nyberg}, \bibinfo{author}{S.~L.
  Herbert}, \bibinfo{title}{Safe autonomous racing via approximate reachability
  on ego-vision}, \bibinfo{year}{2021}.
  \href{http://arxiv.org/abs/2110.07699}{\tt arXiv:2110.07699}.
\bibitem[{Berkenkamp et~al.(2016)Berkenkamp, Schoellig, and
  Krause}]{berkenkamp2016safe}
\bibinfo{author}{F.~Berkenkamp}, \bibinfo{author}{A.~P. Schoellig},
  \bibinfo{author}{A.~Krause},
\newblock \bibinfo{title}{Safe controller optimization for quadrotors with
  gaussian processes},
\newblock in: \bibinfo{booktitle}{Proceedings of the IEEE International
  Conference on Robotics and Automation (ICRA)}, \bibinfo{year}{2016}, pp.
  \bibinfo{pages}{491--496}. \DOIprefix\doi{10.1109/ICRA.2016.7487170}.
\bibitem[{Koller et~al.(2018)Koller, Berkenkamp, Turchetta, and
  Krause}]{koller2018learning}
\bibinfo{author}{T.~Koller}, \bibinfo{author}{F.~Berkenkamp},
  \bibinfo{author}{M.~Turchetta}, \bibinfo{author}{A.~Krause},
\newblock \bibinfo{title}{Learning-based model predictive control for safe
  exploration},
\newblock in: \bibinfo{booktitle}{Proceedings of the IEEE Conference on
  Decision and Control (CDC)}, \bibinfo{year}{2018}, pp.
  \bibinfo{pages}{6059--6066}. \DOIprefix\doi{10.1109/CDC.2018.8619572}.
\bibitem[{Liu et~al.(2020)Liu, Shi, Chung, Anandkumar, and Yue}]{liu2020robust}
\bibinfo{author}{A.~Liu}, \bibinfo{author}{G.~Shi}, \bibinfo{author}{S.-J.
  Chung}, \bibinfo{author}{A.~Anandkumar}, \bibinfo{author}{Y.~Yue},
\newblock \bibinfo{title}{Robust regression for safe exploration in control},
\newblock in: \bibinfo{booktitle}{Proceedings of the 2nd Conference on Learning
  for Dynamics and Control}, volume \bibinfo{volume}{120} of
  \textit{\bibinfo{series}{Proceedings of Machine Learning Research}},
  \bibinfo{publisher}{PMLR}, \bibinfo{year}{2020}, pp.
  \bibinfo{pages}{608--619}. \URLprefix
  \url{https://proceedings.mlr.press/v120/liu20a.html}.
\bibitem[{Vapnik and Chervonenkis(2015)}]{vapnik2015uniform}
\bibinfo{author}{V.~N. Vapnik}, \bibinfo{author}{A.~Y. Chervonenkis},
\newblock \bibinfo{title}{On the uniform convergence of relative frequencies of
  events to their probabilities},
\newblock in: \bibinfo{booktitle}{Measures of complexity},
  \bibinfo{publisher}{Springer}, \bibinfo{year}{2015}, pp.
  \bibinfo{pages}{11--30}.
\bibitem[{Bousquet et~al.(2003)Bousquet, Boucheron, and
  Lugosi}]{bousquet2003introduction}
\bibinfo{author}{O.~Bousquet}, \bibinfo{author}{S.~Boucheron},
  \bibinfo{author}{G.~Lugosi},
\newblock \bibinfo{title}{Introduction to statistical learning theory},
\newblock in: \bibinfo{booktitle}{Summer school on machine learning},
  \bibinfo{organization}{Springer}, \bibinfo{year}{2003}, pp.
  \bibinfo{pages}{169--207}.
\bibitem[{McAllester(1999)}]{mcallester1999some}
\bibinfo{author}{D.~A. McAllester},
\newblock \bibinfo{title}{Some pac-bayesian theorems},
\newblock \bibinfo{journal}{Machine Learning} \bibinfo{volume}{37}
  (\bibinfo{year}{1999}) \bibinfo{pages}{355--363}.
\bibitem[{Dziugaite and Roy(2017)}]{dziugaite2017computing}
\bibinfo{author}{G.~K. Dziugaite}, \bibinfo{author}{D.~M. Roy},
\newblock \bibinfo{title}{Computing nonvacuous generalization bounds for deep
  (stochastic) neural networks with many more parameters than training data},
\newblock in: \bibinfo{booktitle}{Proceedings of the Thirty-Third Conference on
  Uncertainty in Artificial Intelligence (UAI), Sydney, Australia, August
  11-15}, \bibinfo{year}{2017}.
\bibitem[{P{\'e}rez-Ortiz et~al.(2021)P{\'e}rez-Ortiz, Rivasplata,
  Shawe-Taylor, and Szepesv{\'a}ri}]{perez2021tighter}
\bibinfo{author}{M.~P{\'e}rez-Ortiz}, \bibinfo{author}{O.~Rivasplata},
  \bibinfo{author}{J.~Shawe-Taylor}, \bibinfo{author}{C.~Szepesv{\'a}ri},
\newblock \bibinfo{title}{Tighter risk certificates for neural networks},
\newblock \bibinfo{journal}{Journal of Machine Learning Research (JMLR)}
  \bibinfo{volume}{22} (\bibinfo{year}{2021}).
\bibitem[{Ren et~al.(2021)Ren, Veer, and Majumdar}]{ren2020generalization}
\bibinfo{author}{A.~Z. Ren}, \bibinfo{author}{S.~Veer},
  \bibinfo{author}{A.~Majumdar},
\newblock \bibinfo{title}{Generalization guarantees for imitation learning},
\newblock in: \bibinfo{booktitle}{Proceedings of the 2020 Conference on Robot
  Learning (CoRL)}, volume \bibinfo{volume}{155} of
  \textit{\bibinfo{series}{Proceedings of Machine Learning Research}},
  \bibinfo{publisher}{PMLR}, \bibinfo{year}{2021}, pp.
  \bibinfo{pages}{1426--1442}.
\bibitem[{Gurgen et~al.(2021)Gurgen, Majumdar, and Veer}]{gurgen2021learning}
\bibinfo{author}{A.~E. Gurgen}, \bibinfo{author}{A.~Majumdar},
  \bibinfo{author}{S.~Veer},
\newblock \bibinfo{title}{Learning provably robust motion planners using funnel
  libraries},
\newblock \bibinfo{journal}{arXiv preprint arXiv:2111.08733}
  (\bibinfo{year}{2021}).
\bibitem[{Agarwal et~al.(2021)Agarwal, Veer, Ren, and
  Majumdar}]{agarwal2021stronger}
\bibinfo{author}{A.~Agarwal}, \bibinfo{author}{S.~Veer}, \bibinfo{author}{A.~Z.
  Ren}, \bibinfo{author}{A.~Majumdar},
\newblock \bibinfo{title}{Stronger generalization guarantees for robot learning
  by combining generative models and real-world data},
\newblock \bibinfo{journal}{arXiv preprint arXiv:2111.08761}
  (\bibinfo{year}{2021}).
\bibitem[{Farid et~al.(2022)Farid, Snyder, Ren, and
  Majumdar}]{farid2022failure}
\bibinfo{author}{A.~Farid}, \bibinfo{author}{D.~Snyder}, \bibinfo{author}{A.~Z.
  Ren}, \bibinfo{author}{A.~Majumdar},
\newblock \bibinfo{title}{Failure prediction with statistical guarantees for
  vision-based robot control},
\newblock in: \bibinfo{booktitle}{Proceedings of the Robotics: Science and
  Systems (RSS)}, \bibinfo{year}{2022}.
\bibitem[{Eysenbach et~al.(2019)Eysenbach, Gupta, Ibarz, and
  Levine}]{eysenbach2018diayn}
\bibinfo{author}{B.~Eysenbach}, \bibinfo{author}{A.~Gupta},
  \bibinfo{author}{J.~Ibarz}, \bibinfo{author}{S.~Levine},
\newblock \bibinfo{title}{Diversity is all you need: Learning skills without a
  reward function},
\newblock in: \bibinfo{booktitle}{Proceedings of the International Conference
  on Learning Representations (ICLR)}, \bibinfo{year}{2019}.
\bibitem[{Bonin-Font et~al.(2008)Bonin-Font, Ortiz, and
  Oliver}]{bonin2008visual}
\bibinfo{author}{F.~Bonin-Font}, \bibinfo{author}{A.~Ortiz},
  \bibinfo{author}{G.~Oliver},
\newblock \bibinfo{title}{Visual navigation for mobile robots: A survey},
\newblock \bibinfo{journal}{Journal of Intelligent and Robotic Systems}
  \bibinfo{volume}{53} (\bibinfo{year}{2008}) \bibinfo{pages}{263--296}.
\bibitem[{Sim and Little(2006)}]{sim2006autonomous}
\bibinfo{author}{R.~Sim}, \bibinfo{author}{J.~J. Little},
\newblock \bibinfo{title}{Autonomous vision-based exploration and mapping using
  hybrid maps and {Rao-Blackwellised} particle filters},
\newblock in: \bibinfo{booktitle}{Proceedings of the IEEE/RSJ International
  Conference on Intelligent Robots and Systems (IROS)}, \bibinfo{year}{2006},
  pp. \bibinfo{pages}{2082--2089}. \DOIprefix\doi{10.1109/IROS.2006.282485}.
\bibitem[{Thrun and B{\"u}cken(1996)}]{thrun1996integrating}
\bibinfo{author}{S.~Thrun}, \bibinfo{author}{A.~B{\"u}cken},
\newblock \bibinfo{title}{Integrating grid-based and topological maps for
  mobile robot navigation},
\newblock in: \bibinfo{booktitle}{Proceedings of the AAAI Conference on
  Artificial Intelligence}, \bibinfo{year}{1996}, pp.
  \bibinfo{pages}{944--951}.
\bibitem[{Bansal et~al.(2020)Bansal, Tolani, Gupta, Malik, and
  Tomlin}]{bansal2020combining}
\bibinfo{author}{S.~Bansal}, \bibinfo{author}{V.~Tolani},
  \bibinfo{author}{S.~Gupta}, \bibinfo{author}{J.~Malik},
  \bibinfo{author}{C.~Tomlin},
\newblock \bibinfo{title}{Combining optimal control and learning for visual
  navigation in novel environments},
\newblock in: \bibinfo{booktitle}{Proceedings of the 2020 Conference on Robot
  Learning (CoRL)}, volume \bibinfo{volume}{100} of
  \textit{\bibinfo{series}{Proceedings of Machine Learning Research}},
  \bibinfo{publisher}{PMLR}, \bibinfo{year}{2020}, pp.
  \bibinfo{pages}{420--429}.
\bibitem[{Gupta et~al.(2017)Gupta, Davidson, Levine, Sukthankar, and
  Malik}]{gupta2017cognitive}
\bibinfo{author}{S.~Gupta}, \bibinfo{author}{J.~Davidson},
  \bibinfo{author}{S.~Levine}, \bibinfo{author}{R.~Sukthankar},
  \bibinfo{author}{J.~Malik},
\newblock \bibinfo{title}{Cognitive mapping and planning for visual
  navigation},
\newblock in: \bibinfo{booktitle}{Proceedings of the IEEE Conference on
  Computer Vision and Pattern Recognition (CVPR)}, \bibinfo{year}{2017}, pp.
  \bibinfo{pages}{2616--2625}.
\bibitem[{Richter and Roy(2017)}]{richter2017safe}
\bibinfo{author}{C.~Richter}, \bibinfo{author}{N.~Roy},
\newblock \bibinfo{title}{Safe visual navigation via deep learning and novelty
  detection},
\newblock in: \bibinfo{booktitle}{Proceedings of Robotics: Science and Systems
  (RSS)}, \bibinfo{address}{Cambridge, Massachusetts}, \bibinfo{year}{2017}.
  \DOIprefix\doi{10.15607/RSS.2017.XIII.064}.
\bibitem[{Wellhausen et~al.(2020)Wellhausen, Ranftl, and
  Hutter}]{wellhausen2020safe}
\bibinfo{author}{L.~Wellhausen}, \bibinfo{author}{R.~Ranftl},
  \bibinfo{author}{M.~Hutter},
\newblock \bibinfo{title}{Safe robot navigation via multi-modal anomaly
  detection},
\newblock \bibinfo{journal}{IEEE Robotics and Automation Letters (RAL)}
  \bibinfo{volume}{5} (\bibinfo{year}{2020}) \bibinfo{pages}{1326--1333}.
\bibitem[{L{\"u}tjens et~al.(2019)L{\"u}tjens, Everett, and
  How}]{lutjens2019safe}
\bibinfo{author}{B.~L{\"u}tjens}, \bibinfo{author}{M.~Everett},
  \bibinfo{author}{J.~P. How},
\newblock \bibinfo{title}{Safe reinforcement learning with model uncertainty
  estimates},
\newblock in: \bibinfo{booktitle}{Proceedings of the International Conference
  on Robotics and Automation (ICRA)}, \bibinfo{organization}{IEEE},
  \bibinfo{year}{2019}, pp. \bibinfo{pages}{8662--8668}.
\bibitem[{Kahn et~al.(2017)Kahn, Villaflor, Pong, Abbeel, and
  Levine}]{kahn2017uncertainty}
\bibinfo{author}{G.~Kahn}, \bibinfo{author}{A.~Villaflor},
  \bibinfo{author}{V.~Pong}, \bibinfo{author}{P.~Abbeel},
  \bibinfo{author}{S.~Levine},
\newblock \bibinfo{title}{Uncertainty-aware reinforcement learning for
  collision avoidance},
\newblock \bibinfo{journal}{arXiv preprint arXiv:1702.01182}
  (\bibinfo{year}{2017}).
\bibitem[{Bajcsy et~al.(2019)Bajcsy, Bansal, Bronstein, Tolani, and
  Tomlin}]{bajcsy2019efficient}
\bibinfo{author}{A.~Bajcsy}, \bibinfo{author}{S.~Bansal},
  \bibinfo{author}{E.~Bronstein}, \bibinfo{author}{V.~Tolani},
  \bibinfo{author}{C.~J. Tomlin},
\newblock \bibinfo{title}{An efficient reachability-based framework for
  provably safe autonomous navigation in unknown environments},
\newblock in: \bibinfo{booktitle}{Proceedings of the IEEE 58th Conference on
  Decision and Control (CDC)}, \bibinfo{organization}{IEEE},
  \bibinfo{year}{2019}, pp. \bibinfo{pages}{1758--1765}.
\bibitem[{Li et~al.(2020)Li, Bansal, Giovanis, Tolani, Tomlin, and
  Chen}]{li2020generating}
\bibinfo{author}{A.~Li}, \bibinfo{author}{S.~Bansal},
  \bibinfo{author}{G.~Giovanis}, \bibinfo{author}{V.~Tolani},
  \bibinfo{author}{C.~Tomlin}, \bibinfo{author}{M.~Chen},
\newblock \bibinfo{title}{Generating robust supervision for learning-based
  visual navigation using hamilton-jacobi reachability},
\newblock in: \bibinfo{booktitle}{Proceedings of the 2nd Conference on Learning
  for Dynamics and Control}, volume \bibinfo{volume}{120} of
  \textit{\bibinfo{series}{Proceedings of Machine Learning Research}},
  \bibinfo{publisher}{PMLR}, \bibinfo{year}{2020}, pp.
  \bibinfo{pages}{500--510}.
\bibitem[{Ramos et~al.(2019)Ramos, Possas, and Fox}]{ramos2019bayessim}
\bibinfo{author}{F.~Ramos}, \bibinfo{author}{R.~C. Possas},
  \bibinfo{author}{D.~Fox},
\newblock \bibinfo{title}{Bayessim: adaptive domain randomization via
  probabilistic inference for robotics simulators},
\newblock \bibinfo{journal}{arXiv preprint arXiv:1906.01728}
  (\bibinfo{year}{2019}).
\bibitem[{Chebotar et~al.(2019)Chebotar, Handa, Makoviychuk, Macklin, Issac,
  Ratliff, and Fox}]{chebotar2019closing}
\bibinfo{author}{Y.~Chebotar}, \bibinfo{author}{A.~Handa},
  \bibinfo{author}{V.~Makoviychuk}, \bibinfo{author}{M.~Macklin},
  \bibinfo{author}{J.~Issac}, \bibinfo{author}{N.~Ratliff},
  \bibinfo{author}{D.~Fox},
\newblock \bibinfo{title}{Closing the sim-to-real loop: Adapting simulation
  randomization with real world experience},
\newblock in: \bibinfo{booktitle}{2019 International Conference on Robotics and
  Automation (ICRA)}, \bibinfo{year}{2019}.
\bibitem[{Lim et~al.(2021)Lim, Huang, Chen, Wang, Ichnowski, Seita, Laskey, and
  Goldberg}]{lim2021planar}
\bibinfo{author}{V.~Lim}, \bibinfo{author}{H.~Huang}, \bibinfo{author}{L.~Y.
  Chen}, \bibinfo{author}{J.~Wang}, \bibinfo{author}{J.~Ichnowski},
  \bibinfo{author}{D.~Seita}, \bibinfo{author}{M.~Laskey},
  \bibinfo{author}{K.~Goldberg},
\newblock \bibinfo{title}{Planar robot casting with real2sim2real
  self-supervised learning},
\newblock \bibinfo{journal}{arXiv preprint arXiv:2111.04814}
  (\bibinfo{year}{2021}).
\bibitem[{Mehta et~al.(2020)Mehta, Diaz, Golemo, Pal, and
  Paull}]{mehta2020active}
\bibinfo{author}{B.~Mehta}, \bibinfo{author}{M.~Diaz},
  \bibinfo{author}{F.~Golemo}, \bibinfo{author}{C.~J. Pal},
  \bibinfo{author}{L.~Paull},
\newblock \bibinfo{title}{Active domain randomization},
\newblock in: \bibinfo{booktitle}{Conference on Robot Learning},
  \bibinfo{year}{2020}, pp. \bibinfo{pages}{1162--1176}.
\bibitem[{Muratore et~al.(2021)Muratore, Eilers, Gienger, and
  Peters}]{muratore2021data}
\bibinfo{author}{F.~Muratore}, \bibinfo{author}{C.~Eilers},
  \bibinfo{author}{M.~Gienger}, \bibinfo{author}{J.~Peters},
\newblock \bibinfo{title}{Data-efficient domain randomization with bayesian
  optimization},
\newblock \bibinfo{journal}{IEEE Robotics and Automation Letters}
  \bibinfo{volume}{6} (\bibinfo{year}{2021}) \bibinfo{pages}{911--918}.
\bibitem[{Cutler et~al.(2014)Cutler, Walsh, and How}]{cutler2014multifidelity}
\bibinfo{author}{M.~Cutler}, \bibinfo{author}{T.~J. Walsh},
  \bibinfo{author}{J.~P. How},
\newblock \bibinfo{title}{Reinforcement learning with multi-fidelity
  simulators},
\newblock in: \bibinfo{booktitle}{Proceedings of the IEEE/RSJ International
  Conference on Robotics and Automation (ICRA)}, \bibinfo{year}{2014}.
\bibitem[{Shafer and Vovk(2008)}]{shafer2008tutorial}
\bibinfo{author}{G.~Shafer}, \bibinfo{author}{V.~Vovk},
\newblock \bibinfo{title}{A tutorial on conformal prediction.},
\newblock \bibinfo{journal}{Journal of Machine Learning Research}
  \bibinfo{volume}{9} (\bibinfo{year}{2008}).
\bibitem[{Haarnoja et~al.(2018)Haarnoja, Zhou, Abbeel, and
  Levine}]{Haarnoja2018SAC}
\bibinfo{author}{T.~Haarnoja}, \bibinfo{author}{A.~Zhou},
  \bibinfo{author}{P.~Abbeel}, \bibinfo{author}{S.~Levine},
\newblock \bibinfo{title}{{Soft Actor-Critic}: Off-policy maximum entropy deep
  reinforcement learning with a stochastic actor},
\newblock in: \bibinfo{booktitle}{Proceedings of the 35th International
  Conference on Machine Learning}, volume~\bibinfo{volume}{80} of
  \textit{\bibinfo{series}{Proceedings of Machine Learning Research}},
  \bibinfo{publisher}{PMLR}, \bibinfo{year}{2018}, pp.
  \bibinfo{pages}{1861--1870}.
\bibitem[{Alshiekh et~al.(2018)Alshiekh, Bloem, Ehlers, K\"{o}nighofer, Niekum,
  and Topcu}]{alshiekh2017safe}
\bibinfo{author}{M.~Alshiekh}, \bibinfo{author}{R.~Bloem},
  \bibinfo{author}{R.~Ehlers}, \bibinfo{author}{B.~K\"{o}nighofer},
  \bibinfo{author}{S.~Niekum}, \bibinfo{author}{U.~Topcu},
\newblock \bibinfo{title}{Safe reinforcement learning via shielding},
\newblock in: \bibinfo{booktitle}{Proceedings of the Thirty-Second AAAI
  Conference on Artificial Intelligence and Thirtieth Innovative Applications
  of Artificial Intelligence Conference and Eighth AAAI Symposium on
  Educational Advances in Artificial Intelligence}, \bibinfo{publisher}{AAAI
  Press}, \bibinfo{year}{2018}.
\bibitem[{Jabri et~al.(2019)Jabri, Hsu, Gupta, Eysenbach, Levine, and
  Finn}]{jabri2019unsupervised}
\bibinfo{author}{A.~Jabri}, \bibinfo{author}{K.~Hsu},
  \bibinfo{author}{A.~Gupta}, \bibinfo{author}{B.~Eysenbach},
  \bibinfo{author}{S.~Levine}, \bibinfo{author}{C.~Finn},
\newblock \bibinfo{title}{Unsupervised curricula for visual meta-reinforcement
  learning},
\newblock in: \bibinfo{booktitle}{Advances in Neural Information Processing
  Systems (NeurIPS)}, volume~\bibinfo{volume}{32}, \bibinfo{year}{2019}, pp.
  \bibinfo{pages}{10519--10530}.
\bibitem[{Kumar et~al.(2020)Kumar, Kumar, Levine, and Finn}]{kumar2020smerl}
\bibinfo{author}{S.~Kumar}, \bibinfo{author}{A.~Kumar},
  \bibinfo{author}{S.~Levine}, \bibinfo{author}{C.~Finn},
\newblock \bibinfo{title}{One solution is not all you need: Few-shot
  extrapolation via structured {MaxEnt RL}},
\newblock in: \bibinfo{booktitle}{Advances in Neural Information Processing
  Systems (NeurIPS)}, volume~\bibinfo{volume}{33}, \bibinfo{publisher}{Curran
  Associates, Inc.}, \bibinfo{year}{2020}, pp. \bibinfo{pages}{8198--8210}.
\bibitem[{Sharma et~al.(2020)Sharma, Gu, Levine, Kumar, and
  Hausman}]{sharma2020dynamicsaware}
\bibinfo{author}{A.~Sharma}, \bibinfo{author}{S.~Gu},
  \bibinfo{author}{S.~Levine}, \bibinfo{author}{V.~Kumar},
  \bibinfo{author}{K.~Hausman},
\newblock \bibinfo{title}{Dynamics-aware unsupervised discovery of skills},
\newblock in: \bibinfo{booktitle}{Proceedings of the International Conference
  on Learning Representations (ICLR)}, \bibinfo{year}{2020}.
\bibitem[{Langford and Caruana(2002)}]{langford2002not}
\bibinfo{author}{J.~Langford}, \bibinfo{author}{R.~Caruana},
\newblock \bibinfo{title}{{(Not)} bounding the true error},
\newblock in: \bibinfo{booktitle}{Advances in Neural Information Processing
  Systems (NeurIPS)}, volume~\bibinfo{volume}{14}, \bibinfo{publisher}{MIT
  Press}, \bibinfo{year}{2002}.
\bibitem[{Levine et~al.(2018)Levine, Pastor, Krizhevsky, Ibarz, and
  Quillen}]{levine2016learning}
\bibinfo{author}{S.~Levine}, \bibinfo{author}{P.~Pastor},
  \bibinfo{author}{A.~Krizhevsky}, \bibinfo{author}{J.~Ibarz},
  \bibinfo{author}{D.~Quillen},
\newblock \bibinfo{title}{Learning hand-eye coordination for robotic grasping
  with deep learning and large-scale data collection},
\newblock \bibinfo{journal}{The International Journal of Robotics Research}
  \bibinfo{volume}{37} (\bibinfo{year}{2018}) \bibinfo{pages}{421--436}.
\bibitem[{{Q}uartz(2019)}]{amazon19}
\bibinfo{author}{{Q}uartz}, \bibinfo{title}{{A}mazon - this company built one
  of the world’s most efficient warehouses by embracing chaos},
  \bibinfo{howpublished}{\url{https://classic.qz.com/perfect-company-2/1172282/this-company-built-one-of-the-worlds-most-efficient-warehouses-by-embracing-chaos/}},
  \bibinfo{year}{2019}.
\bibitem[{{F}uture{Car}(2022)}]{waymo22}
\bibinfo{author}{{F}uture{Car}}, \bibinfo{title}{{A} look at how waymo's
  self-driving test fleet safely traveled 2.7 million miles in san francisco
  last year},
  \bibinfo{howpublished}{\url{https://www.futurecar.com/5158/A-Look-at-How-Waymos-Self-Driving-Test-Fleet-Safely-Traveled-2-7-Million-Miles-in-San-Francisco-Last-Year}},
  \bibinfo{year}{2022}.
\bibitem[{Ichnowski et~al.(2022)Ichnowski, Chen, Dharmarajan, Adebola,
  Danielczuk, Mayoral-Vilches, Zhan, Xu, Ghassemi, Kubiatowicz
  et~al.}]{ichnowski2022fogros}
\bibinfo{author}{J.~Ichnowski}, \bibinfo{author}{K.~Chen},
  \bibinfo{author}{K.~Dharmarajan}, \bibinfo{author}{S.~Adebola},
  \bibinfo{author}{M.~Danielczuk}, \bibinfo{author}{V.~Mayoral-Vilches},
  \bibinfo{author}{H.~Zhan}, \bibinfo{author}{D.~Xu},
  \bibinfo{author}{R.~Ghassemi}, \bibinfo{author}{J.~Kubiatowicz}, et~al.,
\newblock \bibinfo{title}{Fogros 2: An adaptive and extensible platform for
  cloud and fog robotics using ros 2},
\newblock \bibinfo{journal}{arXiv preprint arXiv:2205.09778}
  (\bibinfo{year}{2022}).
\bibitem[{Eysenbach et~al.(2018)Eysenbach, Gu, Ibarz, and
  Levine}]{eysenbach2018leave}
\bibinfo{author}{B.~Eysenbach}, \bibinfo{author}{S.~Gu},
  \bibinfo{author}{J.~Ibarz}, \bibinfo{author}{S.~Levine},
\newblock \bibinfo{title}{Leave no trace: Learning to reset for safe and
  autonomous reinforcement learning},
\newblock in: \bibinfo{booktitle}{Proceedings of the 6th International
  Conference on Learning Representations (ICLR)}, \bibinfo{year}{2018}.
  \URLprefix \url{https://openreview.net/forum?id=S1vuO-bCW}.
\bibitem[{Borsos et~al.(2020)Borsos, Mutny, and Krause}]{borsos2020coresets}
\bibinfo{author}{Z.~Borsos}, \bibinfo{author}{M.~Mutny},
  \bibinfo{author}{A.~Krause},
\newblock \bibinfo{title}{Coresets via bilevel optimization for continual
  learning and streaming},
\newblock \bibinfo{journal}{Advances in Neural Information Processing Systems}
  \bibinfo{volume}{33} (\bibinfo{year}{2020}) \bibinfo{pages}{14879--14890}.
\bibitem[{Guedj(2019)}]{guedj2019primer}
\bibinfo{author}{B.~Guedj},
\newblock \bibinfo{title}{A primer on pac-bayesian learning},
\newblock \bibinfo{journal}{arXiv preprint arXiv:1901.05353}
  (\bibinfo{year}{2019}).
\bibitem[{Arora et~al.(2018)Arora, Ge, Neyshabur, and
  Zhang}]{arora2018stronger}
\bibinfo{author}{S.~Arora}, \bibinfo{author}{R.~Ge},
  \bibinfo{author}{B.~Neyshabur}, \bibinfo{author}{Y.~Zhang},
\newblock \bibinfo{title}{Stronger generalization bounds for deep nets via a
  compression approach},
\newblock in: \bibinfo{booktitle}{International Conference on Machine
  Learning}, \bibinfo{organization}{PMLR}, \bibinfo{year}{2018}, pp.
  \bibinfo{pages}{254--263}.

\end{thebibliography}

\appendix

\section{Derivations for Inducing Diversity into Backup Policy Update \label{app:diverse_backup}}
We add observation-conditional mutual information term to the loss function of backup policy.

\begin{align}
    L(\theta) := \ & \expect_{\obs, \latent} \bigg[ \expect_{\ctrl \sim \pi_\theta(\cdot|\obs, \latent)} \Big[ Q(\obs, \ctrl; \latent) \Big] \bigg] - \nu I(A; Z|O) \nonumber \\
    = \ & \expect_{\obs, \latent} \bigg[ \expect_{\ctrl \sim \pi_\theta(\cdot|\obs, \latent)} \Big[ Q(\obs, \ctrl; \latent) \Big] \bigg] - \nu \entropy (A | O) + \nu \entropy (A | Z, O) \nonumber \\
    = \ & \expect_{\obs, \latent} \bigg[ \expect_{\ctrl \sim \pi_\theta(\cdot|\obs, \latent)} \Big[ Q(\obs, \ctrl; \latent)  - \nu \log \pi_\theta (\ctrl | \obs, \latent) \Big] \bigg] + \nu \expect_\obs \bigg[ \expect_{\ctrl \sim p(\cdot | \obs)} \Big[ \log  p(\ctrl | \obs) \Big] \bigg ]
\end{align}
We then approximate the expectation by the transitions sampled from the replay buffer as
\begin{equation}
    L(\theta) \approx \expect_{(\obs, \latent) \sim \replay, \ctrl \sim \pi_\theta(\cdot|\obs, \latent)} \bigg[ 
        Q(\obs, \ctrl; \latent)  - \nu \log \pi_\theta (\ctrl | \obs, \latent)
        + \nu \log p(\ctrl | \obs)
    \bigg].
\end{equation}
Finally, we approximate the marginal with the latent variables sampled from the distribution (empirical measure) as
\begin{equation}
    L(\theta) \approx \expect_{(\obs, \latent) \sim \replay, \ctrl \sim \pi_\theta(\cdot|\obs, \latent)} \Bigg[
        Q(\obs, \ctrl; \latent)  - \nu \log \pi_\theta (\ctrl | \obs, \latent)
        + \nu \frac{1}{n_s} \log \sum_{i=1, z_i\sim p(z)}^{n_s} \pi_\theta (\ctrl | \obs, \latent_i)
    \Bigg].
\end{equation}

\section{Training Hyperparameters used in Experiments \label{app:hyperparams}}
We show the training hyperparameters used to generate the results in Fig.~\ref{fig:main}.

\begin{table}[h!]
\footnotesize
\centering
\begin{threeparttable}
    \caption{Hyperparameters for PAC{\myspace}Shield{\myspace}Perf in Sim training. Same neural network architecture is used for performance and backup policies.}
    \label{tab:sim-hyperparams}
    \begin{tabular}{cccc}
        \toprule
        & \multicolumn{3}{c}{Environment Setting} \\
        \cmidrule{2-4}
        \phantom{a} & Vanilla-Normal/Dynamics & Vanilla-Task & Advanced-Env \\ \midrule
        $\#$ training steps & 500000 & 1000000 & 4000000 \\
        Replay buffer size & 50000 (steps) & 100000 (steps) & 5000 (trajectories) \\
        Optimize frequency & 2000 & 2000 & 20000 \\
        $\#$ updater per optimize & 1000 & 1000 & 1000 \\
        Value shielding threshold & -0.05 & -0.05 & -0.05 \\ \midrule
        \multirow{2}{*}{\textbf{Latent Distribution}} \\ \\
        Latent dimension ($n_z$) & 20 & 20 & 30 \\
        Augmented reward coefficient ($\beta$) & 2 & 2 & 2 \\
        Prior standard deviation & 2 & 2 & 2 \\ \midrule
        \multirow{2}{*}{\textbf{Optimization}} \\ \\
        Optimizer & Adam & Adam & Adam \\
        Batch size (Performance) & 128 & 128 & 128 \\
        Discount factor (Performance) & 0.99 & 0.99 & 0.99 \\
        Learning rate (Performance) & 0.0001 & 0.0001 & 0.0001 \\
        Batch size (Backup) & 128 & 128 & 128 \\
        Discount factor (Backup) & 0.8 $\rightarrow$ 0.999 & 0.8 $\rightarrow$ 0.999 & 0.8 $\rightarrow$ 0.99 \\ 
        Learning rate (Backup) & 0.0001 & 0.0001 & 0.001 \\ \midrule
        \multirow{2}{*}{\textbf{NN Architecture}} \\ \\
        Input channels & 3 & 3 & 22\tnote{a} \\
        CNN kernel size & [5,3,3] & [5,3,3] & [7,5,3] \\
        CNN stride & [2,2,2] & [2,2,2] & [4,3,2] \\
        CNN channel size & [8,16,32] & [8,16,32] & [16,32,64] \\
        MLP dimensions & [130+$n_z$\tnote{b} \ ,128] & [132+$n_z$\tnote{b} \ ,128] & [248+$n_z$\tnote{b} \ ,256,256] \\ \midrule
        \multirow{2}{*}{\textbf{Hardware Resource}} \\ \\
        $\#$ CPU threads & 8 & 8 & 16 \\
        GPU & Nvidia V100 (16GB) & Nvidia V100 (16GB) & Nvidia A100 (40GB) \\
        Runtime & 8 hours & 14 hours & 12 hours \\ 
        \bottomrule
    \end{tabular}
    \begin{tablenotes}
        \item[a] We stack 4 previous RGB images while skipping 3 frames between two images and concatenate the stacked images with the first 10 elements of the latent variable (each element is repeated to match the same shape of a channel in an image).
        \item[b] The input of the first linear layer is composed of the output from the convolutional layers, latent variables and auxiliary signals, which is $128 + n_z + 2$ in Vanilla-Normal/Dynamics, $128 + n_z + 4$ in Vanilla-Task and $256 + (n_z-10) + 2$ in Advanced-Env.
    \end{tablenotes}
\end{threeparttable}
\end{table}

\begin{table}[h!]
\footnotesize
\centering
\caption{Hyperparameters for PAC{\myspace}Shield{\myspace}Perf in Lab training.}
\label{tab:lab-hyperparams}
\begin{tabular}{ccc}
    \toprule
    & \multicolumn{2}{c}{Environment Setting} \\
    \cmidrule{2-3}
    \phantom{a} & Vanilla-Env & Advanced-Env \\ \midrule
    $\#$ training steps & 500000 & 3000000 \\
    Replay buffer size & 50000 (steps) & 5000 (trajectories) \\
    Optimize frequency & 2000 & 20000 \\
    $\#$ updater per optimize & 1000 & 1000 \\
    Value shielding threshold & -0.05 & -0.05 \\ 
    The number of environments ($N$) & 1000 & 1000 \\ \midrule
    \multirow{2}{*}{\textbf{Optimization}} \\ \\
    Learning rate for latent mean & 0.0001 & 0.0001 \\
    Learning rate for latent std & 0.0001 & 0.0001 \\
    KL-divergence coefficient ($\alpha$) & 1 & 2 \\
    Optimizer & Adam & Adam \\
    Batch size (Performance) & 1024 & 128 \\
    Discount factor (Performance) & 0.99 & 0.99 \\
    Learning rate (Performance) & 0.0001 & 0.0001 \\ \midrule
    \multirow{2}{*}{\textbf{PAC-Bayes Bound}} \\ \\
    The number of latent variables ($L$) & 1000 & 1000 \\
    Precision ($\delta$) & 0.01 & 0.01 \\ \midrule
    \multirow{2}{*}{\textbf{Hardware Resource}} \\ \\
    $\#$ CPU threads & 8 & 8 \\
    GPU & Nvidia V100 (16GB) & Nvidia A100 (40GB) \\
    Runtime & 6 hours & 16 hours \\ 
    \bottomrule
\end{tabular}
\end{table}

\newpage
\section{Environment Setup for Advanced-Env \label{app:env-setup}}

In order to train the navigating agent in realistic environments before Real deployment, we use the 3D-FRONT (3D Furnished Rooms with layOuts and semaNTics) dataset \cite{fu20213dfront} that offers a larger number of synthetic indoor scenes with professionally designed layouts and high-quality textured furniture. This is the richest dataset we find suitable to indoor navigation task, training with domain randomization and PAC-Bayes Control framework often requires more than $1000$ environments.

\begin{figure}[!ht]
    \centering
    \includegraphics[width=\textwidth]{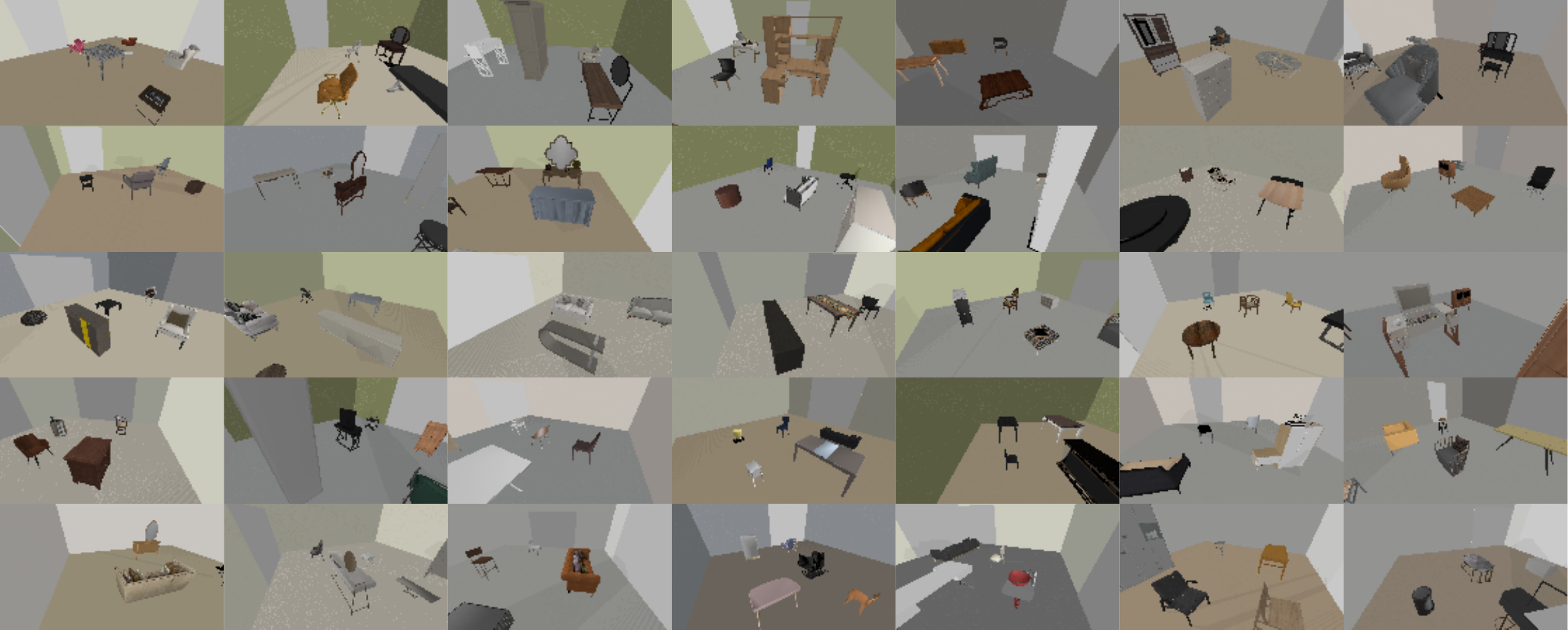}
    \caption{\textbf{Samples of robot observations in Sim training of Advanced-Env:} for better view here, the virtual camera is placed at a higher location than the robot.}
    \label{fig:app-sim-rooms}
\end{figure}

For Sim training, we use $7m \times 7m$ undecorated rooms as room layouts, and randomly placing $5$ pieces of furniture from the dataset. We use $4$ categories of furniture: Soft ($2701$ pieces available), Chair ($1775$ pieces available), Cabinet/Shelf/Desk ($5725$ pieces available), Table ($1090$ pieces available). We also randomly sample textures from the dataset to add to the walls and floor: for walls, we use categories Tile, Wallpaper, and Paint ($911$ images available in total), and for floor, we use Flooring, Stone, Wood, Marble, Solid Wood Flooring ($466$ images available in total). We set the minimum clearance between furniture, around the initial location, and around the goal to be $1m$. The minimum distance between the initial location and the goal is $5m$. Fig.~\ref{fig:app-sim-rooms} shows samples of observations at the initial locations. For Advanced-Dense Lab where the furniture density is higher, we place $6$ instead of $5$ pieces of furniture, and the minimum clearance is $0.8m$ instead of $1m$.

\begin{figure}[!ht]
    \centering
    \includegraphics[width=\textwidth]{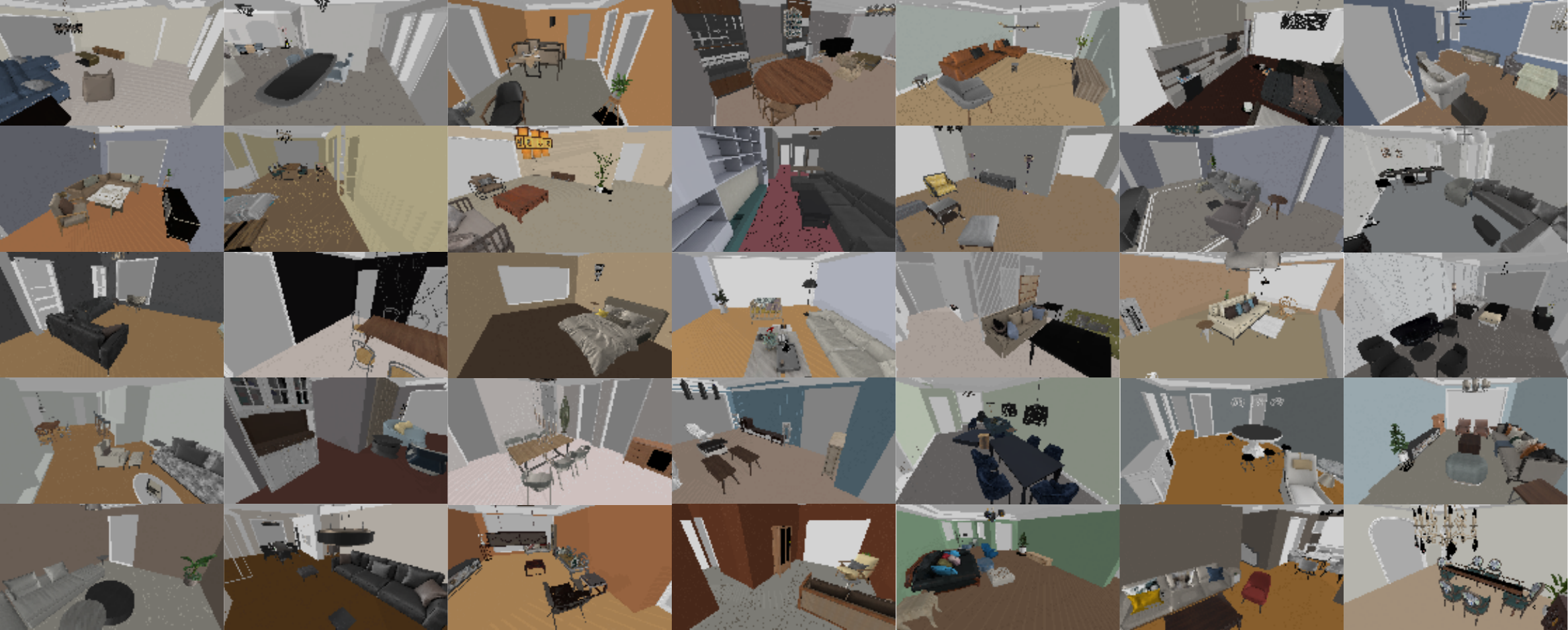}
    \caption{\textbf{Samples of robot observations in Advanced-Realistic Lab:} for better view here, the virtual camera is placed at a higher location than the robot.}
    \label{fig:app-lab-rooms} 
\end{figure}

For Lab training, we instead use the professionally designed room layouts (with furniture configuration) from the dataset. The dataset contains $6813$ different house layouts (each with multiple rooms). Since our focus is on obstacle avoidance with relatively short horizon, in each house, we try to sample initial and goal locations within one room. Unfortunately the dataset does not provide corresponding wall and floor textures in each layout, and we resort to random samples as in Vanilla-Env. Again we maintain a minimum clearance of $1m$ between furniture, around the initial and goal locations. To check the environment is solvable, we extract a 2D occupancy map for each room and run the Dijkstra algorithm. We also ensure there is at least one piece of furniture along the line connecting the initial and goal locations. We tend to find that many rooms are too crowded or the found path does not have enough clearance for the quadrupedal robot (about $0.5m$ wide). At the end, we are able to process about $2000$ room environments, which are then split for training and testing. Fig.~\ref{fig:app-lab-rooms} shows samples of observations at the initial locations. 

\end{document}